\documentclass[preprint,5p,12pt]{elsarticle}

\usepackage{amssymb}

\usepackage{lipsum}
\usepackage{amsmath}
\usepackage[labelformat=simple]{subcaption}
\usepackage{booktabs}
\usepackage{array}
\usepackage[hidelinks]{hyperref}
\usepackage{tikz}
\usepackage{pgfplots}  %
\usepackage{siunitx}  %
\usepackage{textcomp, gensymb}  %
\usepackage{mathtools}  %
\usepackage{tensor}  %
\usepackage{pifont}
\usepackage[subpreambles=false]{standalone}
\usepackage{multirow}
\usepackage{threeparttable}
\usepackage{placeins}
\usepackage{color}
\usepackage{tikz_utils}  %
\usepackage{colortbl}
\usepackage[noabbrev, capitalize]{cleveref}

\NewDocumentCommand{\rot}{O{45} O{1em} m}{\makebox[#2][l]{\rotatebox{#1}{#3}}}%

\definecolor{LightGray}{rgb}{0.96,0.96,0.96}
\definecolor{crimson2143940}{RGB}{214,39,40}
\definecolor{darkgray176}{RGB}{176,176,176}
\definecolor{darkorange25512714}{RGB}{255,127,14}
\definecolor{forestgreen4416044}{RGB}{44,160,44}
\definecolor{goldenrod18818934}{RGB}{188,189,34}
\definecolor{gray127}{RGB}{127,127,127}
\definecolor{mediumpurple148103189}{RGB}{148,103,189}
\definecolor{orchid227119194}{RGB}{227,119,194}
\definecolor{sienna1408675}{RGB}{140,86,75}
\definecolor{steelblue31119180}{RGB}{31,119,180}

\definecolor{gold}{RGB}{255,215,0}
\definecolor{silver}{RGB}{192,192,192}
\definecolor{bronze}{RGB}{204,153,102}

\newcommand\thefontsize{The current font size is: \csname f@size\endcsname pt}

\newcommand{\xmark}{%
\tikz[scale=0.17,opacity=0.1] {
    \node[opacity=0, outer sep=0, inner sep=0, minimum width=0pt, minimum height=0pt] (text) at (0.5, 0.0) {\clap{\smash{\ding{55}}}};%
    \draw[line width=0.4,line cap=round] (0,0) to [bend left=6] (1,1);
    \draw[line width=0.4,line cap=round] (0.2,0.95) to [bend right=3] (0.8,0.05);
}}
\newcommand{\cmark}{%
\tikz[scale=0.17] {
    \node[opacity=0, outer sep=0, inner sep=0, minimum width=0pt, minimum height=0pt] (text) at (0.5, 0.0) {\clap{\smash{\ding{51}}}};%
    \draw[line width=0.4,line cap=round] (0.25,0) to [bend left=10] (1,1);%
    \draw[line width=0.4,line cap=round] (0,0.35) to [bend right=1] (0.23,0);%
}}
\newcommand{\omark}{%
\tikz[scale=0.17, opacity=0.6] {
    \node[opacity=0, outer sep=0, inner sep=0, minimum width=0pt, minimum height=0pt] (text) at (0.5, 0.0) {\smash{\clap{O}}};%
    \draw[rotate=-10, line width=0.4,line cap=round] (0.5, 0.5) circle(0.41 and 0.5);
}}
\newcommand{\gold}{%
\tikz[scale=0.17, opacity=1] {
    \node[opacity=0, outer sep=0, inner sep=0, minimum width=0pt, minimum height=0pt] (text) at (0.5, 0.0) {\smash{\clap{O}}};%
    \draw[gold,rotate=-10, line width=0.4,line cap=round,fill] (0.5, 0.63) circle(0.39 and 0.45);
}}
\newcommand{\silver}{%
\tikz[scale=0.17, opacity=1] {
    \node[opacity=0, outer sep=0, inner sep=0, minimum width=0pt, minimum height=0pt] (text) at (0.5, 0.0) {\smash{\clap{O}}};%
    \draw[silver,rotate=-10, line width=0.4,line cap=round,fill] (0.5, 0.63) circle(0.39 and 0.45);
}}
\newcommand{\bronze}{%
\tikz[scale=0.17, opacity=1] {
    \node[opacity=0, outer sep=0, inner sep=0, minimum width=0pt, minimum height=0pt] (text) at (0.5, 0.0) {\smash{\clap{O}}};%
    \draw[bronze,rotate=-10, line width=0.4,line cap=round,fill] (0.5, 0.63) circle(0.39 and 0.45);
}}

\newcolumntype{L}[1]{>{\raggedright\let\newline\\\arraybackslash\hspace{0pt}}m{#1}}
\newcolumntype{C}[1]{>{\centering\let\newline\\\arraybackslash\hspace{0pt}}m{#1}}
\newcolumntype{R}[1]{>{\raggedleft\let\newline\\\arraybackslash\hspace{0pt}}m{#1}}

\pgfplotsset{compat=1.17} 

\usepgfplotslibrary{units}

\newtheorem{problem}{Problem}
\newtheorem{definition}{Definition}

\journal{Robotics and Autonomous Systems}

\begin{document}\sloppy

\begin{frontmatter}

\title{RGB-D-Based Categorical Object Pose and Shape Estimation:\\Methods, Datasets, and Evaluation}

\author[kth]{Leonard Bruns\texorpdfstring{\corref{cor1}}{}}
\cortext[cor1]{Corresponding author.}
\ead{leonardb@kth.se}
\author[kth]{Patric Jensfelt}
\ead{patric@kth.se}

\address[kth]{Division of Robotics, Perception and Learning (RPL), KTH Royal Institute of Technology, Teknikringen 14, 11428 Stockholm, Sweden}

\begin{abstract}
Recently, various methods for 6D pose and shape estimation of objects at a per-category level have been proposed. This work provides an overview of the field in terms of methods, datasets, and evaluation protocols. First, an overview of existing works and their commonalities and differences is provided. Second, we take a critical look at the predominant evaluation protocol, including metrics and datasets. Based on the findings, we propose a new set of metrics, contribute new annotations for the Redwood dataset, and evaluate state-of-the-art methods in a fair comparison. The results indicate that existing methods do not generalize well to unconstrained orientations and are actually heavily biased towards objects being upright. We provide an easy-to-use evaluation toolbox with well-defined metrics, methods, and dataset interfaces, which allows evaluation and comparison with various state-of-the-art approaches (\url{https://github.com/roym899/pose_and_shape_evaluation}).
\end{abstract}


\end{frontmatter}

\section{Introduction}

We consider the problem of pose and shape estimation on a per-category level. Classic grasp and motion planning methods often assume that full knowledge of pose and shape is available, making them difficult to apply with partial sensor information. Estimating the full shape and pose of objects promises to bridge this gap from partial sensor information to an actionable representation. Although pure shape reconstruction itself is sufficient for some tasks \cite{varley2017shape}, categorical pose estimation additionally provides a reference frame. Such a categorical reference frame could, for example, further enable alignment of objects \cite{sucar2020nodeslam} and pose-dependent grasp computation (e.g., an upside-down mug has to be grasped differently from an upright mug).

In the last two years, various learning-based categorical pose and shape estimation methods have been proposed. With a large number of changes between different works (i.e., varying loss functions, network architecture, training protocols, etc.) the main contributions of individual works and their relation to prior work can become hard to identify. We provide an overview of existing methods and identify notable commonalities and differences, which so far have been less discussed in the literature.

Regarding datasets, most of the current methods are trained and evaluated on the two datasets proposed by \cite{wang2019normalized}, called CAMERA and REAL. The CAMERA dataset is a large dataset of real RGB-D tabletop scenes with synthetic objects superimposed on top of the table. The REAL dataset is a smaller real-world dataset of tabletop sequences with objects that have been scanned and tracked for the purpose of evaluating categorical pose estimation. Notably, both datasets only contain upright objects, which opens the question of how well existing methods generalize to less constrained settings.

To answer this question, we contribute a set of annotations to evaluate unconstrained 6D pose and shape estimation. Our annotations consist of meshes and poses for handheld objects of three categories in the Redwood dataset \cite{choi2016large}. The objects in this dataset are freely rotated in front of the camera by a human hand causing occlusions. The orientations vary significantly more than in the datasets by \cite{wang2019normalized}.

Most methods evaluate pose estimation by following the same evaluation protocol as originally proposed by \cite{wang2019normalized}. The method in~\cite{wang2019normalized} combines mask detection and pose estimation into a single network and therefore evaluates pose estimation with \emph{average precision} (AP), which is a common detection metric. However, many of the subsequent methods assume that the mask is an input to their method, making AP an unnatural evaluation metric, as it makes the results unnecessarily difficult to interpret. Therefore, we propose a set of simpler metrics that use ground-truth masks and categories to evaluate pose estimation.

Pose estimation with shape reconstruction was first demonstrated by \cite{chen2020learning} and \cite{tian2020shape}. Both methods independently used \emph{chamfer distance} (CD) as their reconstruction metric, which is now commonly used by methods performing shape reconstruction. However, \cite{tatarchenko2019single} showed that CD is not a good measure of reconstruction quality and other metrics better correlate with perceived reconstruction quality. Therefore, in this work, we advocate for the use of F-score to evaluate reconstruction quality and propose a new evaluation protocol for both shape reconstruction and pose estimation.

To summarize, our contributions are:
\begin{itemize}
    \item an extensive overview of existing methods for categorical pose and shape estimation,
    \item a well-defined evaluation protocol,
    \item a challenging set of annotations to evaluate unconstrained pose and shape estimation,
    \item a fair evaluation of various state-of-the-art methods, and
    \item an open-source evaluation toolbox for categorical pose and shape estimation including methods, datasets, and metrics.
\end{itemize}

\section{Related Work}

Since the introduction of the BOP benchmark suite \cite{hodavn2020bop}, great progress has been made in the task of instance-level pose estimation, where a mesh of the target object is available. However, the more general task of category-level pose estimation has only recently received more attention. Compared to instance-level pose estimation, category-level pose estimation is more challenging due to the large possible variations in shape and appearance.

\citet{wang2019normalized} introduced the first deep learning-based method to address the 6D pose estimation problem at the per-category level. They introduced two datasets: CAMERA, a synthetic dataset that combines real scenes with meshes from the ShapeNet dataset \cite{chang2015} and REAL, a smaller real-world dataset that is mainly used for fine-tuning and evaluation. Their method is based on the \emph{normalized object coordinate space} (NOCS), in which objects of one category have a common alignment. The projection of the NOCS coordinates in the image plane (also called NOCS map) is predicted by extending Mask R-CNN \cite{he2017mask} with an additional head. From this prediction, the 6D pose and scale can be estimated by employing the Umeyama algorithm \cite{umeyama1991least} with RANSAC \cite{fischler1981random} for outlier removal.

In \cite{wang2019normalized}, the NOCS map was predicted using only RGB information. Since geometry typically varies less than appearance for a fixed category, several methods were proposed to more directly incorporate the observed point set into the prediction. \citet{chen2020learning} introduced \emph{canonical shape space} (CASS), which is a latent shape representation learned in a variational autoencoder \cite{kingma2013auto} framework. Their method regresses the latent shape, orientation, and position directly from the cropped image and the set of observed points. As a byproduct of their method, they also reconstruct the full canonical point set. \citet{tian2020shape} introduced \emph{shape prior deformation} (SPD), which uses a canonical point set and predicts a deformation based on the observed RGB-D information. 

Various follow-up works modify SPD's network architecture and training scheme to achieve further improvements. For example, CR-Net \cite{wang2021category} uses a recurrent architecture to iteratively deform the canonical point set, SGPA \cite{chen2021sgpa} uses a transformer architecture to more effectively adjust the canonical point set, and DPDN \cite{lin2022category} employs consistency-based losses for additional self-supervised learning.

Another notable line of work follows an analysis-by-synthesis approach. Analysis-by-synthesis approaches can be seen as the category-level analog to mesh-based pose refinement \cite{li2020deepim,wang2019densefusion,labbe2020cosypose,shugurov2021multi} for instance-level pose estimation. Since at category-level no ground-truth mesh is available for refinement, analysis-by-synthesis approaches typically integrate a generative shape model into the pipeline to jointly optimize a latent shape representation and the 6D pose at inference time. 

\citet{chen2020category} proposed the first such analysis-by-synthesis framework in which the latent representation of a generative model is iteratively optimized to fit the observed color image. The generative model allows to generate novel views of the object, but a full reconstruction is not readily available. Instead of minimizing the RGB discrepancy, iCaps \cite{deng2022icaps} and SDFEst \cite{bruns2022sdfest} perform iterative optimization based on the depth observation. A full RGB-D-based optimization is proposed in \cite{irshad2022shapo}. %

Aside from these RGB-D-based methods, a few RGB-based methods have been proposed.
\citet{manhardt2020cps++} estimate pose and a point set from monocular images. While their method only uses RGB information during inference, they show that unannotated depth data can be used to close the synthetic-to-real domain gap in a self-supervised fashion. Going beyond point set-based shape representations, \citet{lee2021category} extend Mesh R-CNN \cite{gkioxari2019mesh} to predict pose and mesh from a single RGB image, while \citet{engelmann2021points} propose shape reconstruction in a representation-agnostic way by classifying the closest matching object from a database.

Other methods such as \cite{chen2021fs,li2021leveraging,lin2021dualposenet,di2022gpv,lee2022uda} predict pose and bounding box without reconstructing the full shape of the object. For the evaluation presented in this work, we limit ourselves to methods that perform both reconstruction and pose estimation, although our evaluation protocol could in principle be used for pure pose estimation methods as well.

Other works perform shape estimation on video sequences. FroDO \cite{runz2020frodo} employs DeepSDF \cite{park2019deepsdf} to represent the shape and uses tracked keypoints from an RGB video to optimize the latent shape description such that the keypoints lie on the 0-isosurface of the signed distance field. NodeSLAM \cite{sucar2020nodeslam} uses the depth data to optimize the shape of multiple objects and the camera pose jointly. Both of these works simplify the pose estimation problem, by assuming objects to be upright on a planar surface.

In this work the focus is on categorical 6D pose and shape estimation methods, which do not explicility constrain the poses. An extensive overview of existing methods is provided in Section \ref{sec:methods}.

\paragraph{Evaluation} The most established benchmark dataset for categorical pose estimation is the REAL275 dataset proposed by \cite{wang2019normalized}. We will take a critical look at that dataset in Section \ref{sec:real275} and show that it only evaluates a constrained set of orientations, hiding inherent difficulties of the task, such as multimodal orientation distributions. \cite{wang2019normalized} also proposed \emph{average precision} as a metric to evaluate pose estimation.

To evaluate shape reconstruction most papers currently use \emph{chamfer distance} (CD) \cite{akizuki2021,chen2020learning,tian2020shape}, which was introduced to measure the difference of point sets by \cite{fan2017point}. However, \cite{tatarchenko2019single} noted that CD is not robust to outliers, that is, outlier points can skew the resulting metric based on their distance to the ground truth. Therefore, the authors advocate using a robust thresholded metric such as F-score \cite{knapitsch2017} to measure the quality of reconstruction.

\begin{table*}[h!]
    \centering
    \scriptsize
    \setlength{\tabcolsep}{3pt}
    \renewcommand{\arraystretch}{1.3}
    \begin{threeparttable}
        \caption{Overview of categorical pose and shape estimation methods sorted by publication date. \textbf{Bold} methods are included in our toolbox and evaluation. See text for further explanation and discussion.}\label{tab:overview}
        \begin{tabular}{lccccccccm{4.8cm}}
            \toprule
             \multirow{2}[3]{*}{\bfseries Method} & \multirow{2}[3]{*}{\bfseries Input} & \multirow{2}[3]{*}{\bfseries Shape} & \multirow{2}[3]{*}{\bfseries \shortstack{ Symmetry \\ Handling}} & \multicolumn{3}{c}{\bfseries Properties$\smash{^\ast}$} & \multicolumn{2}{c}{\bfseries Open source$\smash{^\dagger}$} & \multirow{2}[3]{*}{\bfseries Notes} \\
             \cmidrule(lr){5-7}
             \cmidrule(lr){8-9}
              &  &  & & \bfseries Det. & \bfseries MV & \bfseries Tr. & \bfseries  Infer. & \bfseries Train &  \\
             \midrule
             CPS++ \cite{manhardt2020cps++} & RGB & Point set & CD \cite{wang2019densefusion} & \cmark& \xmark & \xmark & \xmark & \xmark & Self-supervised sim-to-real (mask, depth)\\
             \rowcolor{LightGray} \citet{chen2020category} & RGB & Novel views & Input & \xmark & \xmark & \xmark & \cmark & \cmark & RGB-only analysis-by-synthesis \\
             \textbf{CASS} \cite{chen2020learning} & RGB-D & Point set & CD \cite{wang2019densefusion}  & \xmark& \xmark & \xmark & \cmark & \xmark & Variational autoencoder \cite{kingma2013auto} to learn canonical shape space \\
             \rowcolor{LightGray}
             \textbf{SPD} \cite{tian2020shape} & RGB-D & Point set & Norm.\ \cite{pitteri2019object} & \xmark & \xmark & \xmark & \cmark & \cmark & Deformation of shape prior \\
             SAR-Net \cite{lin2022sar} & D & Point set & Norm.\ \cite{pitteri2019object} & \xmark & \xmark & \xmark & \omark & \omark & Symmetry-based shape completion \\
             \rowcolor{LightGray}
             DISP6D \cite{wen2022disp6d} & RGB & Novel views & Codebook \cite{sundermeyer2018implicit} & \xmark & \xmark & \xmark & \cmark & \cmark & Implicit shape and pose learning \\
             \textbf{CR-Net} \cite{wang2021category} & RGB-D & Point set & Norm.\ \cite{pitteri2019object} & \xmark & \xmark & \xmark & \cmark & \cmark & Recurrent extension of SPD \cite{tian2020shape} \\
             \rowcolor{LightGray}
             \citet{lee2021category} & RGB & Mesh & Norm.\ \cite{pitteri2019object} & \xmark & \xmark & \xmark & \xmark & \xmark & Monocular metric mesh estimation \\
             \textbf{SGPA} \cite{chen2021sgpa} & RGB-D & Point set & Norm.\ \cite{pitteri2019object} & \xmark & \xmark & \xmark & \cmark & \cmark & Transformer extension of SPD \cite{tian2020shape} \\
             \rowcolor{LightGray}
             6D-ViT \cite{zou20226d}& RGB-D & Point set & Norm.\ \cite{pitteri2019object} & \xmark  & \xmark  & \xmark & \omark & \omark & Transformer-based feature extraction for SPD \cite{tian2020shape} \\
             ACR-Pose \cite{fan2021acr} & RGB-D & Point set & Norm.\ \cite{pitteri2019object} & \xmark & \xmark & \xmark & \xmark & \xmark & Adversarial extension of SPD \cite{tian2020shape} \\
             \rowcolor{LightGray}
             \textbf{ASM-Net} \cite{akizuki2021} & D & Point set & CD \cite{wang2019densefusion} & \xmark & \xmark & \xmark & \cmark & \xmark  & Active shape models \cite{cootes1995active} \\
             \textbf{iCaps} \cite{deng2022icaps} & D & Cont.\ SDF & Codebook \cite{sundermeyer2018implicit} & \xmark & \xmark & \cmark & \cmark & \cmark & Alternating pose refinement and shape estimation over time \\
             \rowcolor{LightGray}
             \citet{he2022towards} & D & Mesh &  CD \cite{wang2019densefusion} & \xmark & \xmark & \xmark & \xmark & \xmark & Fully self-supervised (depth) \\
             GPV-Pose \cite{di2022gpv} & D & Point set & Sym.\ axis \cite{chen2021fs} & \xmark & \xmark & \xmark & \cmark & \cmark & Consistency-based losses and pointwise bounding box prediction \\
             \rowcolor{LightGray}
             OLD-Net \cite{fan2022old} & RGB & Point set & Norm.\ \cite{pitteri2019object} & \cmark & \xmark & \xmark & \xmark & \xmark & Monocular extension of SPD \cite{tian2020shape} \\
             \citet{peng2022self} & RGB-D & Cont.\ SDF & Norm.\ \cite{pitteri2019object} & \xmark  & \xmark & \xmark & \cmark & \cmark & Self-supervised sim-to-real (depth) with DeepSDF \cite{park2019deepsdf} \\
             \rowcolor{LightGray}
             CenterSnap \cite{irshad2022centersnap} & RGB-D & Point set &  Norm.\ \cite{pitteri2019object} & \cmark & \xmark & \xmark & \cmark & \cmark & Single-stage detection, pose, and shape estimation  \\
             RePoNet \cite{fu2022wild6d} & RGB-D & Point set & Norm.\ \cite{pitteri2019object} / CD \cite{xiangposecnn,wang2021gdr} & \xmark & \xmark & \xmark & \cmark & \omark & Self-supervised sim-to-real (mask) \\
             \rowcolor{LightGray}
             \textbf{SDFEst} \cite{bruns2022sdfest} & D & Disc.\ SDF & Disc.\ SO(3) & \xmark & \cmark  & \xmark & \cmark & \cmark &  Depth-only analysis-by-synthesis \\
             \textbf{DPDN} \cite{lin2022category} & RGB-D & Point set & Norm.\ \cite{pitteri2019object} & \xmark & \xmark & \xmark & \cmark & \cmark & Self-supervised sim-to-real (consistency) \\
             \rowcolor{LightGray}
             SSP-Pose \cite{zhang2022ssp} & D & Point set & Sym.\ axis \cite{chen2021fs} / Min.\ \cite{wang2019normalized} & \xmark & \xmark & \xmark & \xmark & \xmark & Symmetry-aware and direct pose regression extension of SPD \cite{tian2020shape} \\
             ShAPO \cite{irshad2022shapo} & RGB-D & Cont.\ SDF & Norm.\ \cite{pitteri2019object} & \cmark & \xmark & \xmark & \cmark & \cmark & RGB-D analysis-by-synthesis \\
             \rowcolor{LightGray}
             \textbf{RBP-Pose} \cite{zhang2022rbp} & D & Point set & Sym.\ axis only \cite{chen2021fs} & \xmark & \xmark & \xmark & \cmark & \cmark & Integration of SPD \cite{tian2020shape} into GPV-Pose \cite{di2022gpv} \\
             gCasp \cite{li2022generative} & D & Cont.\ SDF & Min.\ \cite{wang2019normalized} & \xmark & \xmark & \xmark & \cmark & \cmark & Iterative optimization with semantic primitives \cite{hao2020dualsdf} \\
             \rowcolor{LightGray}
             \citet{zhang2022self} & RGB-D & Mesh & CD \cite{wang2019densefusion} & \xmark & \xmark & \xmark & \omark & \omark & Fully self-supervised (RGB, depth, mask, consistency) \\
             \bottomrule
        \end{tabular}
        \begin{tablenotes}
            \item[$^\ast$] Whether methods support detection (Det.), multi-view setups (MV) and tracking (Tr.) over time.
            \item[$^\dagger$] \omark\ denotes cases in which the paper mentions publishing the code, but it could not be found as of January 5, 2023.
        \end{tablenotes}
    \end{threeparttable}
\end{table*}

\section{Methods}\label{sec:methods}

An overview of the methods and their most notable differences is given in \cref{tab:overview}. Only methods that estimate both pose and shape and do not require an initial pose estimate are included. Therefore, pure categorial pose estimation methods such as \cite{wang2019normalized,chen2021fs,lee2022uda,you2022cppf} are excluded; tracking methods that require an initial pose estimate such as \cite{wang20206pack,weng2021captra,wen2021bundletrack} are excluded; and finally, methods that explicitly rely on objects being upright such as \cite{runz2020frodo,sucar2020nodeslam,li2021moltr} are excluded. %

Notably, only SDFEst \cite{bruns2022sdfest} supports multi-view setups, only iCaps \cite{deng2022icaps} supports tracking over time, and only three methods include the detection part in their pipeline \cite{manhardt2020cps++,irshad2022centersnap,irshad2022shapo}. The other methods assume that an off-the-shelf detector (typically Mask R-CNN \cite{he2017mask}) is available, but do not train it end-to-end with the pose and shape estimation part. This observation prompts us to question the predominant average precision-based evaluation protocol (see Section \ref{sec:evaluation}). 

We note that in some circumstances modular pipelines can be more desired than end-to-end pipelines. For example, by training the detection part separately from the pose and shape estimation part, larger detection-only datasets can be leveraged, whereas combined training of the detection part and the pose / shape estimation part typically requires annotations for both, which is harder to obtain at a large scale. Therefore, not supporting detection should not be seen as a major disadvantage for a given method. 

Similar arguments can be made for tracking and multi-view handling. For example, most methods could be combined with a specialized tracking method such as \cite{wang20206pack} or \cite{weng2021captra} to enable tracking over time. However, most of the methods in \cref{tab:overview} are purely discriminative in that they regress pose and shape in a feed-forward manner. Such methods are generally less flexible to extend to multi-view settings than methods that include a generative shape model in their pipeline, such as \cite{chen2020learning,deng2022icaps,bruns2022sdfest,irshad2022shapo,li2022generative}.

A notable line of work focusses on self-supervised approaches. Following the early work by \citet{manhardt2020cps++}, most works \cite{peng2022self,fu2022wild6d,lin2022category} focus on self-supervised learning in a sim-to-real context. That is, they first train in fully supervised fashion on synthetic data, and subsequently fine-tune without or with limited annotations on real data. \citet{he2022towards} and \citet{zhang2022self} have proposed fully self-supervised approaches that do not require initial training on synthetic data. In general, methods in this category differ in the used modality (see \textbf{Notes} column in Table \ref{tab:overview}). That is, methods formulate losses based on the depth data, mask, various consistencies that should hold, or combinations thereof. Self-supervised approaches are closely related to analysis-by-synthesis approaches \cite{chen2020category,bruns2022sdfest,irshad2022shapo}, which often employ similar losses, however, to optimize the estimate at inference time instead of the network at training time. 

Next, we will discuss two main dimensions in which the methods differ. First, the utilized shape representation and second, the handling of symmetries, and, in a wider sense, pose ambiguities.

\subsection{Shape Representation}

Various shape representations have been proposed in the literature. Shape representations vary in downstream usability (e.g., collision checking or grasp computation), efficiency, and flexibility. The latter referring to the fact that some representations can easily be converted into another, whereas the conversion for others is more involved or not well defined. Efficiency is often a trade-off with quality and depends on the employed hardware. Therefore, it is difficult to make general statements about efficiency.

\paragraph{Point sets} Most methods use point sets of fixed size to represent the shape. However, the exact way to predict the point sets varies. Most methods follow SPD \cite{tian2020shape} and predict an offset (often referred to as deformation) for each point in a mean shape that is defined per category. Notable exceptions are CASS \cite{chen2020learning}, which instead uses a FoldingNet-based \cite{yang2018foldingnet} variational autoencoder \cite{kingma2013auto} to predict the point set, and ASM-Net \cite{akizuki2021} which regresses the parameters of a previously learned active shape model \cite{cootes1995active}.

Point sets have the disadvantage of being a sparse shape representation and are therefore not a reliable representation for collision detection or grasp computation. Introducing an intermediate dense representation can alleviate this limitation; however, conversion quality is limited by the density of the point set. In our experiments, we observe that some methods predict outlier points or non-uniform point sets (i.e., uneven density). In general, high uniform density, and outlier-free point sets improve the downstream usability.

\paragraph{Meshes} \citet{lee2021category} follow Mesh R-CNN \cite{gkioxari2019mesh} and represent the shape as a mesh. Since meshes are difficult to parameterize directly, Mesh R-CNN first predicts a discretized occupancy grid, which defines the topology of the mesh. This mesh is subsequently refined (similar to the deformation approach for point sets) through multiple refinement stages. Meshes are a dense surface representation and are hence well-suited for geometric downstream tasks.

\paragraph{Novel views} Instead of explicitly regressing the shape, \citet{chen2020category} and DISP6D~\cite{wen2022disp6d} employ neural networks that generate RGB and optionally depth views given a latent shape representation and viewing direction. In contrast to neural field representations (see \emph{Continuous SDFs} below), multi-view consistency is not enforced by these methods, that is, there is no guarantee that the resulting views are consistent with each other. In principle, generated views can be converted to point sets or other representations by employing reconstruction methods on multiple synthesized views. However, this has not been demonstrated in any of the aforementioned works. Therefore, generating novel views on its own is typically not sufficient for many downstream tasks.

\paragraph{Discretized SDFs} SDFEst \cite{bruns2022sdfest} represents shape as a discretized signed distance field of fixed resolution. Similar to \cite{chen2020learning} a variational autoencoder is trained to learn the shape model. Compared to point sets and novel views, signed distance fields can readily be used for collision detection and grasp computation. Furthermore, the 0-isosurface of a signed distance field can be converted to a mesh using the marching cubes algorithm \cite{lorensen1987marching}.

\paragraph{Continuous SDFs} Following promising research on neural fields \cite{xie2022neural} for shape representation \cite{park2019deepsdf,mescheder2019occupancy,chen2019learning}, iCaps \cite{deng2022icaps}, ShAPO \cite{irshad2022shapo}, and gCasp \cite{li2022generative} employ SDF-based neural fields in the context of categorical pose and shape estimation to represent the shape. Neural fields are coordinate-based neural networks (e.g., $f_\theta: \mathbb{R}^3\mapsto d$) that for a given 3D coordinate predict a value (in this case the signed distance to the closest surface) at that coordinate. By conditioning such fields on a latent vector, they allow interpolation between shapes and mesh extraction at arbitrary resolution using the marching cubes algorithm \cite{lorensen1987marching}.

\subsection{Symmetry Handling}

A major difference between existing methods is their approach to handle ambiguities. Ambiguities can occur at a category-level (e.g., bowls and bottles typically have a symmetry axis), but also due to occlusions (including view-dependent self-occlusion). In principle, such ambiguities require learning a one-to-many mapping \cite{deng2022deep}. However, most approaches assume one-to-one mappings, which creates issues at training time when different targets occur for the same or similar inputs. In general, the ideal prediction for networks that can only predict a single estimate is poorly defined, when presented with contradicting targets, and therefore they typically fail to converge to a correct pose estimate.

Different approaches to this issue have been proposed, requiring various levels of annotation and providing various degrees of flexibility. Some approaches only handle predefined category-level symmetries, whereas other approaches can, in principle, handle general ambiguities. The first being likely more data-efficient, however, only applies to constrained settings, the latter being more general. Here we summarize the different approaches and discuss their respective advantages and disadvantages.

\paragraph{Normalization} One way of handling symmetries at a per-category level is to introduce a constraint that removes the degree of freedom around the symmetry axis leaving only one correct orientation. This approach and an analytic solution has first been proposed for instance-level pose estimation by \citet{pitteri2019object} and has subsequently been introduced to category-level pose estimation by \citet{tian2020shape}. Most works in the field that employ NOCS to estimate the object pose adopt the same symmetry handling. This approach has two practical disadvantages: it requires a priori specified category-level symmetries, and it cannot handle other forms of ambiguities. The former makes the assumption that all instances of a category exhibit the same axis-symmetry, which might not be the case. This could be alleviated, by per-instance symmetry annotations, which, however, requires additional annotation effort. The latter creates an issue when training on less constrained datasets. For example, when side-views of mugs are included, there can be two contradicting targets for such views. However, for view-constrained datasets with consistent per-category symmetries like the REAL275 dataset, this approach represents an efficient way of making an ambiguous dataset unimodal.

\paragraph{Minimum} \citet{wang2019normalized} proposed to handle symmetries by generating multiple ground-truth targets by rotating the ground-truth pose around the predefined symmetry axis in discrete steps. The loss will then be computed for each target, and only the smallest of these losses will be used to update the weights. Compared to the normalization approach, this approach introduces a small overhead at training time. Furthermore, an ambiguous interval remains due to the discrete nature of the approach.

\paragraph{Symmetry axis} Following FS-Net \cite{chen2021fs} some approaches \cite{di2022gpv,zhang2022ssp,zhang2022rbp} tailor the orientation representation to the predefined axis-symmetries. Specifically, these methods parameterize the rotation by two axes (similar to \cite{zhou2019continuity}), which is sufficient to construct a rotation matrix (i.e., the third vector must be orthogonal to the first two). For objects with category-level symmetries, one of the axes will be defined as the symmetry axis (e.g., the up-axis of a bottle), and only that one is supervised during training and used during inference.

\paragraph{Chamfer distance} Instead of directly supervising the pose, some methods \cite{chen2020learning,akizuki2021} 
employ chamfer distance (see Section \ref{sec:cd} for definition) between transformed point sets as their loss function. Since the chamfer distance is based on the minimum distance between nearest neighbors, it automatically handles symmetries. However, chamfer distance as a loss function introduces local minima that do not exist for direct pose supervision, which can yield undesired solutions.

\paragraph{Codebook} Codebook-based pose estimation was introduced by \citet{sundermeyer2018implicit} in the context of instance-based pose estimation. In short, an autoencoder is trained on synthetic data, which learns to reconstruct a given input image. Indistinguishable views will map to the same points in the latent space. A codebook is stored, which maps orientations to the corresponding latent representation. At inference time, the orientation can be inferred by finding the closest stored latent representation (and its associated orientation) to the encoded input. This approach automatically learns any form of ambiguity and has been adapted to categorical pose estimation by iCaps \cite{deng2022icaps} and DISP6D \cite{wen2022disp6d}. iCaps uses a categorical reference object when training the decoder (i.e., implying category-level ambiguities), whereas DISP6D modifies the codebook at inference time, which in principle supports inference of object-level ambiguities.

\paragraph{Input} As discussed in the beginning of this section, one-to-many mappings are more difficult to learn. However, many-to-one mappings are well defined and, therefore, easier to learn. \citet{chen2020category} propose a network that takes as input a latent shape description and the pose of the object and predicts the expected RGB view. To infer the pose, a large number of poses need to be sampled, and the resulting RGB image is compared with the observed one. While simplifying the learning, inferring the pose using such a network is more difficult and less efficient than with feed-forward approaches. However, similar to the codebook approach, all ambiguities are, in principle, supported, and no symmetry annotations are required.

\paragraph{Discretized SO(3)} SDFEst \cite{bruns2022sdfest} proposes to estimate the orientation by classifying into which cell of a discretized SO(3) grid the orientation falls. Similar to the codebook approach, this approach is, in theory, capable of handling general forms of ambiguities, since it learns all ambiguities purely from data and can represent arbitrary distributions over SO(3). A downside of this approach is that learning such a classification is likely less data-efficient compared to implicit rotation learning with a codebook as noted by \cite{sundermeyer2018implicit}.

\subsection{Methods Evaluated}\label{sec:evalmethods}

Next, we will summarize the categorical pose and shape estimation methods that are included in our evaluation. We limited ourselves to methods which are available open-source and perform both pose and shape estimation.

\paragraph{CASS} CASS \cite{chen2020learning} represents one of the first methods for estimating pose and shape. To learn the shape, it includes a generative model (based on a variational autoencoder \cite{kingma2013auto}) in its pipeline and regresses the pose, parametrized by a rotation matrix and translation. In contrast to most NOCS-based methods (see below), the pose is predicted in a correspondence-free manner, without the need for subsequent alignment with the Umeyama algorithm \cite{umeyama1991least}.

\paragraph{SPD} Shape prior deformation (SPD) \cite{tian2020shape} forms the basis of many subsequent methods. It is based on a categorical prior shape (represented as a point set) and predicts an offset (i.e., a deformation) for each point in the prior shape based on the observed image crop and masked 3D points. In addition, the method predicts an assignment matrix, assigning each observed point to a point in the prior shape. The assignment and deformation yield points in the NOCS \cite{wang2019normalized} which can be used to retrieve the similarity transform (i.e., 6D pose and scale) using RANSAC with the Umeyama algorithm. Approaches following this approach are also referred to as correspondence-based, since dense correspondences between points in the observation and in a reference space (i.e., NOCS) are established.

\paragraph{CR-Net} \citet{wang2021category} generally follow SPD \cite{tian2020shape} to represent the shape and estimate the pose, but modify the network architecture to iteratively deform the prior shape instead of predicting the full deformation in a single step.

\paragraph{SGPA} SGPA \cite{chen2021sgpa} also follows SPD \cite{tian2020shape} to represent the shape and estimate the pose, however, they make several architectural changes. Most notably, PointNet++ \cite{qi2017pointnet++} is used to extract features from the prior shape and input point sets, and a transformer is used to extract features to estimate the deformation and assignment matrix.

\paragraph{ASM-Net} \citet{akizuki2021} propose to represent the shape using active shape models \cite{cootes1995active}. To predict the pose, ASM-Net uses a correspondence-free estimation approach similar to CASS \cite{chen2020learning}. Compared with other works, they further use the \emph{iterative closest point} (ICP) algorithm to align their estimated shape with the observed points.

\paragraph{iCaps} \citet{deng2022icaps} proposes a pipeline for pose and shape estimation focused on tracking an object over time. It is the first method to use a coordinate-based neural field representation (DeepSDF \cite{park2019deepsdf}) in the context of categorical pose and shape estimation. The approach alternates between shape estimation given a pose and pose optimization given the shape. In this way, it iteratively optimizes the pose and shape estimates. We only evaluate iCaps' single-view performance for fair comparison.

\paragraph{SDFEst} \cite{bruns2022sdfest} describes a modular pipeline which first estimates a coarse pose and initial shape. These are subsequently refined in a render-and-compare fashion. Similar to iCaps, SDFEst uses signed distance fields to represent the shape, however, a discretized representation is used instead of a coordinate-based neural field representation. The approach is trained using only synthetic data and only uses the depth information for pose and shape estimation. Furthermore, the refinement step readily supports optimization with multiple views. However, similar to iCaps, we only evaluate the single-view performance for fair comparison.

\paragraph{DPDN} Deep prior deformation network (DPDN) \cite{lin2022category} follows SGPA \cite{chen2021sgpa} in using PointNet++ \cite{qi2017pointnet++} as their underlying feature extractor. Their network follows SPD \cite{tian2020shape} in predicting an assignment and deformation, however, pose estimation is learned instead of solved using the Umeyama algorithm. This allows them to directly supervise the pose, instead of just the assignment matrix. Furthermore, self-supervised learning is employed by augmenting the input point sets in two different ways, which enables the use of a consistency loss when no ground truth is available.

\paragraph{RBP-Pose} RBP-Pose \cite{zhang2022rbp} generally follows FS-Net \cite{chen2021fs} in directly regressing the pose in a correspondence-free manner, but extends it with shape reconstruction and various additional losses. In particular, the pose is estimated by two parallel branches: first, by a simple regression branch (like FS-Net \cite{chen2021fs}); second, by predicting the deformation and assignment (like SPD \cite{chen2021fs}) and subsequently the residual pointwise bounding box projections (hence RBP; see \cite{di2022gpv} for details on pointwise bounding box projection). The second part can be interpreted as an additional refinement step, based on the previously predicted pose and shape. This second branch is, however, only used for additional supervision during training; during inference, only the correspondence-free pose regression is used and optionally the shape deformation branch can be evaluated to retrieve the shape reconstruction.

\section{Evaluation Protocol}\label{sec:evaluation}

In this section, the existing and proposed evaluation protocol will be discussed. We will start by formally defining the problem of categorical pose and shape estimation; then discuss metrics to evaluate and compare different solutions to this problem; and finally, discuss the evaluation datasets.

\subsection{Problem Definition}\label{sec:definition}

Let $\mathbf{I}\in\mathbb{R}^{H\times W\times 3}$ be an RGB image, $\mathbf{D}\in\mathbb{R}^{H\times W}$ be a depth map, and $\mathbf{P}\in\mathbb{R}^{3\times 4}$ be the projection matrix of the associated camera. Further, let $\tensor[^i]{\mathbf{T}}{_j}$ be the homogeneous transformation matrix, that transforms a point $\tensor[^j]{\mathbf{p}}{}$ from frame $j$ to frame $i$, that is, $\tensor[^i]{\mathbf{p}}{}=\tensor[^i]{\mathbf{T}}{_j}\tensor[^j]{\mathbf{p}}{}$. Note that depending on the context, $\tensor[^i]{\mathbf{T}}{_j}$ can also be interpreted as the 6D pose of frame $j$ in frame $i$. Let $\tensor[^i]{\mathbf{R}}{_j}$ and $\tensor[^i]{\mathbf{t}}{_j}$ further denote the rotation matrix and translation vector of which $\tensor[^i]{\mathbf{T}}{_j}$ is composed.

We will use $\mathrm{o}$ to denote the object's coordinate frame and $\mathrm{c}$ to denote the camera's coordinate frame. We will use $\mathcal{O}$ to denote a 3D object and $\mathcal{B}(\mathcal{O})$ to denote the axis-aligned bounding box of $\mathcal{O}$ in $\mathcal{O}$'s frame $\mathrm{o}$. We define the origin of frame $\mathrm{o}$ as the center of $\mathcal{B}(\mathcal{O})$. We assume that 3D objects and bounding boxes are defined such that transforms can be applied to them, for example, $\tensor[^{\mathrm{c}}]{\mathcal{O}}{}=\tensor[^{\mathrm{c}}]{\mathbf{T}}{_{\mathrm{o}}}\mathcal{O}$. Following this notation, note that there is a difference between $\mathcal{B}({\tensor[^{\mathrm{c}}]{\mathcal{O}}{}})$ and $\tensor[^{\mathrm{c}}]{\mathbf{T}}{_{\mathrm{o}}}\mathcal{B}(\mathcal{O})$. The first one is an \emph{axis-aligned bounding box} (AABB), the second is an \emph{oriented bounding box} (OBB).

\begin{problem}{(Categorical Pose and Shape Estimation)}\label{problem}
    Given $(\mathbf{I},\mathbf{D},\mathbf{P})$ imaging an object $\mathcal{O}$ of known category $c$ at pose $\tensor[^{\mathrm{c}}]{\mathbf{T}}{_{\mathrm{o}}}$, and given the mask $\mathbf{M}$ of visible points of the object in the image, find estimates $\smash{\widetilde{\mathcal{O}}}$ and $\tensor[^{\mathrm{c}}]{\widetilde{\mathbf{T}}}{_{\mathrm{o}}}$ of $\mathcal{O}$ and $\tensor[^{\mathrm{c}}]{\mathbf{T}}{_{\mathrm{o}}}$, respectively.
\end{problem}

Similarly, one could define the problems of categorical pose estimation (estimate $\tensor[^{\mathrm{c}}]{\mathbf{T}}{_{\mathrm{o}}}$ only) and categorical pose and size estimation (estimate $\tensor[^{\mathrm{c}}]{\mathbf{T}}{_{\mathrm{o}}}$ and $\mathcal{B}(\mathcal{O})$). Extensions to multiple images are possible by introducing a world coordinate system, but are not further considered in this work.

\subsection{Metrics}\label{sec:metrics}

Various metrics exist to assess how well a method solves Problem \ref{problem}. Currently, the predominant evaluation metric is \emph{average precision} \cite{wang2019normalized,chen2020learning,wang2021category,chen2021sgpa,akizuki2021,lin2021dualposenet,chen2021fs,lee2021category} for pose estimation and \emph{chamfer distance} \cite{fan2017point,chen2020learning,wang2021category,chen2021sgpa,akizuki2021,lee2021category} for shape reconstruction. In this section, we will introduce these metrics and discuss several issues with them. Subsequently, we will advocate for \emph{precision} (contrary to average precision) and \emph{F-score} to evaluate pose estimation and shape reconstruction, respectively.

We first define similarity measures for transforms and objects. These are later used to define the evaluation metrics for Problem \ref{problem}.
\begin{definition}
    Let $d(\tensor[^{\mathrm{c}}]{\mathbf{T}}{_{\mathrm{o}}}, \tensor[^{\mathrm{c}}]{\widetilde{\mathbf{T}}}{_{\mathrm{o}}})$ denote the translation error between the ground-truth transform and the estimated transform, that is,
    \begin{equation}
        d(\tensor[^{\mathrm{c}}]{\mathbf{T}}{_{\mathrm{o}}},\tensor[^{\mathrm{c}}]{\widetilde{\mathbf{T}}}{_{\mathrm{o}}})=\lVert\tensor[^{\mathrm{c}}]{\mathbf{t}}{_{\mathrm{o}}} - \tensor[^{\mathrm{c}}]{\tilde{\mathbf{t}}}{_{\mathrm{o}}}\rVert_2.
    \end{equation}
\end{definition}

\begin{definition}
    Let $\delta(\tensor[^{\mathrm{c}}]{\mathbf{T}}{_{\mathrm{o}}}, \tensor[^{\mathrm{c}}]{\widetilde{\mathbf{T}}}{_{\mathrm{o}}})$ denote the rotation error between the ground-truth transform and the estimated transform, that is,
    \begin{equation}
        \delta(\tensor[^{\mathrm{c}}]{\mathbf{T}}{_{\mathrm{o}}},\tensor[^{\mathrm{c}}]{\widetilde{\mathbf{T}}}{_{\mathrm{o}}})=\left|\frac{\mathrm{trace}(\tensor[^{\mathrm{c}}]{\mathbf{R}}{_{\mathrm{o}}}\tensor*[^{\mathrm{c}}]{\widetilde{\mathbf{R}}}{_{\mathrm{o}}^{-1}})-1}{2}\right|.
    \end{equation}
\end{definition}

\begin{definition}
    Let $\mathrm{IoU}(\mathcal{B}_1,\mathcal{B}_2)$ denote the true intersection over union (IoU) of two oriented bounding boxes \cite{ahmadyan2021objectron}. Further, let the axis-aligned IoU between two objects be defined by
    \begin{equation}
        \mathrm{IoU}^+(\tensor[^{\mathrm{c}}]{\mathcal{O}}{},\tensor[^{\mathrm{c}}]{\widetilde{\mathcal{O}}}{})=\mathrm{IoU}\left(\mathcal{B}({\tensor[^{\mathrm{c}}]{\mathcal{O}}{}}),\mathcal{B}({\tensor[^{\mathrm{c}}]{\widetilde{\mathcal{O}}}{}})\right),
    \end{equation}
    and the true (oriented) IoU using oriented bounding boxes by
    \begin{equation}
        \mathrm{IoU}(\tensor[^{\mathrm{c}}]{\mathcal{O}}{},\tensor[^{\mathrm{c}}]{\widetilde{\mathcal{O}}}{})=\mathrm{IoU}\left(\tensor[^{\mathrm{c}}]{\mathbf{T}}{_{\mathrm{o}}}\mathcal{B}(\mathcal{O}),\tensor[^{\mathrm{c}}]{\widetilde{\mathbf{T}}}{_{\mathrm{o}}}\mathcal{B}(\widetilde{\mathcal{O}})\right).
    \end{equation}
\end{definition}

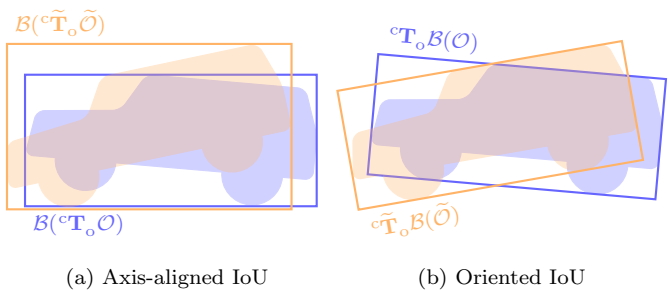
\begin{figure}[htb]
    \centering
    \begin{subfigure}[b]{0.24\textwidth}
        \centering
        \scriptsize
\begin{tikzpicture}[scale=1]
    \coordinate (virtual) at (3, 1.1);
    \coordinate (object) at (3.25, 0.9);
    \coordinate (grid) at (2.5, 0.5);

    \begin{scope}[blue!60,rotate=-5,shift={(object)},local bounding box=object]
        \begin{scope}[transparency group,opacity=0.3]
            \draw[fill] (0.8,0) circle (0.4);
            \draw[fill] (-1.4,0) circle (0.4);
            \draw[fill, rounded corners=3pt] (-2.2, 0) -- (-2.1, 0.6) -- (-1.0,0.7) -- (-0.7, 1.2) -- (1.4, 1.2) -- (1.6, 0.6) -- (1.6, 0) -- cycle;
        \end{scope}
    \end{scope}
    \draw[blue!60,thick] (object.north east) rectangle (object.south west);
    \node[anchor=north west, blue!60] (olabel) at (object.south west)  {$\mathcal{B}(\tensor[^{\mathrm{c}}]{\mathbf{T}}{_{\mathrm{o}}}\mathcal{O})$};
    
    \begin{scope}[orange!60,rotate=12,shift={(virtual)},local bounding box=virtual]
        \begin{scope}[transparency group,opacity=0.3]
            \draw[fill] (0.8,0) circle (0.4);
            \draw[fill] (-1.4,0) circle (0.4);
            \draw[fill, rounded corners=3pt] (-2.2, 0) -- (-2.1, 0.6) -- (-1.0,0.7) -- (-0.7, 1.2) -- (1.4, 1.2) -- (1.6, 0.6) -- (1.6, 0) -- cycle;
        \end{scope}
    \end{scope}
    \draw[orange!60,thick] (virtual.north east) rectangle (virtual.south west);
    \node[anchor=south west, orange!60] (vlabel) at (virtual.north west) {$\mathcal{B}(\tensor[^{\mathrm{c}}]{\widetilde{\mathbf{T}}}{_{\mathrm{o}}}\widetilde{\mathcal{O}})$};
\end{tikzpicture}
        \vspace{0.1cm}
        \subcaption{Axis-aligned IoU}
    \end{subfigure}%
    \begin{subfigure}[b]{0.24\textwidth}
        \centering
        \scriptsize
\begin{tikzpicture}[scale=1]
    \coordinate (virtual) at (3, 1);
    \coordinate (object) at (3.25, 0.9);
    \coordinate (grid) at (2.5, 0.5);

    \begin{scope}[blue!60,rotate=-5,shift={(object)}]
        \node[transform shape,anchor=south west, blue!60] (olabel) at (-2.2,1.2) {$\tensor[^{\mathrm{c}}]{\mathbf{T}}{_{\mathrm{o}}}\mathcal{B}(\mathcal{O})$};
        \begin{scope}[transparency group,opacity=0.3,local bounding box=object]
            \draw[fill] (0.8,0) circle (0.4);
            \draw[fill] (-1.4,0) circle (0.4);
            \draw[fill, rounded corners=3pt] (-2.2, 0) -- (-2.1, 0.6) -- (-1.0,0.7) -- (-0.7, 1.2) -- (1.4, 1.2) -- (1.6, 0.6) -- (1.6, 0) -- cycle;
        \end{scope}
        \draw[thick] (-2.2,-0.4) rectangle (1.6,1.2);
    \end{scope}
    
    \begin{scope}[orange!60,rotate=10,shift={(virtual)}]
        \node[transform shape,anchor=north west, orange!60] (vlabel) at (-2.2,-0.4) {$\tensor[^{\mathrm{c}}]{\widetilde{\mathbf{T}}}{_{\mathrm{o}}}\mathcal{B}(\widetilde{\mathcal{O}})$};
        \begin{scope}[transparency group,opacity=0.3,local bounding box=virtual]
            \draw[fill] (0.8,0) circle (0.4);
            \draw[fill] (-1.4,0) circle (0.4);
            \draw[fill, rounded corners=3pt] (-2.2, 0) -- (-2.1, 0.6) -- (-1.0,0.7) -- (-0.7, 1.2) -- (1.4, 1.2) -- (1.6, 0.6) -- (1.6, 0) -- cycle;
        \end{scope}
        \draw[thick] (-2.2,-0.4) rectangle (1.6,1.2);
    \end{scope}
\end{tikzpicture}
        \subcaption{Oriented IoU}
    \end{subfigure}%
    \caption{Visualization of axis-aligned IoU and true (oriented) IoU in 2D. Note that axis-aligned IoU is less accurate and depends on the object shape.}
    \label{fig:iou}
\end{figure}

The difference between these two IoU definitions is visualized in Figure \ref{fig:iou}. The current evaluation protocol \cite{wang2019normalized} uses axis-aligned $\mathrm{IoU}^+$ instead of oriented $\mathrm{IoU}$, although the former is less accurate. Our implementation follows \cite{ahmadyan2021objectron} and computes oriented $\mathrm{IoU}$.

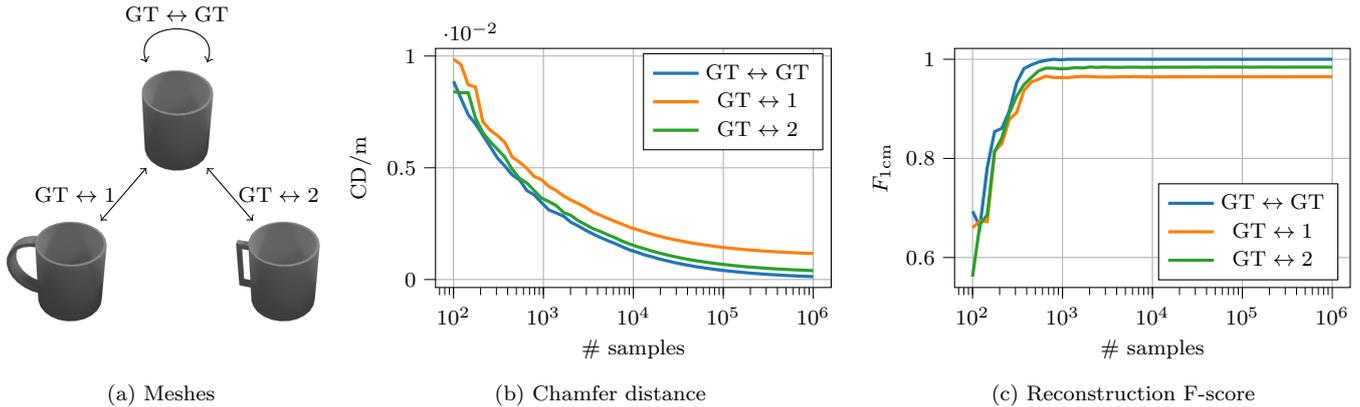
\begin{figure*}
    \centering
    \scriptsize
    \begin{subfigure}[b]{0.25\textwidth}
        \centering
        \begin{tikzpicture}
            \node (1) at (0,0) {\includegraphics[width=1.8cm]{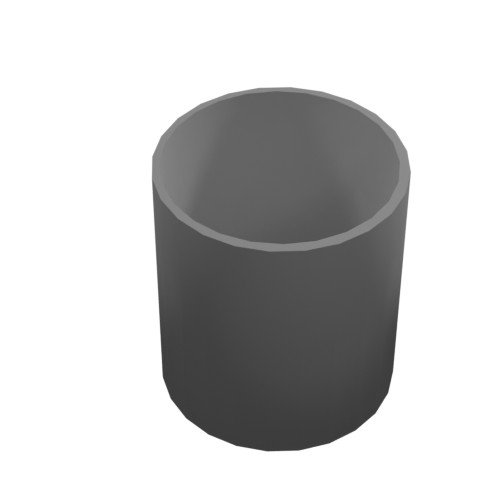}};
            \node (2) at (-1.4,-2) {\includegraphics[width=1.8cm]{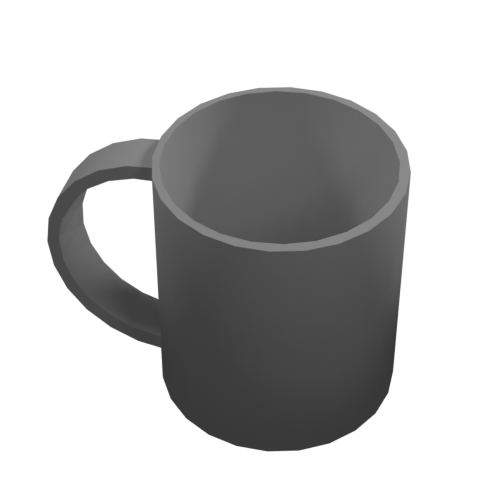}};
            \node (3) at (1.4,-2) {\includegraphics[width=1.8cm]{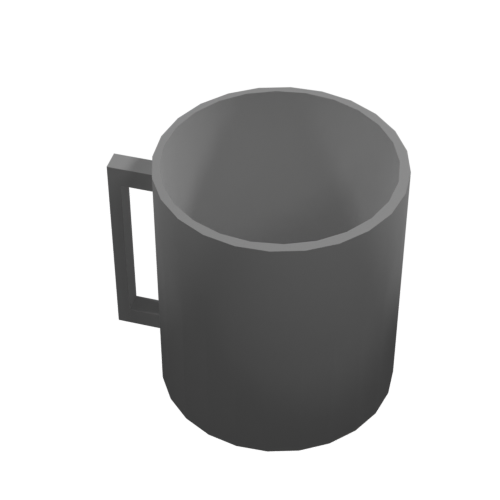}};
            \draw[<->] (-0.3,0.7) to [out=110,in=70,looseness=2]  node[midway, above] {$\mathrm{GT}\leftrightarrow\mathrm{GT}$} (0.5,0.7);
            \draw[<->] (-0.3,-0.7) -- (-0.9,-1.4) node[midway,left] {$\mathrm{GT}\leftrightarrow\mathrm{1}$};
            \draw[<->] (0.5,-0.7) -- (1.1,-1.4) node[midway,right] {$\mathrm{GT}\leftrightarrow\mathrm{2}$};
        \end{tikzpicture}
        \vspace{0cm}
        \subcaption{Meshes}\label{fig:example_meshes}
    \end{subfigure}
    \begin{subfigure}[b]{0.37\textwidth}
        \begin{tikzpicture}

\definecolor{darkgray176}{RGB}{176,176,176}
\definecolor{darkorange25512714}{RGB}{255,127,14}
\definecolor{forestgreen4416044}{RGB}{44,160,44}
\definecolor{steelblue31119180}{RGB}{31,119,180}

\begin{axis}[
width=\textwidth,
height=0.7\textwidth,
log basis x={10},
tick align=outside,
ylabel near ticks,
tick pos=left,
x grid style={darkgray176},
xlabel={\# samples},
xmajorgrids,
xmin=63.0957344480193, xmax=1584893.19246111,
xmode=log,
xtick style={color=black},
y grid style={darkgray176},
ylabel={$\mathrm{CD} / \mathrm{m}$},
ymajorgrids,
ymin=-0.000357484703618492, ymax=0.0103406887394132,
ytick style={color=black}
]
\addplot [very thick, steelblue31119180]
table {%
100 0.00887207639653745
120.679264063933 0.00811490515495286
145.634847750124 0.00737640696402916
175.751062485479 0.00695920815044473
212.095088792019 0.00645543000659581
255.954792269954 0.0059451813218451
308.884359647748 0.00543289692207538
372.759372031494 0.00507654931112869
449.843266896944 0.00468115914739195
542.867543932386 0.00445268615741683
655.128556859551 0.00398067513402519
790.60432109077 0.0037874671856085
954.095476349994 0.00343718972446623
1151.39539932645 0.0031133275135468
1389.49549437314 0.00297498601622347
1676.83293681101 0.00283442254715392
2023.58964772516 0.00256129121685061
2442.05309454865 0.00239621792731292
2947.05170255181 0.00220447533126299
3556.48030622313 0.00203953458271211
4291.93426012878 0.00188323481923328
5179.47467923121 0.00174113533545296
6250.55192527397 0.00161242185762149
7543.12006335462 0.0014613526413073
9102.98177991522 0.00132308859683913
10985.4114198756 0.0012215272464772
13257.1136559011 0.00110455593555721
15998.5871960606 0.00101882455580813
19306.9772888325 0.000921454244140444
23299.5181051537 0.000843655138763264
28117.6869797423 0.000765090979120931
33932.2177189533 0.000697895210917357
40949.1506238042 0.000636828998799428
49417.1336132384 0.000577418015645057
59636.2331659464 0.000527715202747646
71968.5673001151 0.000479635930448023
86851.1373751352 0.000434957964785117
104811.313415469 0.000397714997977767
126485.52168553 0.000361722779532943
152641.796717523 0.000329396955295915
184206.996932672 0.000299906837432821
222299.648252619 0.000273720327817266
268269.579527973 0.000249049251272778
323745.754281764 0.000226618350209181
390693.993705461 0.00020587969481896
471486.636345739 0.000187550025506476
568986.602901829 0.000171009185360244
686648.8450043 0.000155394285924479
828642.772854684 0.000141521151359487
1000000 0.000128795907428403
};
\addplot [very thick, darkorange25512714]
table {%
100 0.0098544081283663
120.679264063933 0.00962652215559299
145.634847750124 0.00871739929352431
175.751062485479 0.00861149963842358
212.095088792019 0.00705867486833976
255.954792269954 0.00668049429555589
308.884359647748 0.00644181601503226
372.759372031494 0.00613072822287869
449.843266896944 0.00548191864150449
542.867543932386 0.00525096216375182
655.128556859551 0.00497317586246658
790.60432109077 0.00460763199779269
954.095476349994 0.00446576338699611
1151.39539932645 0.00416249074950407
1389.49549437314 0.00398510145224972
1676.83293681101 0.00374062820383371
2023.58964772516 0.00355717170772655
2442.05309454865 0.00340009360767452
2947.05170255181 0.00322651918997893
3556.48030622313 0.00301841729856438
4291.93426012878 0.00287739274449161
5179.47467923121 0.00273454560434042
6250.55192527397 0.00260538875676169
7543.12006335462 0.00247338948015978
9102.98177991522 0.00233461605278349
10985.4114198756 0.00223530355250698
13257.1136559011 0.00213103868571683
15998.5871960606 0.00202804833443362
19306.9772888325 0.00194276518539997
23299.5181051537 0.00186040937090073
28117.6869797423 0.00178794026856257
33932.2177189533 0.00172376734160445
40949.1506238042 0.00166123787363394
49417.1336132384 0.00160529434551679
59636.2331659464 0.00155685082883245
71968.5673001151 0.00150897009930137
86851.1373751352 0.00146569545489431
104811.313415469 0.00142658055001089
126485.52168553 0.00139234505520267
152641.796717523 0.00135945238207842
184206.996932672 0.00133038133924684
222299.648252619 0.00130538776142812
268269.579527973 0.00128001706057
323745.754281764 0.00125953347342331
390693.993705461 0.00123845451696843
471486.636345739 0.00122125940061886
568986.602901829 0.00120481098984016
686648.8450043 0.00118910474316047
828642.772854684 0.00117534673574433
1000000 0.00116292540672353
};
\addplot [very thick, forestgreen4416044]
table {%
100 0.00839988869119809
120.679264063933 0.00835949652358131
145.634847750124 0.00835434029461695
175.751062485479 0.00721793222931058
212.095088792019 0.00656990781298667
255.954792269954 0.00617076557924616
308.884359647748 0.0058276374068993
372.759372031494 0.00547335196372112
449.843266896944 0.00492880321600788
542.867543932386 0.00452648536686067
655.128556859551 0.00431980578638745
790.60432109077 0.00397668861503106
954.095476349994 0.00363937041418539
1151.39539932645 0.00348773195750739
1389.49549437314 0.00330799232536578
1676.83293681101 0.00299811414573868
2023.58964772516 0.002870135104484
2442.05309454865 0.00262952435943032
2947.05170255181 0.00245728501768351
3556.48030622313 0.00228008169161143
4291.93426012878 0.00215124842140402
5179.47467923121 0.00200537545972758
6250.55192527397 0.0018698734703572
7543.12006335462 0.00171509240465301
9102.98177991522 0.00159105084932836
10985.4114198756 0.00147416182514654
13257.1136559011 0.00137870346986043
15998.5871960606 0.00127567259806342
19306.9772888325 0.00119081631096961
23299.5181051537 0.00110525153643973
28117.6869797423 0.00103052821670086
33932.2177189533 0.000964132627335628
40949.1506238042 0.000899538792831904
49417.1336132384 0.000843889997326636
59636.2331659464 0.000795266772089992
71968.5673001151 0.000745811491478122
86851.1373751352 0.00070509139105524
104811.313415469 0.000665281446276757
126485.52168553 0.000628973294919481
152641.796717523 0.000597832743348116
184206.996932672 0.000568245935881892
222299.648252619 0.000542283444245131
268269.579527973 0.000517794789822584
323745.754281764 0.000494619492178697
390693.993705461 0.000474928658104126
471486.636345739 0.000456973967035564
568986.602901829 0.000440197083590331
686648.8450043 0.000425171137549512
828642.772854684 0.00041081125649559
1000000 0.000398668872015543
};
\legend{$\mathrm{GT}\leftrightarrow\mathrm{GT}$,$\mathrm{GT}\leftrightarrow\mathrm{1}$,$\mathrm{GT}\leftrightarrow\mathrm{2}$}
\end{axis}

\end{tikzpicture}
        \subcaption{Chamfer distance}\label{fig:cd}
    \end{subfigure}
    \begin{subfigure}[b]{0.37\textwidth}
        \begin{tikzpicture}

\definecolor{darkgray176}{RGB}{176,176,176}
\definecolor{darkorange25512714}{RGB}{255,127,14}
\definecolor{forestgreen4416044}{RGB}{44,160,44}
\definecolor{steelblue31119180}{RGB}{31,119,180}

\begin{axis}[
width=\textwidth,
height=0.7\textwidth,
log basis x={10},
tick align=outside,
tick pos=left,
x grid style={darkgray176},
legend pos=south east,
xlabel={\# samples},
xmajorgrids,
xmin=63.0957344480193, xmax=1584893.19246111,
xmode=log,
xtick style={color=black},
y grid style={darkgray176},
ylabel={$F_{1\mathrm{cm}}$},
ymajorgrids,
ymin=0.539486725663717, ymax=1.02192920353982,
ytick style={color=black}
]
\addplot [very thick, steelblue31119180]
table {%
100 0.693237410071943
120.679264063933 0.662264150943396
145.634847750124 0.781528178641957
175.751062485479 0.854046822742475
212.095088792019 0.860687516154045
255.954792269954 0.895559274037843
308.884359647748 0.952454700325228
372.759372031494 0.981175430844012
449.843266896944 0.988864142538975
542.867543932386 0.99446152160965
655.128556859551 0.997709339609969
790.60432109077 1
954.095476349994 0.998950682056663
1151.39539932645 1
1389.49549437314 1
1676.83293681101 1
2023.58964772516 1
2442.05309454865 1
2947.05170255181 1
3556.48030622313 1
4291.93426012878 1
5179.47467923121 1
6250.55192527397 1
7543.12006335462 1
9102.98177991522 1
10985.4114198756 1
13257.1136559011 1
15998.5871960606 1
19306.9772888325 1
23299.5181051537 1
28117.6869797423 1
33932.2177189533 1
40949.1506238042 1
49417.1336132384 1
59636.2331659464 1
71968.5673001151 1
86851.1373751352 1
104811.313415469 1
126485.52168553 1
152641.796717523 1
184206.996932672 1
222299.648252619 1
268269.579527973 1
323745.754281764 1
390693.993705461 1
471486.636345739 1
568986.602901829 1
686648.8450043 1
828642.772854684 1
1000000 1
};
\addplot [very thick, darkorange25512714]
table {%
100 0.659848484848485
120.679264063933 0.674074074074074
145.634847750124 0.672254641909814
175.751062485479 0.814195488721804
212.095088792019 0.830161878216123
255.954792269954 0.878815668806498
308.884359647748 0.892278630460449
372.759372031494 0.93706288319931
449.843266896944 0.954127535419295
542.867543932386 0.959976202449165
655.128556859551 0.966061562746646
790.60432109077 0.963254593175853
954.095476349994 0.963043478260869
1151.39539932645 0.963063063063063
1389.49549437314 0.964591874767052
1676.83293681101 0.965432098765432
2023.58964772516 0.965217391304348
2442.05309454865 0.964376590330789
2947.05170255181 0.96414762741652
3556.48030622313 0.964322120285423
4291.93426012878 0.964651948365303
5179.47467923121 0.964614154338265
6250.55192527397 0.965060440470277
7543.12006335462 0.964797913950456
9102.98177991522 0.964387302309705
10985.4114198756 0.964558393816571
13257.1136559011 0.96441823223841
15998.5871960606 0.964631265572922
19306.9772888325 0.964631432172258
23299.5181051537 0.964603932896345
28117.6869797423 0.964649998158854
33932.2177189533 0.964856619890177
40949.1506238042 0.964744845719198
49417.1336132384 0.96468788954883
59636.2331659464 0.9646887260859
71968.5673001151 0.964720098395322
86851.1373751352 0.964703778757897
104811.313415469 0.964747831927461
126485.52168553 0.964714259989851
152641.796717523 0.964770272607352
184206.996932672 0.964652056857973
222299.648252619 0.964734208399988
268269.579527973 0.96469603004002
323745.754281764 0.964691849160868
390693.993705461 0.964704728869409
471486.636345739 0.964737449401824
568986.602901829 0.964695992510692
686648.8450043 0.964719859901681
828642.772854684 0.964722256406284
1000000 0.964758837983205
};
\addplot [very thick, forestgreen4416044]
table {%
100 0.56141592920354
120.679264063933 0.666666666666667
145.634847750124 0.686275862068966
175.751062485479 0.814195488721804
212.095088792019 0.841657417684055
255.954792269954 0.891187244128421
308.884359647748 0.925279106858054
372.759372031494 0.948856194218527
449.843266896944 0.963188588642713
542.867543932386 0.976622797389447
655.128556859551 0.982128982128982
790.60432109077 0.982002805689678
954.095476349994 0.980808426956226
1151.39539932645 0.98141592920354
1389.49549437314 0.98316251830161
1676.83293681101 0.982392228293868
2023.58964772516 0.984437751004016
2442.05309454865 0.983558792924037
2947.05170255181 0.984318455971049
3556.48030622313 0.983709631323235
4291.93426012878 0.983657034580767
5179.47467923121 0.983912105159898
6250.55192527397 0.98407022106632
7543.12006335462 0.983765577635567
9102.98177991522 0.983868266815518
10985.4114198756 0.983764281419122
13257.1136559011 0.984099007624813
15998.5871960606 0.984060455959865
19306.9772888325 0.984081879653748
23299.5181051537 0.984018314618991
28117.6869797423 0.984230776178176
33932.2177189533 0.98419350975931
40949.1506238042 0.983996625645097
49417.1336132384 0.984223887193085
59636.2331659464 0.98418430353873
71968.5673001151 0.984170001340927
86851.1373751352 0.984133482270789
104811.313415469 0.984108983585583
126485.52168553 0.984056224899599
152641.796717523 0.984100045257301
184206.996932672 0.984251447477254
222299.648252619 0.984140548420338
268269.579527973 0.984077753730623
323745.754281764 0.984150941028071
390693.993705461 0.984170772610861
471486.636345739 0.984098460265543
568986.602901829 0.984168528623854
686648.8450043 0.984157605821231
828642.772854684 0.984160025302769
1000000 0.984135240043763
};
\legend{$\mathrm{GT}\leftrightarrow\mathrm{GT}$,$\mathrm{GT}\leftrightarrow\mathrm{1}$,$\mathrm{GT}\leftrightarrow\mathrm{2}$}
\end{axis}

\end{tikzpicture}
        \subcaption{Reconstruction F-score}\label{fig:fscore}
    \end{subfigure}
    \caption{Visualization of the effect of varying number of samples on the chamfer distance (b) and reconstruction F-score (c). We consider two reconstructions (denoted by 1 and 2) of the ground truth (denoted by GT). Note that particularly the relative difference between $\mathrm{CD}(\mathcal{S}_{\mathrm{GT}}, \mathcal{S}_1)$ and $\mathrm{CD}(\mathcal{S}_{\mathrm{GT}}, \mathcal{S}_2)$ varies significantly. This is because the majority of the error stems from sparse sampling, not from actual differences in geometry. All mugs have been scaled to be \SI{10}{cm} tall.}
\end{figure*}

\subsubsection{Chamfer Distance}\label{sec:cd}
In the context of shape reconstruction, \emph{chamfer distance} (CD) was introduced by \cite{fan2017point} to differentiably measure the difference of point sets. 
\begin{definition}
    Let $\mathcal{S}\subset \mathbb{R}^3$ and $\widetilde{\mathcal{S}}\subset\mathbb{R}^3$ denote point sets sampled from the surfaces of $\mathcal{O}$ and $\widetilde{\mathcal{O}}$, respectively. We define CD as
    \begin{equation}
        \mathrm{CD}(\mathcal{S},\widetilde{\mathcal{S}})=\frac{1}{2}\mathrm{AD}(\mathcal{S}\rightarrow\widetilde{\mathcal{S}}) + \frac{1}{2}\mathrm{AD}(\widetilde{\mathcal{S}}\rightarrow\mathcal{S})
    \end{equation}
    based on the directed average distance
    \begin{equation}
        \mathrm{AD}(\mathcal{X}\rightarrow\mathcal{Y})=\frac{1}{|\mathcal{X}|}\sum_{\mathbf{x}\in \mathcal{X}}\min_{\mathbf{y}\in\widetilde{\mathcal{Y}}}\lVert \mathbf{x} - \mathbf{y} \rVert_2.\label{eq:averagedistance}
    \end{equation}
\end{definition}
It is easiest to interpret as the mean Euclidean distance from a point in one point set to the closest point in the other set. Note that slightly different CD versions exist, such as squared versions and ones using the sum instead of the arithmetic mean. 

In \cref{fig:cd} we visualize potential issues with CD as an evaluation metric. Consider the two mugs with different handles (denoted 1 and 2) as reconstructions of the mug without handle (denoted GT). The relative quality of these reconstructions measured by CD varies significantly depending on the number of samples. Furthermore, a large number of samples is required for CD to converge.

Notably, the number of ground-truth samples is left unspecified by most methods. This problem becomes further amplified because methods perform reconstruction by predicting a varying number of samples as noted by \cite{akizuki2021}. Furthermore, \citet{tatarchenko2019single} note that chamfer distance is not robust to outliers, since it is an average of distances (i.e., a single outlier can strongly influence the metric). Therefore, we discourage further use of CD for evaluation purposes.

\citet{manhardt2020cps++} introduce a size-normalized reconstruction metric, closely related to CD\footnote{\citet{manhardt2020cps++} introduced this metric as \emph{average distance of predicted point sets} in an already thresholded manner. We have changed it to a non-thresholded metric, which can be thresholded later, yielding the same result as the originally proposed metric. This fits better into our formalism.}.
\begin{definition}
    Let $\mathcal{S}\subset \mathbb{R}^3$ and $\widetilde{\mathcal{S}}\subset\mathbb{R}^3$ denote point sets sampled from the surfaces of $\mathcal{O}$ and $\widetilde{\mathcal{O}}$, respectively. We define normalized average distance (NAD) as 
    \begin{equation}
        \mathrm{NAD}(\mathcal{S},\widetilde{\mathcal{S}})=\max\left(\frac{\mathrm{AD}(\mathcal{S}\rightarrow \widetilde{\mathcal{S}})}{\mathrm{diam}(\mathcal{S})}, \frac{\mathrm{AD}(\widetilde{\mathcal{S}} \rightarrow \mathcal{S})}{\mathrm{diam}(\widetilde{\mathcal{S}})}\right)
    \end{equation}
    based on the directed average distance \eqref{eq:averagedistance} and the diameter of the point sets
    \begin{equation}
        \mathrm{diam}(\mathcal{X})=\max_{\mathbf{x}\in\mathcal{X}}\max_{\mathbf{y}\in\mathcal{X}}\lVert \mathbf{x}-\mathbf{y} \rVert_2.
    \end{equation}
\end{definition}
In general, this metric exhibits similar pitfalls as chamfer distance, however, normalizing metrics (or, equivalently, setting thresholds) based on the diameter can be a useful tool for datasets with large size differences between objects.

\subsubsection{Reconstruction F-Score}\label{sec:fscore}

\citet{tatarchenko2019single} introduced F-score in the context of reconstruction as a robust alternative to chamfer distance.
\begin{definition}
    Let $\mathcal{S}\subset \mathbb{R}^3$ and $\widetilde{\mathcal{S}}\subset\mathbb{R}^3$ denote point sets sampled from the surfaces of $\mathcal{O}$ and $\widetilde{\mathcal{O}}$, respectively.
    Given an application-specific threshold $\Delta$, we define reconstruction recall as
    \begin{equation}
        r_\Delta(\mathcal{S},\widetilde{\mathcal{S}}) = \frac{1}{|\mathcal{S}|}\sum_{\mathbf{x}\in \mathcal{S}}\left[\min_{\mathbf{y}\in\widetilde{\mathcal{S}}}\lVert \mathbf{x} - \mathbf{y} \rVert_2 < \Delta\right]
    \end{equation}
    and reconstruction precision as
    \begin{equation}
        p_\Delta(\mathcal{S},\widetilde{\mathcal{S}}) = \frac{1}{|\widetilde{\mathcal{S}}|}\sum_{\mathbf{y}\in \widetilde{\mathcal{S}}}\left[\min_{\mathbf{x}\in\mathcal{S}}\lVert \mathbf{x} - \mathbf{y} \rVert_2 < \Delta\right],
    \end{equation}
    where $\left[\cdot\right]$ denotes the Iverson bracket. Finally, we define F-score as the harmonic mean of precision and recall
    \begin{equation}
        F_\Delta = \frac{2}{p_\Delta(\mathcal{S},\widetilde{\mathcal{S}}^{-1})+r_\Delta(\mathcal{S},\widetilde{\mathcal{S}})^{-1}}.
    \end{equation}
\end{definition}

Note that $\Delta$ should be adjusted depending on the application and the sensor. For the table-top items contained in the datasets, we propose to use $\Delta=\SI{1}{cm}$.

In \cref{fig:fscore} we show $F_{1\mathrm{cm}}$ for varying numbers of samples for the meshes in \cref{fig:example_meshes}. $F_{1\mathrm{cm}}$ converges significantly faster than CD and can easily be interpreted as the percentage of correct (i.e., error below $\Delta=\SI{1}{cm}$) surfaces or points \cite{tatarchenko2019single}.

Due to these reasons, following \cite{tatarchenko2019single}, we advocate for using F-score instead of CD to evaluate shape reconstruction. Furthermore, note that shape reconstruction can be evaluated in the object frame or in the camera frame, taking into account $\tensor*[^{\mathrm{c}}]{\mathbf{T}}{_{\mathrm{o}}}$ and $\tensor*[^{\mathrm{c}}]{\widetilde{\mathbf{T}}}{_{\mathrm{o}}}$. Previous works evaluated in the object frame (i.e., assuming perfect alignment based on the canonical reference frame); however, we believe it is better to evaluate posed reconstruction in the camera frame, as it correlates more directly with downstream usability of the full estimate.

All metrics so far (i.e., $d$, $\delta$, $\mathrm{IoU}$, $\mathrm{CD}$, $\mathrm{NAD}$, and $F_\Delta$) assess the quality of a single estimate. Next, we discuss average precision and precision, which attempt to summarize a method's performance on a dataset. In principle, one could also compute different averages of the aforementioned metrics, but those are typically affected by outliers and harder to interpret in comparison to thresholded evaluation metrics that classify estimates as correct or incorrect.

\subsubsection{Average Precision}\label{sec:ap}

\emph{Average precision} (AP) summarizes precision-recall curves in a single value \cite{salton1983introduction} and has been the standard evaluation metric for object detection on the PASCAL VOC \cite{everingham2010pascal} and COCO datasets \cite{lin2014microsoft}. In general, average precision is calculated based on the interpolated precision-recall curve, which is constructed by varying a confidence threshold. 

\citet{wang2019normalized} proposed to use AP with different thresholds on $\mathrm{IoU}^+$, $d$, and $\delta$ to evaluate their pose estimation (all specified thresholds must hold for a prediction to count as a true positive). Their method includes a Mask R-CNN architecture to detect objects and therefore had a confidence threshold to compute AP. However, most of the following pose and shape estimation methods do not include such a confidence threshold\footnote{Only methods marked \textbf{Det.}\ in \cref{tab:overview} include a confidence threshold.}. Instead they assume $\mathbf{M}$ to be given as stated in \cref{problem}. 

To still follow the same evaluation protocol as \cite{wang2019normalized}, all other methods rely on the same, suboptimal Mask R-CNN predictions that \cite{wang2019normalized} provided. This protocol effectively limits the achievable AP due to wrong classifications, missing detections and poor masks. Furthermore, AP is inherently difficult to interpret compared to simpler metrics.

Therefore, we believe that AP is ill-suited to compare pose and shape estimation methods that do not include detection. Rather, simpler metrics, such as precision (see below), should be used, assuming that mask $\mathbf{M}$ and category $c$ are provided.

\subsubsection{Precision}\label{sec:precision} We propose to use precision, contrary to average precision, to assess categorical pose and shape estimation.
\begin{definition}
    Given inputs $(\mathbf{I}^i,\mathbf{D}^i,\mathbf{P}^i,\mathbf{M}^i,c^i)$, ground truths $(\mathcal{O}^i,\tensor*[^{\mathrm{c}}]{\mathbf{T}}{^i_{\mathrm{o}}})$, and associated predictions $(\widetilde{\mathcal{O}}^i,\tensor*[^{\mathrm{c}}]{\widetilde{\mathbf{T}}}{^i_{\mathrm{o}}})$ with $i=1,...,N$, let precision be defined as
    \begin{equation}
        P=\frac{\sum_{i=1}^N\left[ \mathrm{c}(\mathcal{O}^i,\tensor*[^{\mathrm{c}}]{\mathbf{T}}{^i_{\mathrm{o}}}, \widetilde{\mathcal{O}}^i,\tensor*[^{\mathrm{c}}]{\widetilde{\mathbf{T}}}{^i_{\mathrm{o}}}) \right]}{N},
    \end{equation}
    where $\left[\cdot\right]$ denotes the Iverson bracket and $\mathrm{c}$ determines whether a prediction is correct or not based on a single or multiple thresholds on $d$, $\delta$, $\mathrm{IoU}$, $\mathrm{CD}$, $\mathrm{NAD}$, and $F_\Delta$.
\end{definition}
Precision measures the percentage of correct estimates based on thresholds on translation error, rotation error, IoU, NAD and F-score. Note that we can use this simpler metric instead of average precision because we decouple pose estimation from detection and classification.

\begin{table*}[htb]
    \centering
    \scriptsize
    \setlength{\tabcolsep}{3pt}
    \renewcommand{\arraystretch}{1.3}
    \begin{threeparttable}
        \caption{Overview of categorical pose and shape estimation datasets. All datasets include RGB-D data and categorical pose ground truth. \textbf{Bold} datasets are included in our toolbox and evaluation. See text for further explanation and discussion.}\label{tab:datasets}
        \begin{tabular}{lccccccm{9.5cm}}
            \toprule
            \bfseries Dataset & \rot{\bfseries \# Instances} & \rot{\bfseries \# Sequences} & \rot{\bfseries  \# Categories} & \rot{\bfseries Meshes} & \rot{\bfseries Real} & \rot{\bfseries Train} & \multicolumn{1}{l}{\rot{\bfseries  Notes}} \\
            \midrule
            CAMERA25 \cite{wang2019normalized} & 1085 & 31$\mathrlap{\smash{^\ast}}$ & 6 & \cmark & \xmark & \cmark & Synthetic objects on table \\
            \rowcolor{LightGray}
            \textbf{REAL275} \cite{wang2019normalized} & 36 & 13 & 6 & \cmark & \cmark & \cmark & Sequences of multiple objects on table \\
            TOD \cite{liu2020keypose} & 15 & 600 & 3 & \cmark & \cmark & \cmark & Transparent objects on plane \\
            \rowcolor{LightGray}
            PhoCaL \cite{wang2022phocal} & 60 & 24 & 8 & \cmark & \cmark & \cmark & Robotic ground-truth annotations; polarization data; transparent and reflective objects \\
            Wild6D \cite{fu2022wild6d} & 162 & 486 & 5 & \xmark & \cmark & \omark$\smash{^\dagger}$ & Dataset for self-supervised learning \\
            \rowcolor{LightGray}
            \textbf{REDWOOD75} & 15 & 15 & 3 & \cmark & \cmark & \xmark & Annotations for Redwood dataset \cite{choi2016large}; objects freely rotated in hands \\
            HouseCat6D \cite{jung2022housecat} & 194 & 41 & 10 & \cmark & \cmark & \cmark & Larger dataset following PhoCaL \cite{wang2022phocal}; better viewpoint coverage \\
            \bottomrule
        \end{tabular}
        \begin{tablenotes}
            \item[$^\ast$] 31 different background scenes, 300K different object arrangements
            \item[$^\dagger$] Statistics for evaluation split with ground-truth poses; larger training split only has mask annotations
        \end{tablenotes}
    \end{threeparttable}
\end{table*}

\subsubsection{Summary}\label{sec:metrics_summary}

We propose to evaluate categorical pose and shape estimation methods by calculating precision at a few informative thresholds of $d$, $\delta$ and $F_\mathrm{\Delta}$. Furthermore, this evaluation procedure can be adjusted for categorical pose estimation by using only $d$ and $\delta$ and for categorical pose and size estimation by using $\mathrm{IoU}$ instead of $F_\mathrm{\Delta}$. The thresholds for these metrics must be adjusted based on application requirements, sensor accuracy, and annotation quality. Furthermore, when combining multiple thresholds, care should be taken that they are roughly equally strict.

Note that when evaluating $\delta$ and $\mathrm{IoU}$, extra care must be taken with respect to the categories containing symmetric objects. We follow \cite{wang2019normalized} and ignore rotations around the up-axis for the bottle, bowl, and can categories. Further issues with ambiguities are discussed in Section \ref{sec:limitations}.

\subsection{Datasets}\label{sec:datasets}

An overview of existing datasets is given in Table \ref{tab:datasets}. So far, most methods have been evaluated on the synthetic CAMERA25 dataset and on the smaller real-world dataset REAL275 \cite{wang2019normalized}. The transparent object dataset (TOD) \cite{liu2020keypose} focusses on transparent objects and provides the real erroneous depth data and depth data captured using opaque twins of the transparent objects. The PhoCaL dataset \cite{wang2022phocal} is a larger, highly accurate dataset. Similar to the REAL275 dataset, most objects are upright on a table (see \cref{sec:real275}). However, it contains significantly more occlusions, and some objects in unconstrained poses. The recently released HouseCat6D \cite{jung2022housecat} is a follow-up dataset, that significantly improves with regards to limited viewpoints. Finally, the Wild6D dataset \cite{fu2022wild6d} is a large-scale dataset for self-supervised learning. However, it does not contain ground-truth shapes, and is therefore not suited to assess shape estimation methods.

Given its popularity in the literature, we evaluate on the REAL275 dataset (see Section \ref{sec:real275}) and additionally on the REDWOOD75 dataset (see Section \ref{sec:redwood}). All of REDWOOD75's categories are included in the REAL275 dataset, which simplifies evaluation across datasets. While not included in this evaluation, we note that PhoCaL, TOD, and especially HouseCat6D are promising datasets for future research on categorical pose and shape estimation.

\subsubsection{REAL275}\label{sec:real275}

The REAL dataset was proposed by \citet{wang2019normalized} and consists of 4300 training images (7 video sequences) and 2750 test images (6 video sequences). An additional validation set containing 950 images is mentioned in the publication, but has not been made publicly available.
The dataset contains 6 categories (bottle, bowl, camera, can, laptop, and mug) and contains 4 to 7 objects per scene. Meshes for each object are provided, obtained using an RGB-D reconstruction algorithm. Since we are primarily interested in evaluation, we will focus on the evaluation split, called REAL275, from here.

\begin{figure}
    \centering
    \setlength{\lineskip}{0pt}
    \setlength{\lineskiplimit}{0pt}
    \setlength{\parindent}{0pt}
    \setlength{\baselineskip}{0pt}
    \includegraphics[width=0.33\linewidth]{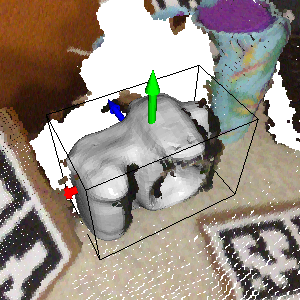}%
    \includegraphics[width=0.33\linewidth]{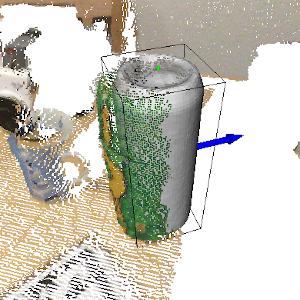}%
    \includegraphics[width=0.33\linewidth]{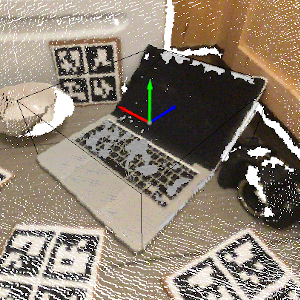}%
    
    \includegraphics[width=0.33\linewidth]{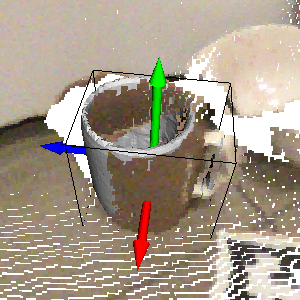}%
    \includegraphics[width=0.33\linewidth]{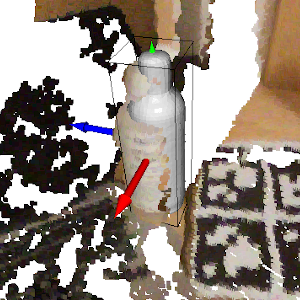}%
    \includegraphics[width=0.33\linewidth]{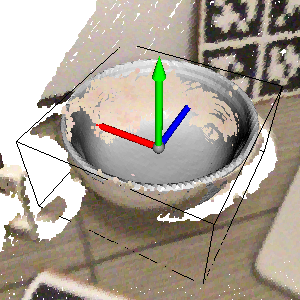}%
    \caption{Examples of REAL275 samples for the 6 object categories. Note that all objects are positioned upright on a table.}
    \label{fig:real275_examples}
\end{figure}

\begin{figure*}
    \centering
    \includegraphics[width=\linewidth]{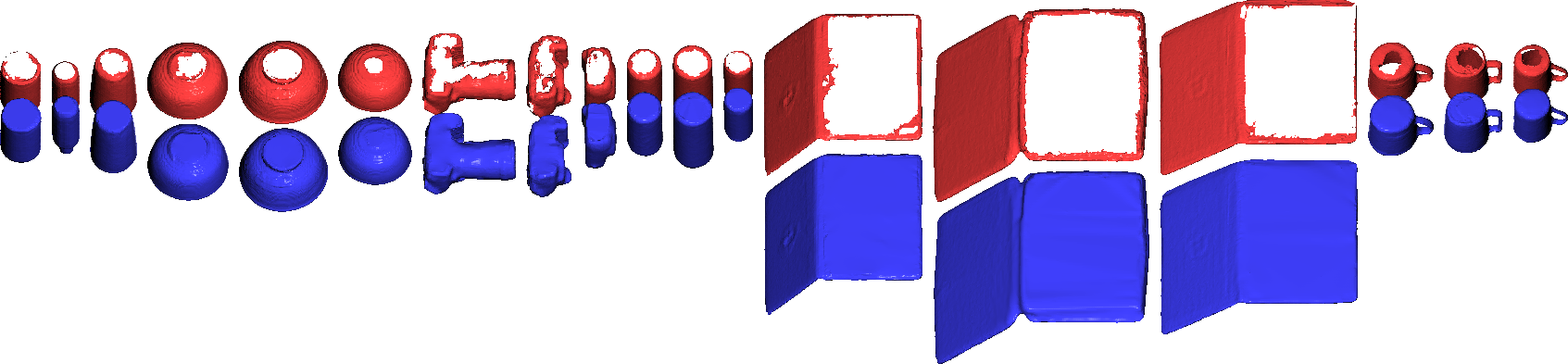}%
    \caption{Bottom view of the original meshes of the REAL275 dataset (red) and our completed meshes (blue). Backfaces are not rendered highlighting the missing surfaces in the original meshes.}
    \label{fig:real275_meshes}
\end{figure*}

\begin{figure}[htb]
    \centering
    \begin{subfigure}{0.35\textwidth}
        \includegraphics[width=0.49\textwidth]{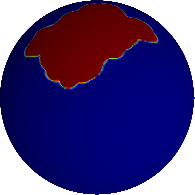}
        \includegraphics[width=0.49\textwidth]{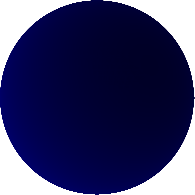}
        \caption{REAL275}\label{fig:orientation_real275}
    \end{subfigure}
    \begin{subfigure}{0.35\textwidth}
        \includegraphics[width=0.49\textwidth]{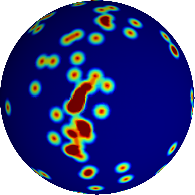}
        \includegraphics[width=0.49\textwidth]{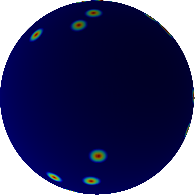}
        \caption{REDWOOD75}\label{fig:orientation_redwood75}
    \end{subfigure}
    \caption{Distribution of the up-axis in REAL275 and REDWOOD75 datasets. REDWOOD75 covers a larger variety of orientations.\thefontsize}
    \label{fig:orientation_dist}
\end{figure}

\Cref{fig:real275_examples} shows point sets with their corresponding ground-truth annotations. Note that all objects are upright on planar surfaces. Similar constrained orientations can be found in the training splits of the CAMERA and REAL datasets. \cref{fig:orientation_real275} visualizes the distribution of orientations contained in the REAL275 dataset. Note that such constraints, present in training and test data, can significantly simplify the learning problem as pose and shape ambiguities disappear (e.g., upright or upside-down can).

The REAL275 dataset was originally proposed for pose estimation and was only later used to evaluate shape estimates. However, the ground-truth meshes included in the dataset are not complete. In particular, the whole bottom is missing (\cref{fig:real275_meshes}, red meshes at bottom), which can cause correct estimates to be considered wrong. To fix this, we manually completed the missing surfaces in Blender \cite{blender} and include the fixed meshes in our toolbox.

\subsubsection{Redwood}\label{sec:redwood}

To evaluate methods on less constrained orientations, we contribute annotations for a set of images in the Redwood dataset \cite{choi2016large}. The Redwood dataset contains sequences of handheld objects being freely rotated in front of the camera. No ground-truth reconstructions are provided for these sequences. 

We annotated pose and shape for 3 categories (bottle, bowl, mug) for 5 sequences each. These annotations were created by manually creating OBBs in multiple frames and exploiting potential symmetries of the object. Alignment of OBBs with previous annotations was sped up and refined by using the ICP algorithm. For each of the annotated sequences we took a subset of 5 frames covering various orientations. We will refer to this set of annotations as REDWOOD75.

\begin{figure}[tb]
    \centering
    \includegraphics[scale=0.7]{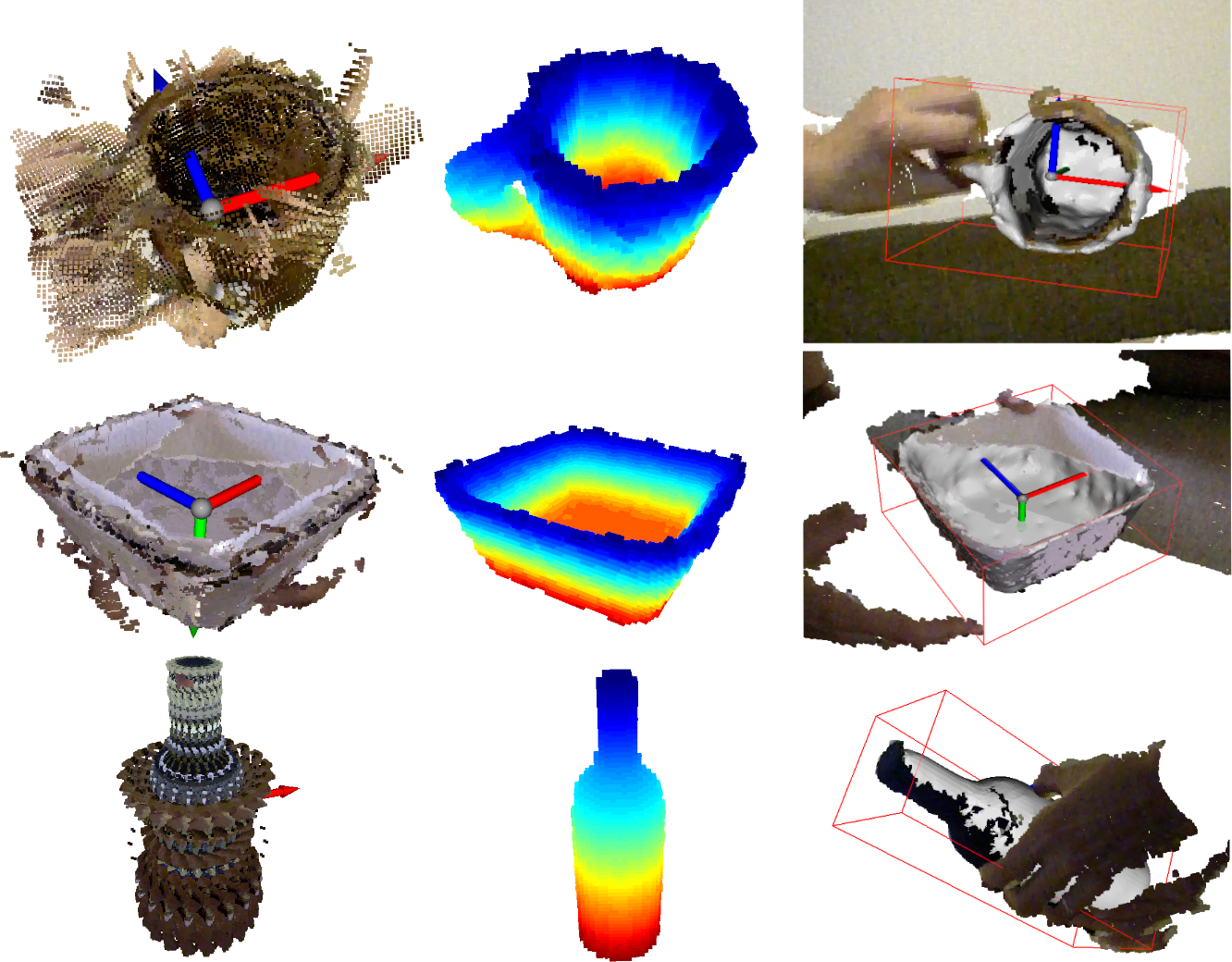}
    \caption{Manual annotations for Redwood dataset. The left column shows the cropped, accumulated point sets (including symmetries) extracted from annotated bounding boxes. The middle columns shows the voxel grid after carving. The right column shows the extracted mesh, overlaid with the point set.}\label{fig:redwood_annotations}
\end{figure}

To reconstruct the shape from the partially occluded and noisy depth data, we start from a dense voxel grid inside the bounding box and apply voxel carving using the annotated frames to remove hands and other temporary occlusions. The remaining voxel grid contains only voxels that are not observed to be free in any of the annotated frames. From this voxel grid, we extract a mesh and apply Laplacian smoothing. \cref{fig:redwood_annotations} visualizes the annotation process.

Note that this method only approximates the real shape and is sensitive to misaligned bounding boxes and missing depth data. Especially thin surfaces and details, such as mug handles, are difficult to extract accurately due to sensor noise. Additionally, alignment errors can easily accumulate, resulting in too large or too small objects. However, the annotations are accurate enough to evaluate the performance of current methods on unconstrained orientations.

\begin{figure}[htb]
    \centering
    \setlength{\lineskip}{0pt}
    \setlength{\lineskiplimit}{0pt}
    \setlength{\parindent}{0pt}
    \setlength{\baselineskip}{0pt}
    \includegraphics[width=0.33\linewidth]{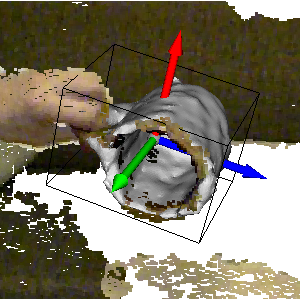}%
    \includegraphics[width=0.33\linewidth]{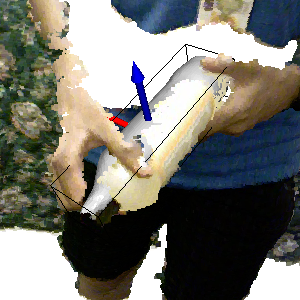}%
    \includegraphics[width=0.33\linewidth]{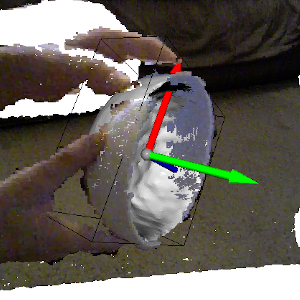}%
    
    \includegraphics[width=0.33\linewidth]{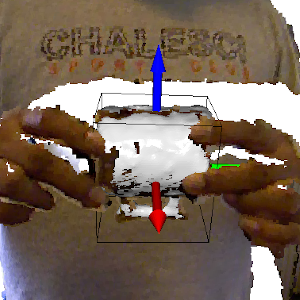}%
    \includegraphics[width=0.33\linewidth]{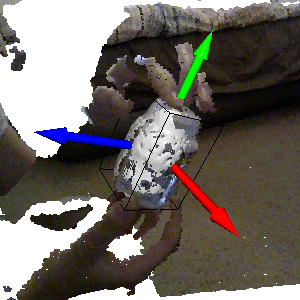}%
    \includegraphics[width=0.33\linewidth]{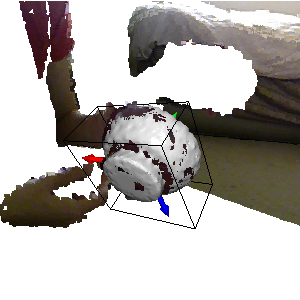}%
    \caption{Examples of REDWOOD75 samples.}
    \label{fig:redwood_examples}
\end{figure}

To produce the final ground truth, we compute tight bounding boxes based on the extracted meshes. For both datasets we normalize the orientation convention to be consistent with the ShapeNet \cite{chang2015} dataset. \Cref{fig:redwood_examples} shows examples of the final annotations. In \cref{fig:orientation_dist} we compare the orientation distribution of REDWOOD75 and REAL275. Note that the orientations in REDWOOD75 are significantly less constrained than in REAL275.

\section{Experiments}\label{sec:experiments}

We follow our proposed evaluation protocol and compare the nine methods described in Section \ref{sec:evalmethods}. All of the methods estimate 6D pose and reconstruct the shape. Most methods represent the shape as point sets; for iCaps \cite{deng2022icaps} and SDFEst \cite{bruns2022sdfest} we convert the estimated signed distance field first into a mesh and then into a point set by sampling 10000 points on the mesh surface.

For all methods, we closely followed the published inference code and verified that our method interface produced results similar to their evaluation code. For most methods we were able to achieve results on par with their published results. However, CASS' reconstructed point sets were significantly worse except for the laptop category and ASM-Net often predicted negative scales, which caused some reconstructions to be upside down, while the object frame $\mathrm{o}$ is predicted in the correct orientation.

We have implemented the metrics, and interfaces to the datasets and methods described in the previous sections using Open3D \cite{zhou2018open3d} and PyTorch \cite{paszke2019pytorch}. We open-source our code as a benchmarking toolbox, with the goal of simplifying fair comparison with state-of-the-art methods. We plan to extend the toolbox as new methods are released.

\subsection{Qualitative Results}\label{sec:qualitative}

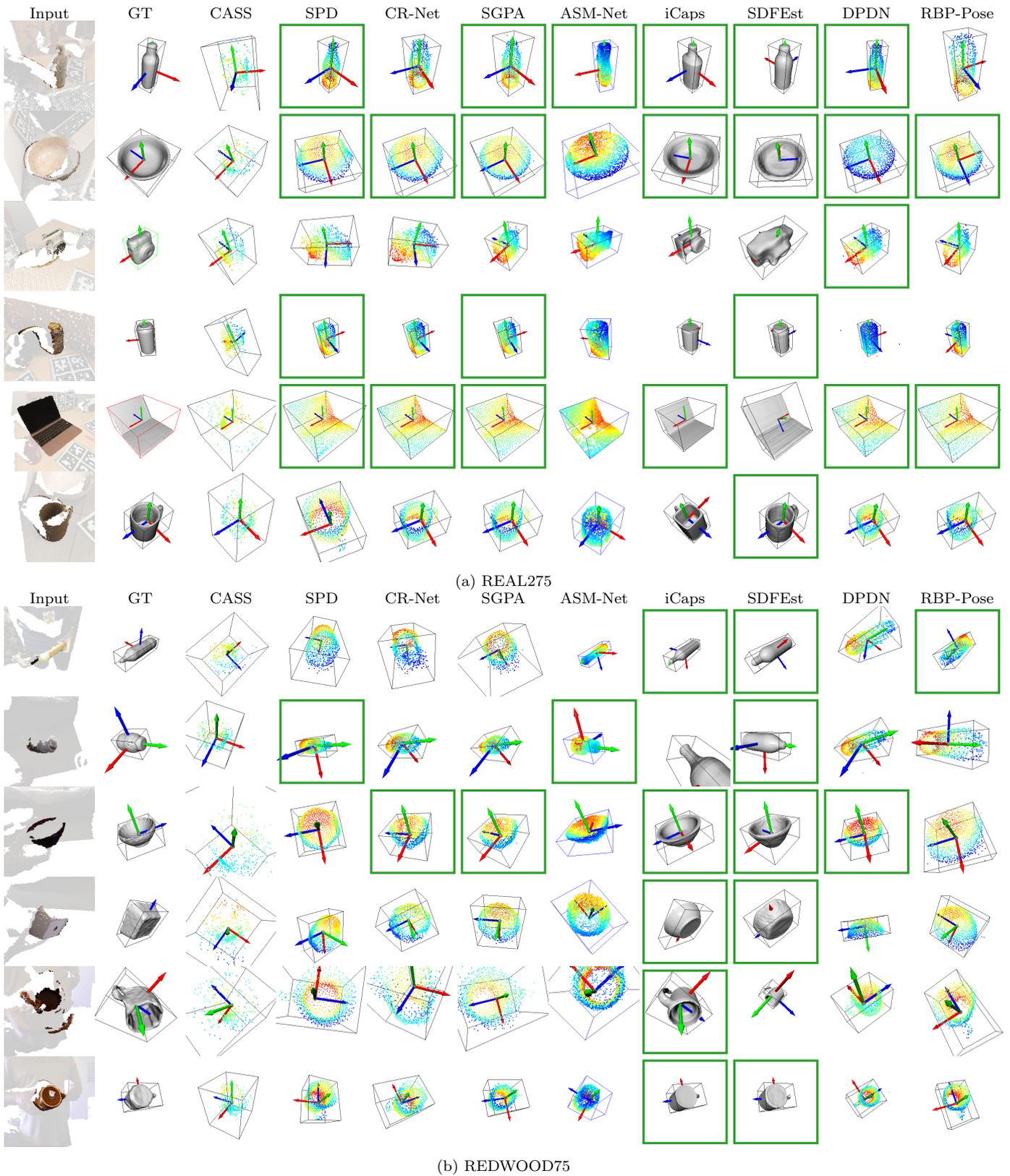
\begin{figure*}[p!]
    \centering
    \begin{subfigure}{\textwidth}
        \centering
        \scriptsize
        \setlength{\fboxsep}{0pt}
        \setlength{\tabcolsep}{0pt}
        \renewcommand{\arraystretch}{0}
        \begin{tabular}{C{0.09\linewidth}C{0.09\linewidth}C{0.09\linewidth}C{0.09\linewidth}C{0.09\linewidth}C{0.09\linewidth}C{0.09\linewidth}C{0.09\linewidth}C{0.09\linewidth}C{0.09\linewidth}C{0.09\linewidth}}
            Input & GT & CASS & SPD & CR-Net & SGPA & ASM-Net & iCaps & SDFEst & DPDN & RBP-Pose 
        \end{tabular}
        \setlength{\lineskip}{0pt}
        \setlength{\lineskiplimit}{0pt}
        \setlength{\parindent}{0pt}
        \setlength{\baselineskip}{0pt}
        \includegraphics[trim={0 0 0 0},clip,width=0.09\textwidth,height=0.09\textwidth]{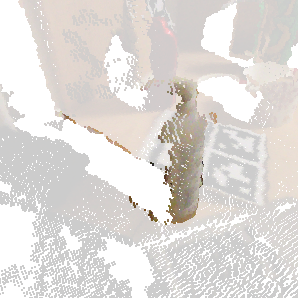}%
        \includegraphics[trim={0 0 0 0},clip,width=0.09\textwidth,height=0.09\textwidth]{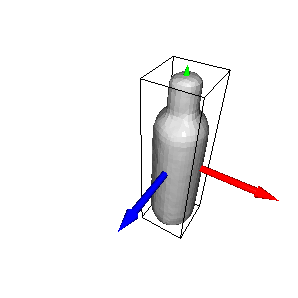}%
        \includegraphics[trim={0 0 0 0},clip,width=0.09\textwidth,height=0.09\textwidth]{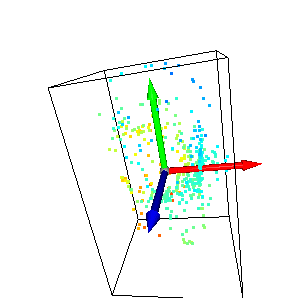}%
        \includegraphics[trim={0 0 0 0},clip,width=0.09\textwidth,height=0.09\textwidth]{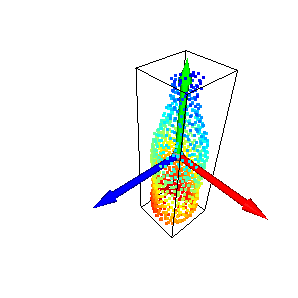}%
        \begin{tikzpicture}[overlay]%
            \draw[forestgreen4416044, very thick,contour=0.7mm] (0,0) rectangle ++(-0.09\textwidth,0.09\textwidth);%
        \end{tikzpicture}%
        \includegraphics[trim={0 0 0 0},clip,width=0.09\textwidth,height=0.09\textwidth]{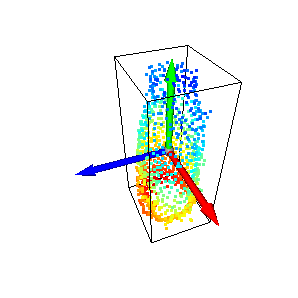}%
        \includegraphics[trim={0 0 0 0},clip,width=0.09\textwidth,height=0.09\textwidth]{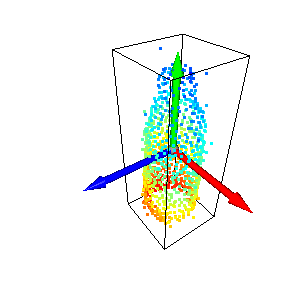}%
        \begin{tikzpicture}[overlay]%
            \draw[forestgreen4416044, very thick,contour=0.7mm] (0,0) rectangle ++(-0.09\textwidth,0.09\textwidth);%
        \end{tikzpicture}%
        \includegraphics[trim={0 0 0 0},clip,width=0.09\textwidth,height=0.09\textwidth]{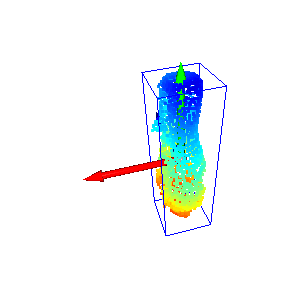}%
        \begin{tikzpicture}[overlay]%
            \draw[forestgreen4416044, very thick,contour=0.7mm] (0,0) rectangle ++(-0.09\textwidth,0.09\textwidth);%
        \end{tikzpicture}%
        \includegraphics[trim={0 0 0 0},clip,width=0.09\textwidth,height=0.09\textwidth]{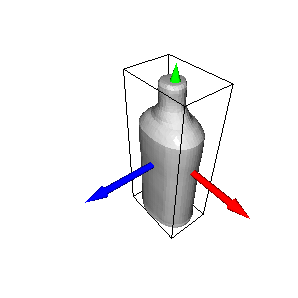}%
        \begin{tikzpicture}[overlay]%
            \draw[forestgreen4416044, very thick,contour=0.7mm] (0,0) rectangle ++(-0.09\textwidth,0.09\textwidth);%
        \end{tikzpicture}%
        \includegraphics[trim={0 0 0 0},clip,width=0.09\textwidth,height=0.09\textwidth]{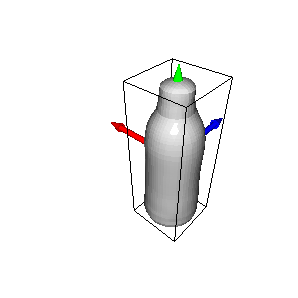}%
        \begin{tikzpicture}[overlay]%
            \draw[forestgreen4416044, very thick,contour=0.7mm] (0,0) rectangle ++(-0.09\textwidth,0.09\textwidth);%
        \end{tikzpicture}%
        \includegraphics[trim={0 0 0 0},clip,width=0.09\textwidth,height=0.09\textwidth]{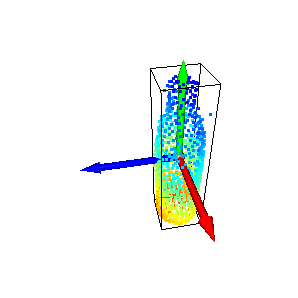}%
        \begin{tikzpicture}[overlay]%
            \draw[forestgreen4416044, very thick,contour=0.7mm] (0,0) rectangle ++(-0.09\textwidth,0.09\textwidth);%
        \end{tikzpicture}%
        \includegraphics[trim={0 0 0 0},clip,width=0.09\textwidth,height=0.09\textwidth]{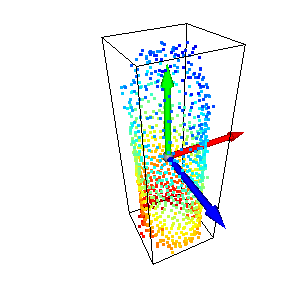}%
        
        \includegraphics[trim={0 0 0 0},clip,width=0.09\textwidth,height=0.09\textwidth]{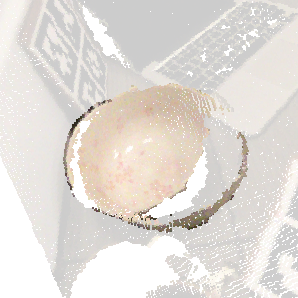}%
        \includegraphics[trim={0 0 0 0},clip,width=0.09\textwidth,height=0.09\textwidth]{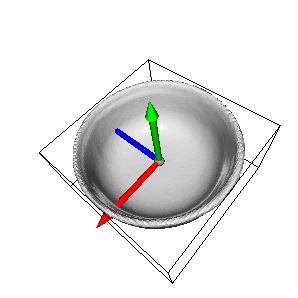}%
        \includegraphics[trim={0 0 0 0},clip,width=0.09\textwidth,height=0.09\textwidth]{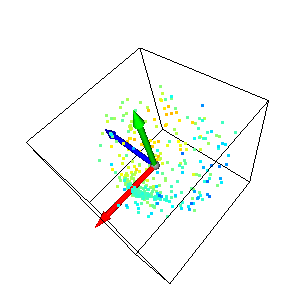}%
        \includegraphics[trim={0 0 0 0},clip,width=0.09\textwidth,height=0.09\textwidth]{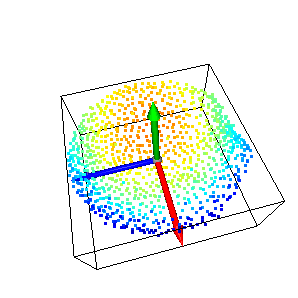}%
        \begin{tikzpicture}[overlay]%
            \draw[forestgreen4416044, very thick,contour=0.7mm] (0,0) rectangle ++(-0.09\textwidth,0.09\textwidth);%
        \end{tikzpicture}%
        \includegraphics[trim={0 0 0 0},clip,width=0.09\textwidth,height=0.09\textwidth]{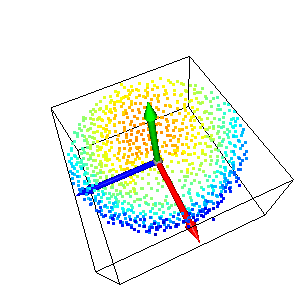}%
        \begin{tikzpicture}[overlay]%
            \draw[forestgreen4416044, very thick,contour=0.7mm] (0,0) rectangle ++(-0.09\textwidth,0.09\textwidth);%
        \end{tikzpicture}%
        \includegraphics[trim={0 0 0 0},clip,width=0.09\textwidth,height=0.09\textwidth]{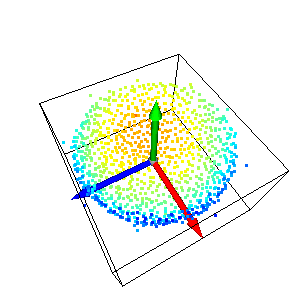}%
        \begin{tikzpicture}[overlay]%
            \draw[forestgreen4416044, very thick,contour=0.7mm] (0,0) rectangle ++(-0.09\textwidth,0.09\textwidth);%
        \end{tikzpicture}%
        \includegraphics[trim={0 0 0 0},clip,width=0.09\textwidth,height=0.09\textwidth]{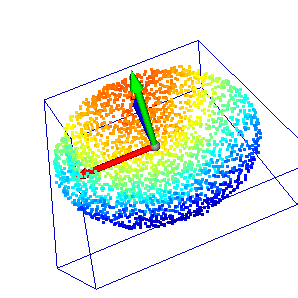}%
        \includegraphics[trim={0 0 0 0},clip,width=0.09\textwidth,height=0.09\textwidth]{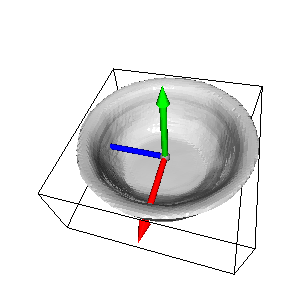}%
        \begin{tikzpicture}[overlay]%
            \draw[forestgreen4416044, very thick,contour=0.7mm] (0,0) rectangle ++(-0.09\textwidth,0.09\textwidth);%
        \end{tikzpicture}%
        \includegraphics[trim={0 0 0 0},clip,width=0.09\textwidth,height=0.09\textwidth]{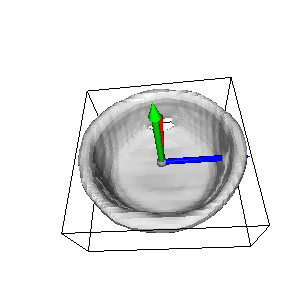}%
        \begin{tikzpicture}[overlay]%
            \draw[forestgreen4416044, very thick,contour=0.7mm] (0,0) rectangle ++(-0.09\textwidth,0.09\textwidth);%
        \end{tikzpicture}%
        \includegraphics[trim={0 0 0 0},clip,width=0.09\textwidth,height=0.09\textwidth]{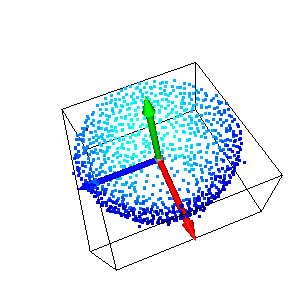}%
        \begin{tikzpicture}[overlay]%
            \draw[forestgreen4416044, very thick,contour=0.7mm] (0,0) rectangle ++(-0.09\textwidth,0.09\textwidth);%
        \end{tikzpicture}%
        \includegraphics[trim={0 0 0 0},clip,width=0.09\textwidth,height=0.09\textwidth]{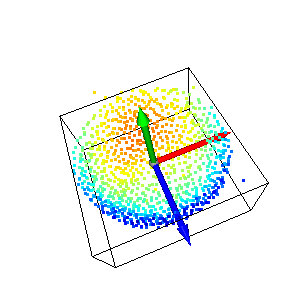}%
        \begin{tikzpicture}[overlay]%
            \draw[forestgreen4416044, very thick,contour=0.7mm] (0,0) rectangle ++(-0.09\textwidth,0.09\textwidth);%
        \end{tikzpicture}%
        
        \includegraphics[trim={0 0 0 0},clip,width=0.09\textwidth,height=0.09\textwidth]{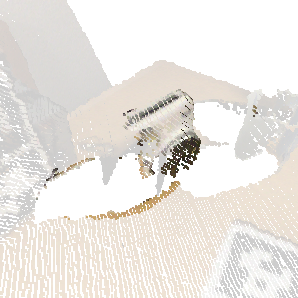}%
        \includegraphics[trim={0 0 0 0},clip,width=0.09\textwidth,height=0.09\textwidth]{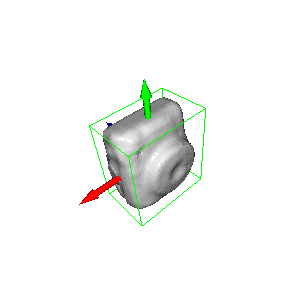}%
        \includegraphics[trim={0 0 0 0},clip,width=0.09\textwidth,height=0.09\textwidth]{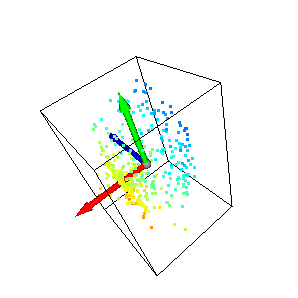}%
        \includegraphics[trim={0 0 0 0},clip,width=0.09\textwidth,height=0.09\textwidth]{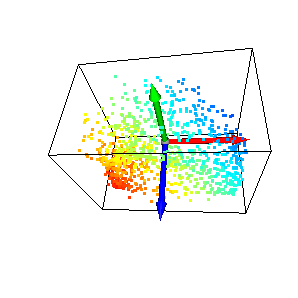}%
        \includegraphics[trim={0 0 0 0},clip,width=0.09\textwidth,height=0.09\textwidth]{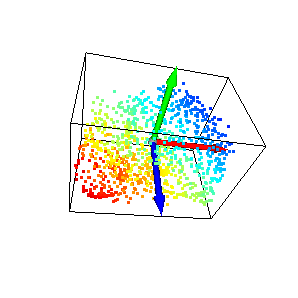}%
        \includegraphics[trim={0 0 0 0},clip,width=0.09\textwidth,height=0.09\textwidth]{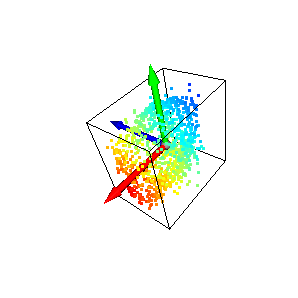}%
        \includegraphics[trim={0 0 0 0},clip,width=0.09\textwidth,height=0.09\textwidth]{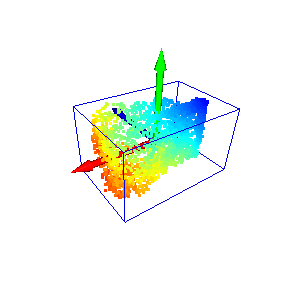}%
        \includegraphics[trim={0 0 0 0},clip,width=0.09\textwidth,height=0.09\textwidth]{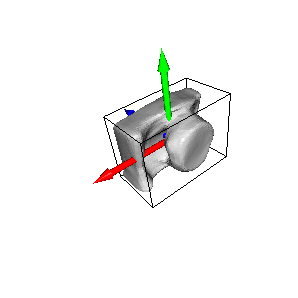}%
        \includegraphics[trim={0 0 0 0},clip,width=0.09\textwidth,height=0.09\textwidth]{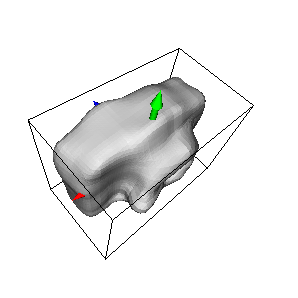}%
        \includegraphics[trim={0 0 0 0},clip,width=0.09\textwidth,height=0.09\textwidth]{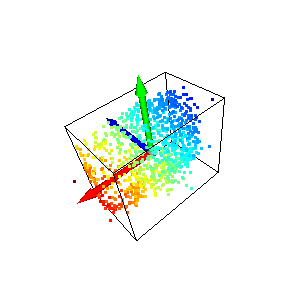}%
        \begin{tikzpicture}[overlay]%
            \draw[forestgreen4416044, very thick,contour=0.7mm] (0,0) rectangle ++(-0.09\textwidth,0.09\textwidth);%
        \end{tikzpicture}%
        \includegraphics[trim={0 0 0 0},clip,width=0.09\textwidth,height=0.09\textwidth]{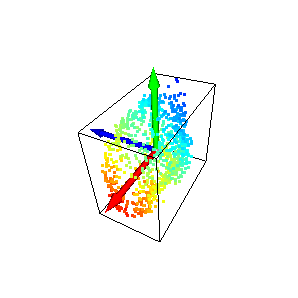}%
        
        \includegraphics[trim={0 0 0 0},clip,width=0.09\textwidth,height=0.09\textwidth]{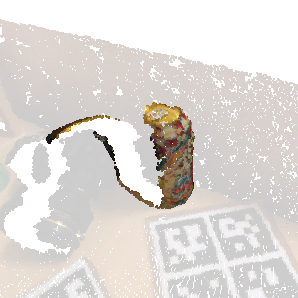}%
        \includegraphics[trim={0 0 0 0},clip,width=0.09\textwidth,height=0.09\textwidth]{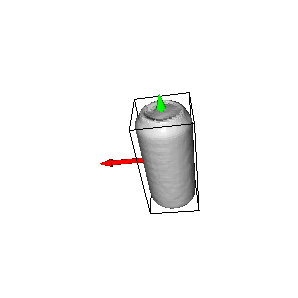}%
        \includegraphics[trim={0 0 0 0},clip,width=0.09\textwidth,height=0.09\textwidth]{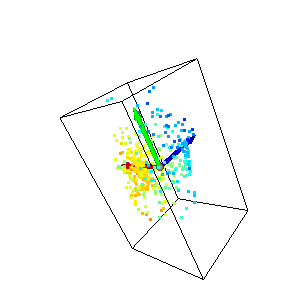}%
        \includegraphics[trim={0 0 0 0},clip,width=0.09\textwidth,height=0.09\textwidth]{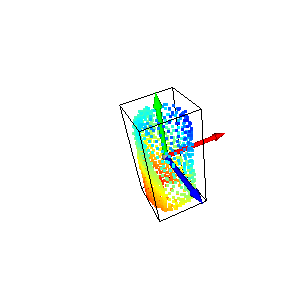}%
        \begin{tikzpicture}[overlay]%
            \draw[forestgreen4416044, very thick,contour=0.7mm] (0,0) rectangle ++(-0.09\textwidth,0.09\textwidth);%
        \end{tikzpicture}%
        \includegraphics[trim={0 0 0 0},clip,width=0.09\textwidth,height=0.09\textwidth]{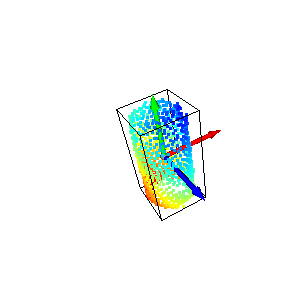}%
        \includegraphics[trim={0 0 0 0},clip,width=0.09\textwidth,height=0.09\textwidth]{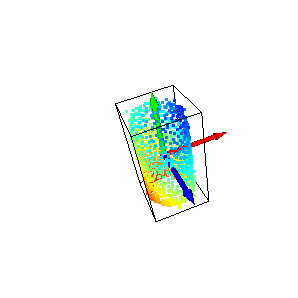}%
        \begin{tikzpicture}[overlay]%
            \draw[forestgreen4416044, very thick,contour=0.7mm] (0,0) rectangle ++(-0.09\textwidth,0.09\textwidth);%
        \end{tikzpicture}%
        \includegraphics[trim={0 0 0 0},clip,width=0.09\textwidth,height=0.09\textwidth]{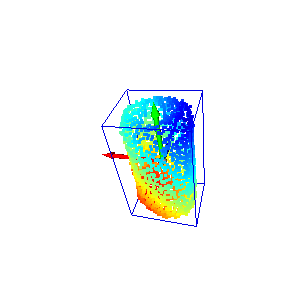}%
        \includegraphics[trim={0 0 0 0},clip,width=0.09\textwidth,height=0.09\textwidth]{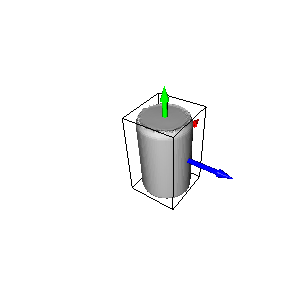}%
        \includegraphics[trim={0 0 0 0},clip,width=0.09\textwidth,height=0.09\textwidth]{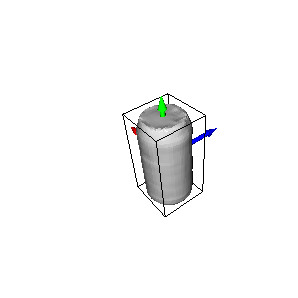}%
        \begin{tikzpicture}[overlay]%
            \draw[forestgreen4416044, very thick,contour=0.7mm] (0,0) rectangle ++(-0.09\textwidth,0.09\textwidth);%
        \end{tikzpicture}%
        \includegraphics[trim={0 0 0 0},clip,width=0.09\textwidth,height=0.09\textwidth]{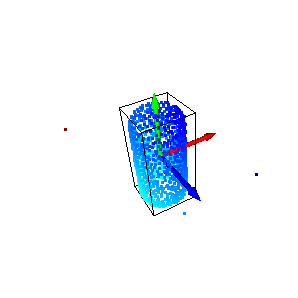}%
        \includegraphics[trim={0 0 0 0},clip,width=0.09\textwidth,height=0.09\textwidth]{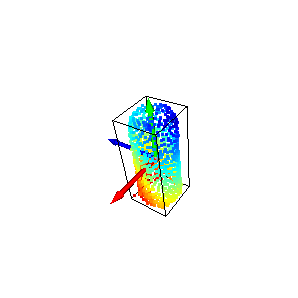}%
        
        \includegraphics[trim={0 0 0 0},clip,width=0.09\textwidth,height=0.09\textwidth]{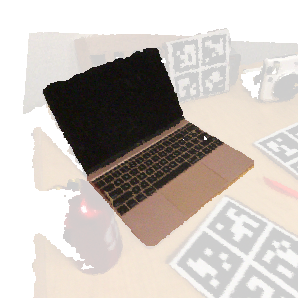}%
        \includegraphics[trim={0 0 0 0},clip,width=0.09\textwidth,height=0.09\textwidth]{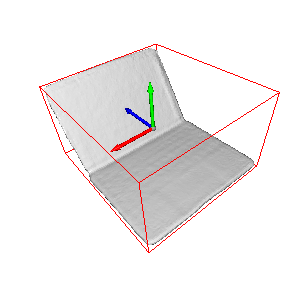}%
        \includegraphics[trim={0 0 0 0},clip,width=0.09\textwidth,height=0.09\textwidth]{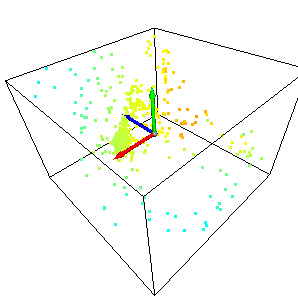}%
        \includegraphics[trim={0 0 0 0},clip,width=0.09\textwidth,height=0.09\textwidth]{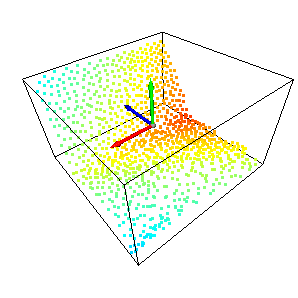}%
        \begin{tikzpicture}[overlay]%
            \draw[forestgreen4416044, very thick,contour=0.7mm] (0,0) rectangle ++(-0.09\textwidth,0.09\textwidth);%
        \end{tikzpicture}%
        \includegraphics[trim={0 0 0 0},clip,width=0.09\textwidth,height=0.09\textwidth]{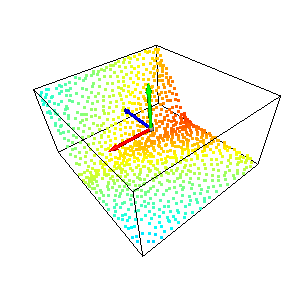}%
        \begin{tikzpicture}[overlay]%
            \draw[forestgreen4416044, very thick,contour=0.7mm] (0,0) rectangle ++(-0.09\textwidth,0.09\textwidth);%
        \end{tikzpicture}%
        \includegraphics[trim={0 0 0 0},clip,width=0.09\textwidth,height=0.09\textwidth]{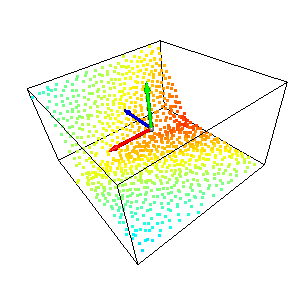}%
        \begin{tikzpicture}[overlay]%
            \draw[forestgreen4416044, very thick,contour=0.7mm] (0,0) rectangle ++(-0.09\textwidth,0.09\textwidth);%
        \end{tikzpicture}%
        \includegraphics[trim={0 0 0 0},clip,width=0.09\textwidth,height=0.09\textwidth]{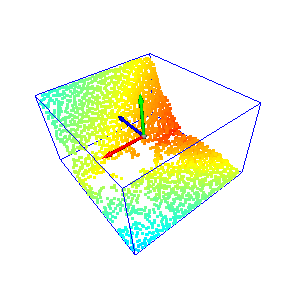}%
        \includegraphics[trim={0 0 0 0},clip,width=0.09\textwidth,height=0.09\textwidth]{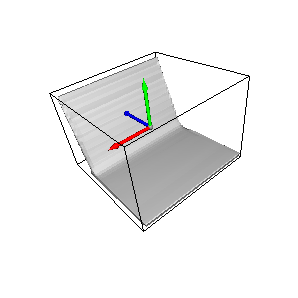}%
        \begin{tikzpicture}[overlay]%
            \draw[forestgreen4416044, very thick,contour=0.7mm] (0,0) rectangle ++(-0.09\textwidth,0.09\textwidth);%
        \end{tikzpicture}%
        \includegraphics[trim={0 0 0 0},clip,width=0.09\textwidth,height=0.09\textwidth]{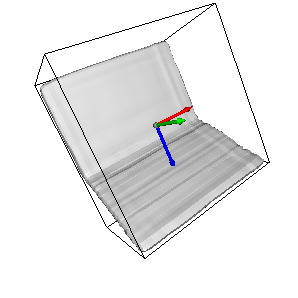}%
        \includegraphics[trim={0 0 0 0},clip,width=0.09\textwidth,height=0.09\textwidth]{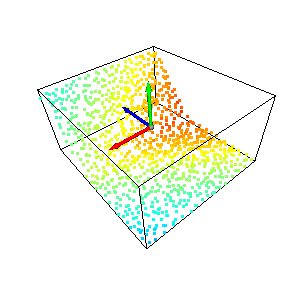}%
        \begin{tikzpicture}[overlay]%
            \draw[forestgreen4416044, very thick,contour=0.7mm] (0,0) rectangle ++(-0.09\textwidth,0.09\textwidth);%
        \end{tikzpicture}%
        \includegraphics[trim={0 0 0 0},clip,width=0.09\textwidth,height=0.09\textwidth]{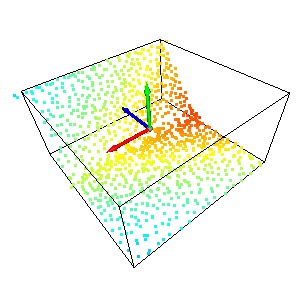}%
        \begin{tikzpicture}[overlay]%
            \draw[forestgreen4416044, very thick,contour=0.7mm] (0,0) rectangle ++(-0.09\textwidth,0.09\textwidth);%
        \end{tikzpicture}%
        
        \includegraphics[trim={0 0 0 0},clip,width=0.09\textwidth,height=0.09\textwidth]{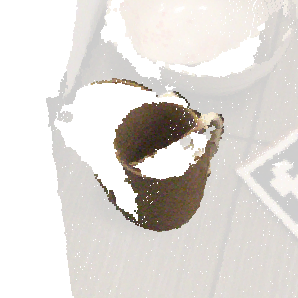}%
        \includegraphics[trim={0 0 0 0},clip,width=0.09\textwidth,height=0.09\textwidth]{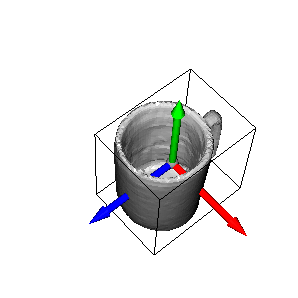}%
        \includegraphics[trim={0 0 0 0},clip,width=0.09\textwidth,height=0.09\textwidth]{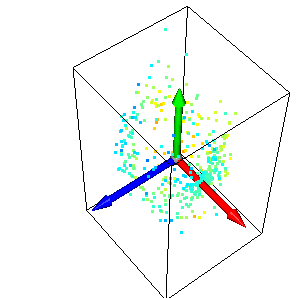}%
        \includegraphics[trim={0 0 0 0},clip,width=0.09\textwidth,height=0.09\textwidth]{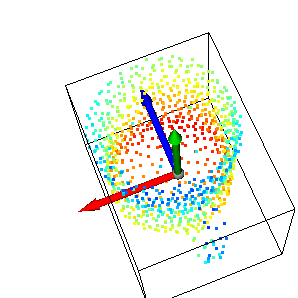}%
        \includegraphics[trim={0 0 0 0},clip,width=0.09\textwidth,height=0.09\textwidth]{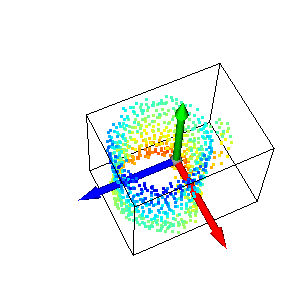}%
        \includegraphics[trim={0 0 0 0},clip,width=0.09\textwidth,height=0.09\textwidth]{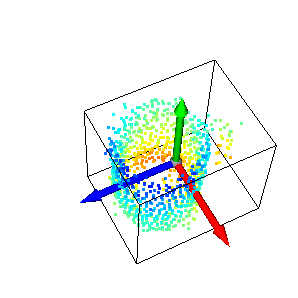}%
        \includegraphics[trim={0 0 0 0},clip,width=0.09\textwidth,height=0.09\textwidth]{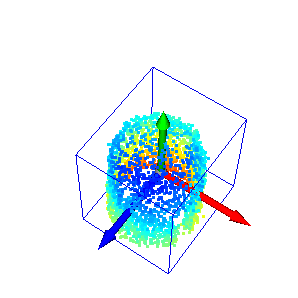}%
        \includegraphics[trim={0 0 0 0},clip,width=0.09\textwidth,height=0.09\textwidth]{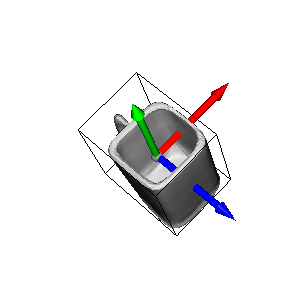}%
        \includegraphics[trim={0 0 0 0},clip,width=0.09\textwidth,height=0.09\textwidth]{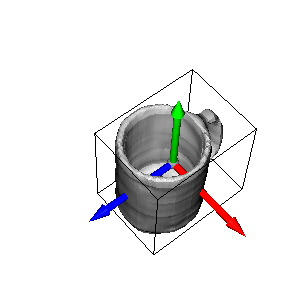}%
        \begin{tikzpicture}[overlay]%
            \draw[forestgreen4416044, very thick,contour=0.7mm] (0,0) rectangle ++(-0.09\textwidth,0.09\textwidth);%
        \end{tikzpicture}%
        \includegraphics[trim={0 0 0 0},clip,width=0.09\textwidth,height=0.09\textwidth]{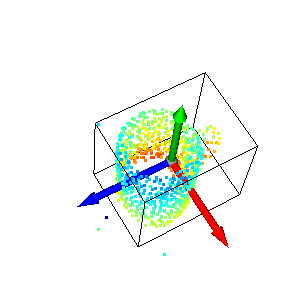}%
        \includegraphics[trim={0 0 0 0},clip,width=0.09\textwidth,height=0.09\textwidth]{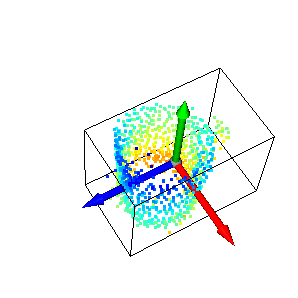}%
        \caption{REAL275}\label{fig:qualitativer275}
    \end{subfigure}
    \begin{subfigure}{\textwidth}
        \centering
        \scriptsize
        \setlength{\fboxsep}{0pt}
        \setlength{\tabcolsep}{0pt}
        \renewcommand{\arraystretch}{0}
        \begin{tabular}{C{0.09\linewidth}C{0.09\linewidth}C{0.09\linewidth}C{0.09\linewidth}C{0.09\linewidth}C{0.09\linewidth}C{0.09\linewidth}C{0.09\linewidth}C{0.09\linewidth}C{0.09\linewidth}C{0.09\linewidth}}
            Input & GT & CASS & SPD & CR-Net & SGPA & ASM-Net & iCaps & SDFEst & DPDN & RBP-Pose 
        \end{tabular}
        \setlength{\lineskip}{0pt}
        \setlength{\lineskiplimit}{0pt}
        \setlength{\parindent}{0pt}
        \setlength{\baselineskip}{0pt}
        \includegraphics[trim={0 0 0 0},clip,width=0.09\textwidth,height=0.09\textwidth]{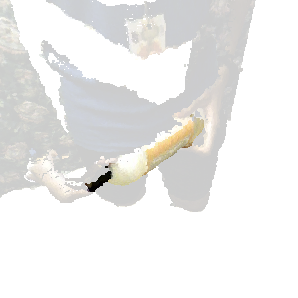}%
        \includegraphics[trim={0 0 0 0},clip,width=0.09\textwidth,height=0.09\textwidth]{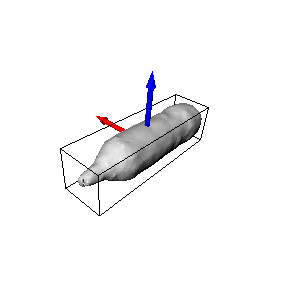}%
        \includegraphics[trim={0 0 0 0},clip,width=0.09\textwidth,height=0.09\textwidth]{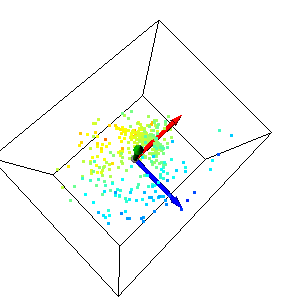}%
        \includegraphics[trim={0 0 0 0},clip,width=0.09\textwidth,height=0.09\textwidth]{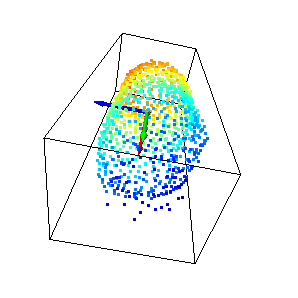}%
        \includegraphics[trim={0 0 0 0},clip,width=0.09\textwidth,height=0.09\textwidth]{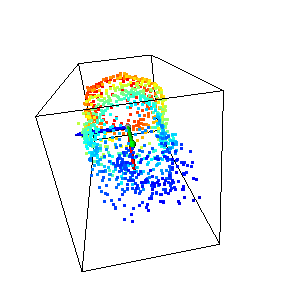}%
        \includegraphics[trim={0 0 0 0},clip,width=0.09\textwidth,height=0.09\textwidth]{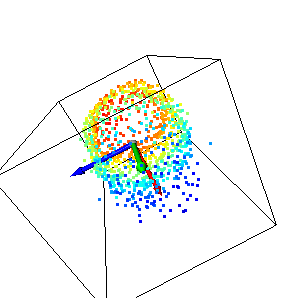}%
        \includegraphics[trim={0 0 0 0},clip,width=0.09\textwidth,height=0.09\textwidth]{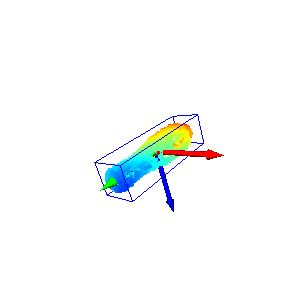}%
        \includegraphics[trim={0 0 0 0},clip,width=0.09\textwidth,height=0.09\textwidth]{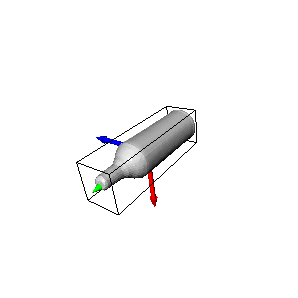}%
        \begin{tikzpicture}[overlay]%
            \draw[forestgreen4416044, very thick,contour=0.7mm] (0,0) rectangle ++(-0.09\textwidth,0.09\textwidth);%
        \end{tikzpicture}%
        \includegraphics[trim={0 0 0 0},clip,width=0.09\textwidth,height=0.09\textwidth]{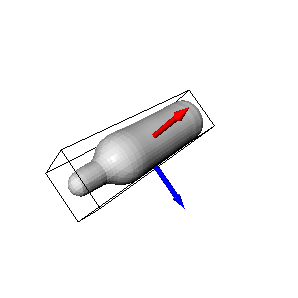}%
        \begin{tikzpicture}[overlay]%
            \draw[forestgreen4416044, very thick,contour=0.7mm] (0,0) rectangle ++(-0.09\textwidth,0.09\textwidth);%
        \end{tikzpicture}%
        \includegraphics[trim={0 0 0 0},clip,width=0.09\textwidth,height=0.09\textwidth]{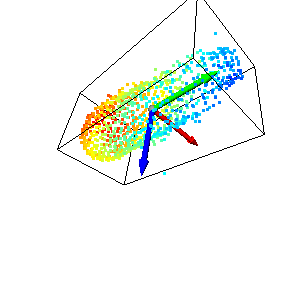}%
        \includegraphics[trim={0 0 0 0},clip,width=0.09\textwidth,height=0.09\textwidth]{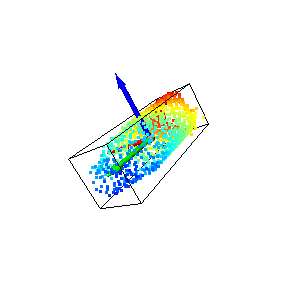}%
        \begin{tikzpicture}[overlay]%
            \draw[forestgreen4416044, very thick,contour=0.7mm] (0,0) rectangle ++(-0.09\textwidth,0.09\textwidth);%
        \end{tikzpicture}%
        
        \includegraphics[trim={0 0 0 0},clip,width=0.09\textwidth,height=0.09\textwidth]{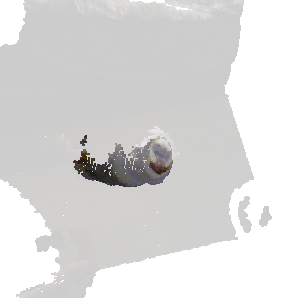}%
        \includegraphics[trim={0 0 0 0},clip,width=0.09\textwidth,height=0.09\textwidth]{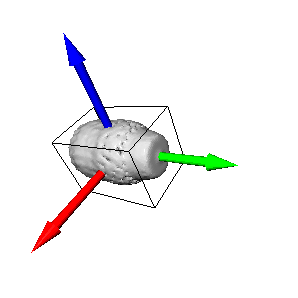}%
        \includegraphics[trim={0 0 0 0},clip,width=0.09\textwidth,height=0.09\textwidth]{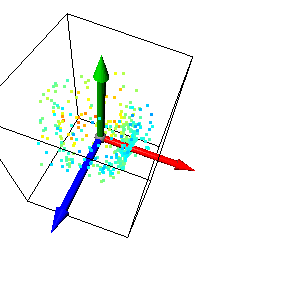}%
        \includegraphics[trim={0 0 0 0},clip,width=0.09\textwidth,height=0.09\textwidth]{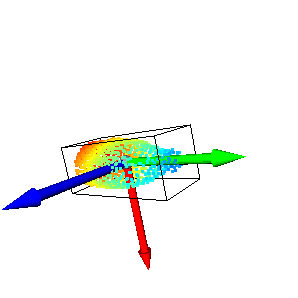}%
        \begin{tikzpicture}[overlay]%
            \draw[forestgreen4416044, very thick,contour=0.7mm] (0,0) rectangle ++(-0.09\textwidth,0.09\textwidth);%
        \end{tikzpicture}%
        \includegraphics[trim={0 0 0 0},clip,width=0.09\textwidth,height=0.09\textwidth]{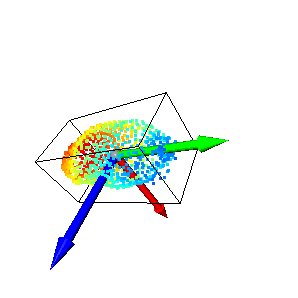}%
        \includegraphics[trim={0 0 0 0},clip,width=0.09\textwidth,height=0.09\textwidth]{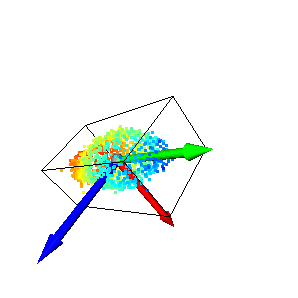}%
        \includegraphics[trim={0 0 0 0},clip,width=0.09\textwidth,height=0.09\textwidth]{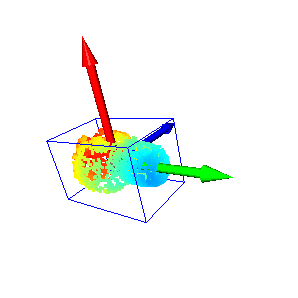}%
        \begin{tikzpicture}[overlay]%
            \draw[forestgreen4416044, very thick,contour=0.7mm] (0,0) rectangle ++(-0.09\textwidth,0.09\textwidth);%
        \end{tikzpicture}%
        \includegraphics[trim={0 0 0 0},clip,width=0.09\textwidth,height=0.09\textwidth]{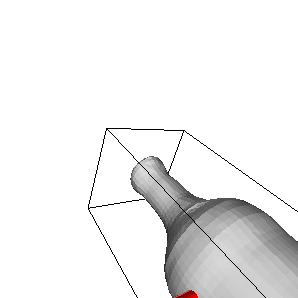}%
        \includegraphics[trim={0 0 0 0},clip,width=0.09\textwidth,height=0.09\textwidth]{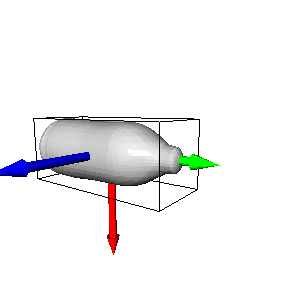}%
        \begin{tikzpicture}[overlay]%
            \draw[forestgreen4416044, very thick,contour=0.7mm] (0,0) rectangle ++(-0.09\textwidth,0.09\textwidth);%
        \end{tikzpicture}%
        \includegraphics[trim={0 0 0 0},clip,width=0.09\textwidth,height=0.09\textwidth]{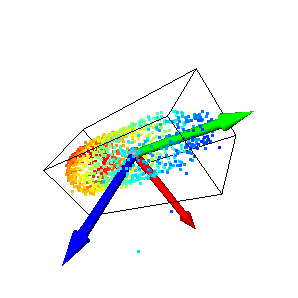}%
        \includegraphics[trim={0 0 0 0},clip,width=0.09\textwidth,height=0.09\textwidth]{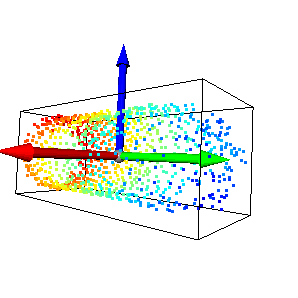}%
        
        \includegraphics[trim={0 0 0 0},clip,width=0.09\textwidth,height=0.09\textwidth]{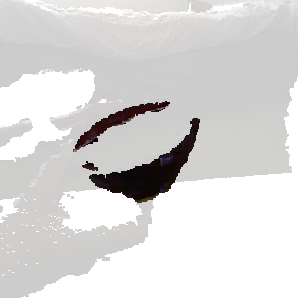}%
        \includegraphics[trim={0 0 0 0},clip,width=0.09\textwidth,height=0.09\textwidth]{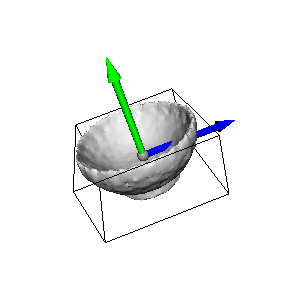}%
        \includegraphics[trim={0 0 0 0},clip,width=0.09\textwidth,height=0.09\textwidth]{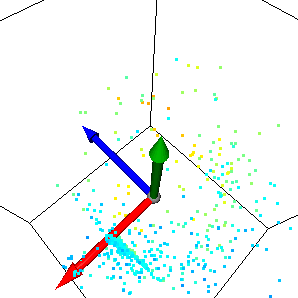}%
        \includegraphics[trim={0 0 0 0},clip,width=0.09\textwidth,height=0.09\textwidth]{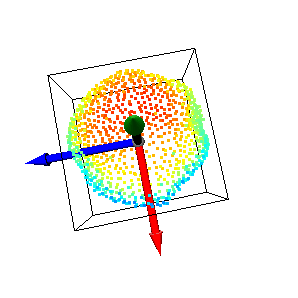}%
        \includegraphics[trim={0 0 0 0},clip,width=0.09\textwidth,height=0.09\textwidth]{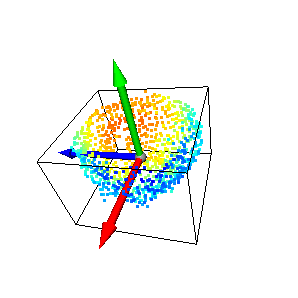}%
        \begin{tikzpicture}[overlay]%
            \draw[forestgreen4416044, very thick,contour=0.7mm] (0,0) rectangle ++(-0.09\textwidth,0.09\textwidth);%
        \end{tikzpicture}%
        \includegraphics[trim={0 0 0 0},clip,width=0.09\textwidth,height=0.09\textwidth]{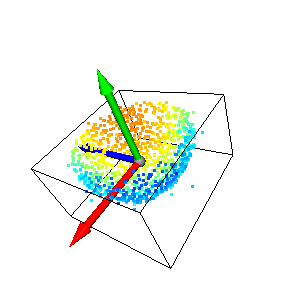}%
        \begin{tikzpicture}[overlay]%
            \draw[forestgreen4416044, very thick,contour=0.7mm] (0,0) rectangle ++(-0.09\textwidth,0.09\textwidth);%
        \end{tikzpicture}%
        \includegraphics[trim={0 0 0 0},clip,width=0.09\textwidth,height=0.09\textwidth]{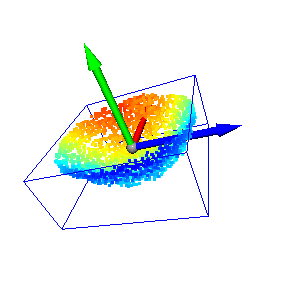}%
        \includegraphics[trim={0 0 0 0},clip,width=0.09\textwidth,height=0.09\textwidth]{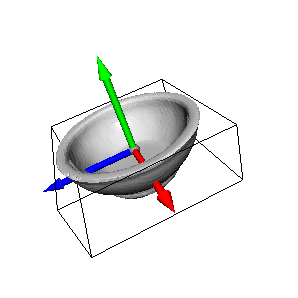}%
        \begin{tikzpicture}[overlay]%
            \draw[forestgreen4416044, very thick,contour=0.7mm] (0,0) rectangle ++(-0.09\textwidth,0.09\textwidth);%
        \end{tikzpicture}%
        \includegraphics[trim={0 0 0 0},clip,width=0.09\textwidth,height=0.09\textwidth]{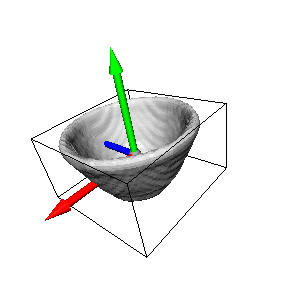}%
        \begin{tikzpicture}[overlay]%
            \draw[forestgreen4416044, very thick,contour=0.7mm] (0,0) rectangle ++(-0.09\textwidth,0.09\textwidth);%
        \end{tikzpicture}%
        \includegraphics[trim={0 0 0 0},clip,width=0.09\textwidth,height=0.09\textwidth]{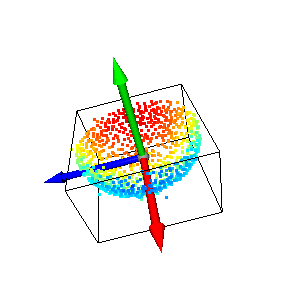}%
        \begin{tikzpicture}[overlay]%
            \draw[forestgreen4416044, very thick,contour=0.7mm] (0,0) rectangle ++(-0.09\textwidth,0.09\textwidth);%
        \end{tikzpicture}%
        \includegraphics[trim={0 0 0 0},clip,width=0.09\textwidth,height=0.09\textwidth]{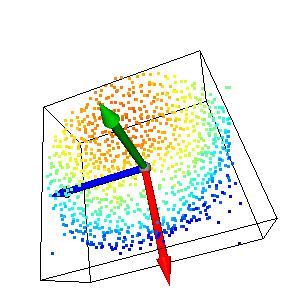}%
        
        \includegraphics[trim={0 0 0 0},clip,width=0.09\textwidth,height=0.09\textwidth]{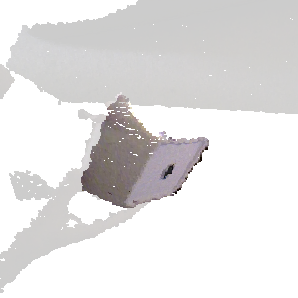}%
        \includegraphics[trim={0 0 0 0},clip,width=0.09\textwidth,height=0.09\textwidth]{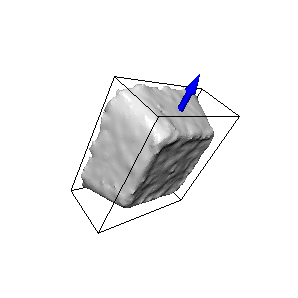}%
        \includegraphics[trim={0 0 0 0},clip,width=0.09\textwidth,height=0.09\textwidth]{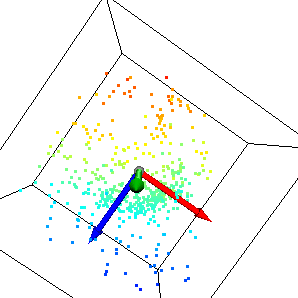}%
        \includegraphics[trim={0 0 0 0},clip,width=0.09\textwidth,height=0.09\textwidth]{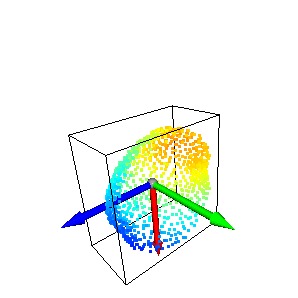}%
        \includegraphics[trim={0 0 0 0},clip,width=0.09\textwidth,height=0.09\textwidth]{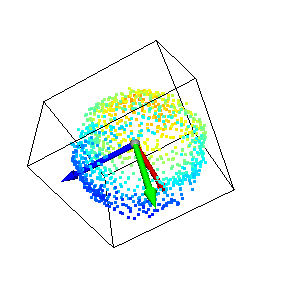}%
        \includegraphics[trim={0 0 0 0},clip,width=0.09\textwidth,height=0.09\textwidth]{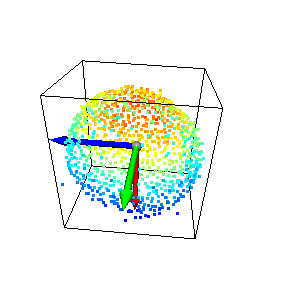}%
        \includegraphics[trim={0 0 0 0},clip,width=0.09\textwidth,height=0.09\textwidth]{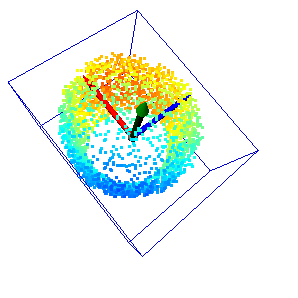}%
        \includegraphics[trim={0 0 0 0},clip,width=0.09\textwidth,height=0.09\textwidth]{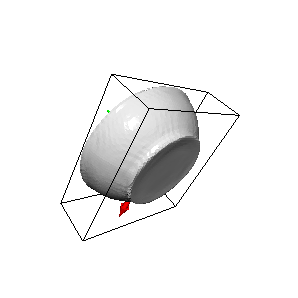}%
        \begin{tikzpicture}[overlay]%
            \draw[forestgreen4416044, very thick,contour=0.7mm] (0,0) rectangle ++(-0.09\textwidth,0.09\textwidth);%
        \end{tikzpicture}%
        \includegraphics[trim={0 0 0 0},clip,width=0.09\textwidth,height=0.09\textwidth]{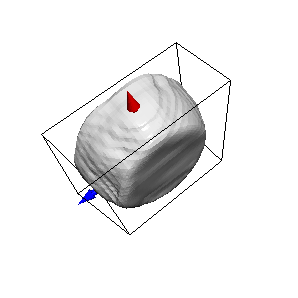}%
        \begin{tikzpicture}[overlay]%
            \draw[forestgreen4416044, very thick,contour=0.7mm] (0,0) rectangle ++(-0.09\textwidth,0.09\textwidth);%
        \end{tikzpicture}%
        \includegraphics[trim={0 0 0 0},clip,width=0.09\textwidth,height=0.09\textwidth]{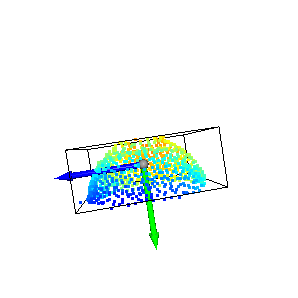}%
        \includegraphics[trim={0 0 0 0},clip,width=0.09\textwidth,height=0.09\textwidth]{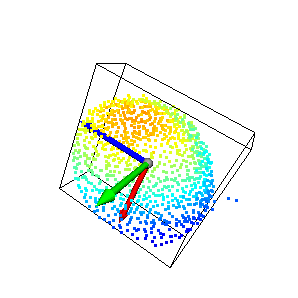}%
        
        \includegraphics[trim={0 0 0 0},clip,width=0.09\textwidth,height=0.09\textwidth]{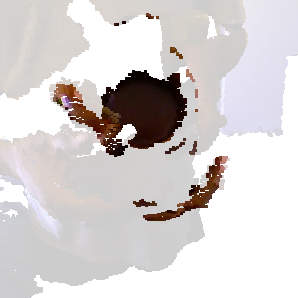}%
        \includegraphics[trim={0 0 0 0},clip,width=0.09\textwidth,height=0.09\textwidth]{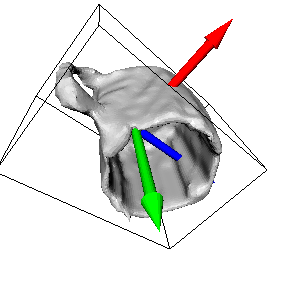}%
        \includegraphics[trim={0 0 0 0},clip,width=0.09\textwidth,height=0.09\textwidth]{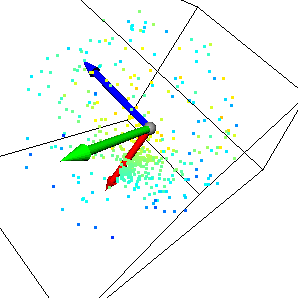}%
        \includegraphics[trim={0 0 0 0},clip,width=0.09\textwidth,height=0.09\textwidth]{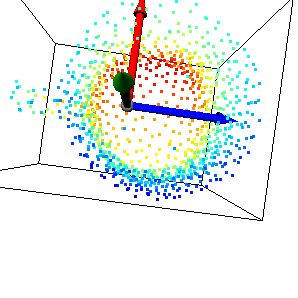}%
        \includegraphics[trim={0 0 0 0},clip,width=0.09\textwidth,height=0.09\textwidth]{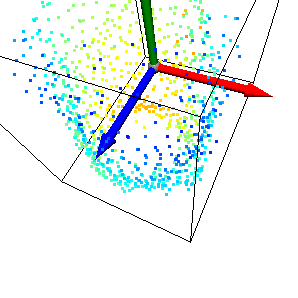}%
        \includegraphics[trim={0 0 0 0},clip,width=0.09\textwidth,height=0.09\textwidth]{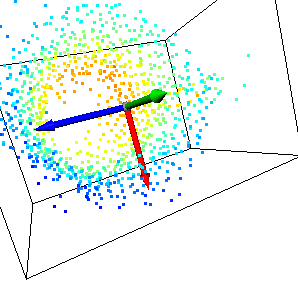}%
        \includegraphics[trim={0 0 0 0},clip,width=0.09\textwidth,height=0.09\textwidth]{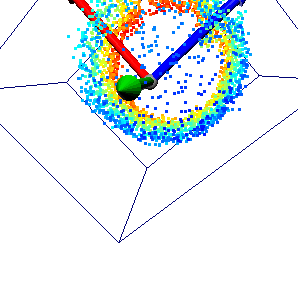}%
        \includegraphics[trim={0 0 0 0},clip,width=0.09\textwidth,height=0.09\textwidth]{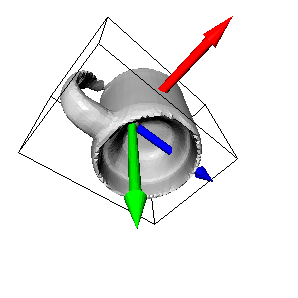}%
        \begin{tikzpicture}[overlay]%
            \draw[forestgreen4416044, very thick,contour=0.7mm] (0,0) rectangle ++(-0.09\textwidth,0.09\textwidth);%
        \end{tikzpicture}%
        \includegraphics[trim={0 0 0 0},clip,width=0.09\textwidth,height=0.09\textwidth]{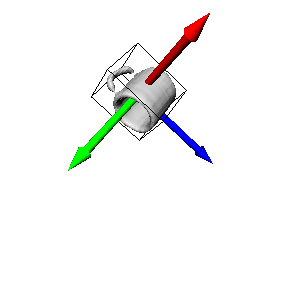}%
        \includegraphics[trim={0 0 0 0},clip,width=0.09\textwidth,height=0.09\textwidth]{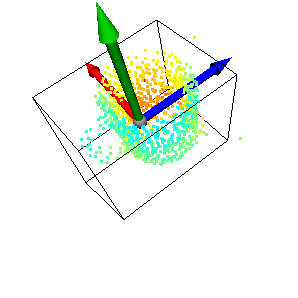}%
        \includegraphics[trim={0 0 0 0},clip,width=0.09\textwidth,height=0.09\textwidth]{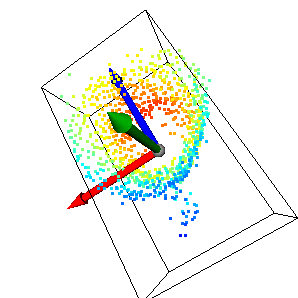}%
        
        \includegraphics[trim={0 0 0 0},clip,width=0.09\textwidth,height=0.09\textwidth]{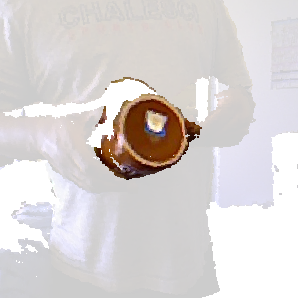}%
        \includegraphics[trim={0 0 0 0},clip,width=0.09\textwidth,height=0.09\textwidth]{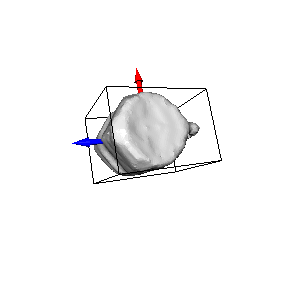}%
        \includegraphics[trim={0 0 0 0},clip,width=0.09\textwidth,height=0.09\textwidth]{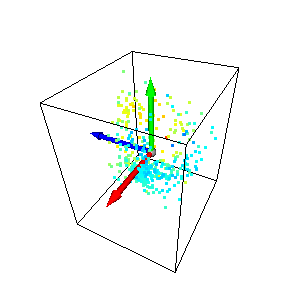}%
        \includegraphics[trim={0 0 0 0},clip,width=0.09\textwidth,height=0.09\textwidth]{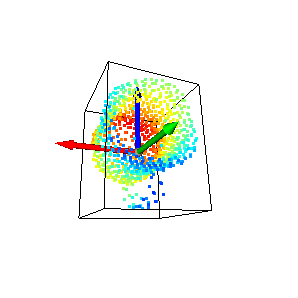}%
        \includegraphics[trim={0 0 0 0},clip,width=0.09\textwidth,height=0.09\textwidth]{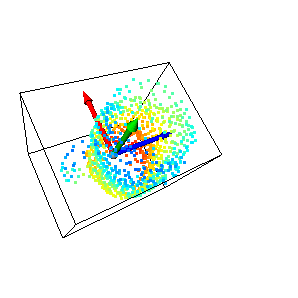}%
        \includegraphics[trim={0 0 0 0},clip,width=0.09\textwidth,height=0.09\textwidth]{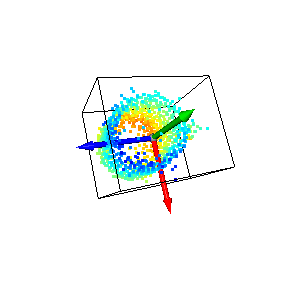}%
        \includegraphics[trim={0 0 0 0},clip,width=0.09\textwidth,height=0.09\textwidth]{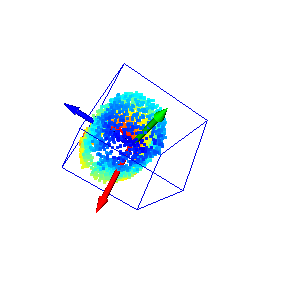}%
        \includegraphics[trim={0 0 0 0},clip,width=0.09\textwidth,height=0.09\textwidth]{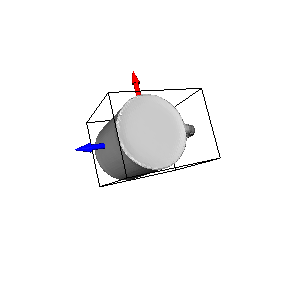}%
        \begin{tikzpicture}[overlay]%
            \draw[forestgreen4416044, very thick,contour=0.7mm] (0,0) rectangle ++(-0.09\textwidth,0.09\textwidth);%
        \end{tikzpicture}%
        \includegraphics[trim={0 0 0 0},clip,width=0.09\textwidth,height=0.09\textwidth]{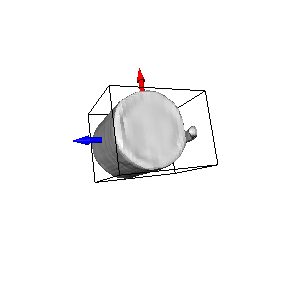}%
        \begin{tikzpicture}[overlay]%
            \draw[forestgreen4416044, very thick,contour=0.7mm] (0,0) rectangle ++(-0.09\textwidth,0.09\textwidth);%
        \end{tikzpicture}%
        \includegraphics[trim={0 0 0 0},clip,width=0.09\textwidth,height=0.09\textwidth]{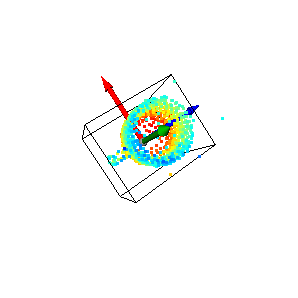}%
        \includegraphics[trim={0 0 0 0},clip,width=0.09\textwidth,height=0.09\textwidth]{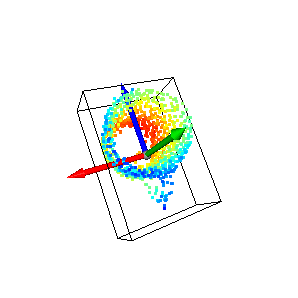}%
        \caption{REDWOOD75}\label{fig:qualitativer75}
    \end{subfigure}
    \setlength{\fboxsep}{2pt}
    \setlength{\fboxrule}{1pt}
    \caption{Randomly selected results on REAL275 and REDWOOD75 datasets. Results that are considered correct under $\delta=10\degree,d=\SI{2}{cm},F_{1\mathrm{cm}}=0.6$ thresholds are \protect\fcolorbox{forestgreen4416044}{white}{highlighted}. Input shows the full point set, with points outside the instance mask greyed out.}
    \label{fig:qualitative}
\end{figure*}

\Cref{fig:qualitative} shows randomly selected results on the REAL275 and REDWOOD75 datasets. For the REAL275 dataset, all methods perform pose estimation with a similar quality as shown in the respective publications. On REDWOOD75, on the other hand, only iCaps and SDFEst show generalization capability to arbitrary orientations. Only objects that are in a similar configuration as those in the REAL275 dataset are estimated successfully (see third row of \cref{fig:qualitativer75}) Most other methods predict upright objects consistent with the orientation distribution of REAL275 independent of the observed point sets.

A notable failure case is SDFEst's laptop prediction shown in \cref{fig:qualitativer275}. While the predicted shape aligns well, the pose is incorrect, that is, the keyboard is aligned with the screen and vice versa. This shows a downside of unbiased depth-only approaches. Similar cases were observed by \citet{deng2022icaps} for the can category, for which pure depth observations are not sufficient to determine whether the can is upright or not without introducing a bias into the estimation. In general, we observe more severe outliers produced by the unbiased methods, iCaps and SDFEst, on the REAL275 dataset.

ASM-Net's reconstructions are often flipped, but surfaces typically align well. As noted above, CASS completely fails to reconstruct any object except laptops. Most point set-based methods and, in particular, DPDN predict a few outlier points (outside the estimated bounding box) in their reconstruction.

\subsection{Quantitative Results}\label{sec:quantitative}

\begin{table*}[htb]
    \centering
    \caption{Precision at varying position, orientation and F-score thresholds. The \textbf{best} \protect\gold, second best \protect\silver, and third best~\protect\bronze\ are highlighted.}\label{tab:precision}
    
    \renewcommand{\arraystretch}{1.1}
    \scriptsize
    
    \begin{tabular}{
      @{}
      l
      S[table-format=1.4]@{\hspace{0.3\tabcolsep}}c
      S[table-format=1.4]@{\hspace{0.3\tabcolsep}}c
      S[table-format=1.4]@{\hspace{0.3\tabcolsep}}c
      S[table-format=1.4]@{\hspace{0.3\tabcolsep}}c
      @{}c
      S[table-format=1.4]@{\hspace{0.3\tabcolsep}}c
      S[table-format=1.4]@{\hspace{0.3\tabcolsep}}c
      S[table-format=1.4]@{\hspace{0.3\tabcolsep}}c
      S[table-format=1.4]@{\hspace{0.3\tabcolsep}}c
    }
        \toprule
         & \multicolumn{8}{c}{REAL275} & & \multicolumn{8}{c}{REDWOOD75} \\
         \cmidrule(lr){2-9} \cmidrule(lr){11-18}
         \multicolumn{1}{r}{$\delta,d \rightarrow\hspace{-0.3cm}$} & \multicolumn{4}{c}{$10\degree,\SI{2}{cm}$} &  \multicolumn{4}{c}{$5\degree,\SI{1}{cm}$} & & \multicolumn{4}{c}{$10\degree,\SI{2}{cm}$} &  \multicolumn{4}{c}{$5\degree,\SI{1}{cm}$} \\
         \cmidrule(lr){2-5} \cmidrule(lr){6-9} \cmidrule(lr){11-14} \cmidrule(lr){15-18}
         \multicolumn{1}{r}{$F_{1\mathrm{cm}}\rightarrow\hspace{-0.3cm}$}& \multicolumn{2}{c}{$\infty$} & \multicolumn{2}{c}{$0.6$} & \multicolumn{2}{c}{$\infty$} &  \multicolumn{2}{c}{$0.8$} & & \multicolumn{2}{c}{$\infty$} & \multicolumn{2}{c}{$0.6$} & \multicolumn{2}{c}{$\infty$} &  \multicolumn{2}{c}{$0.8$} \\
        \midrule
        {CASS} & 0.3319 & & 0.0293 & & 0.0723 & & 0.0000 & & & 0.0133 & & 0.0000 & & 0.0000 & & 0.0000 & \\
        {SPD} & 0.5439 & & 0.4818 & & 0.2082 & & 0.1741 & & & 0.1867 & & 0.1600 & & 0.0400 & & 0.0267 & \\
        {CR-Net} & 0.5993 & & 0.4781 & & 0.2397 & & 0.1777 & & & 0.2533 & & 0.2400 & \bronze & 0.0800 & & 0.0800 & \\
        {SGPA} & 0.6939 & \bronze & 0.6302 & \silver & 0.3142 & \silver & 0.2497 & \silver & & 0.2267 & & 0.2267 & & 0.2000 & \silver & 0.1733 & \silver \\
        {ASM-Net} & 0.3255 & & 0.2101 & & 0.0678 & & 0.0508 & & & 0.3067 & & 0.1733 & & 0.0667 & & 0.0533 & \\
        {iCaps} & 0.2976 & & 0.2243 & & 0.0565 & & 0.0379 & & & 0.4133 & \silver & 0.3467 & \silver & 0.1200 & \bronze & 0.0800 & \\
        {SDFEst} & 0.5060 & & 0.4710 & & 0.2237 & & 0.2008 & \bronze & & \bfseries 0.6533 & \gold & \bfseries 0.6400 & \gold & \bfseries 0.4667 & \gold & \bfseries 0.4133 & \gold \\
        {DPDN} & \bfseries 0.7589 & \gold & \bfseries 0.6315 & \gold & \bfseries 0.3336 & \gold & \bfseries 0.2591 & \gold & & 0.3333 & \bronze & 0.3200 & \bronze & 0.0933 & & 0.0933 & \bronze \\
        {RBP-Pose} & 0.7397 & \silver & 0.5559 & \bronze & 0.2847 & \bronze & 0.1514 & & & 0.2133 & & 0.0933 & & 0.0667 & & 0.0267 & \\

        \bottomrule
    \end{tabular}
\end{table*}

\begin{table*}[htb]
    \centering
    \caption{Precision for different categories with $\delta=10\degree,d=\SI{2}{cm},F_{1\mathrm{cm}}=0.6$. The \textbf{best} \protect\gold, second best \protect\silver, and third best~\protect\bronze\ are highlighted.}\label{tab:categories}
    
    \renewcommand{\arraystretch}{1.1}
    \scriptsize
    
    \begin{tabular}{
      @{}
      l
      S[table-format=1.4]@{\hspace{0.3\tabcolsep}}c
      S[table-format=1.4]@{\hspace{0.3\tabcolsep}}c
      S[table-format=1.4]@{\hspace{0.3\tabcolsep}}c
      S[table-format=1.4]@{\hspace{0.3\tabcolsep}}c
      S[table-format=1.4]@{\hspace{0.3\tabcolsep}}c
      S[table-format=1.4]@{\hspace{0.3\tabcolsep}}c
      @{}c
      S[table-format=1.4]@{\hspace{0.3\tabcolsep}}c
      S[table-format=1.4]@{\hspace{0.3\tabcolsep}}c
      S[table-format=1.4]@{\hspace{0.3\tabcolsep}}c
      @{}
    }
        \toprule
        & \multicolumn{12}{c}{REAL275} & & \multicolumn{6}{c}{REDWOOD75} \\
        \cmidrule(lr){2-13} \cmidrule(lr){15-20}
        & \multicolumn{2}{c}{Bottle} & \multicolumn{2}{c}{Bowl} & \multicolumn{2}{c}{Camera} & \multicolumn{2}{c}{Can} & \multicolumn{2}{c}{Laptop} & \multicolumn{2}{c}{Mug} & & \multicolumn{2}{c}{Bottle} & \multicolumn{2}{c}{Bowl} & \multicolumn{2}{c}{Mug} \\
        \midrule
        {CASS} & 0.0018 & & 0.0920 & & 0.0000 & & 0.0264 & & 0.0000 & & 0.0623 & & & 0.0000 & & 0.0000 & & 0.0000 \\
        {SPD} & 0.6268 & \bronze & 0.8987 & & 0.0543 & & \bfseries 0.8560 & \gold & 0.2269 & & 0.2825 & & & 0.3200 & \bronze & 0.0800 & & 0.0800 \\
        {CR-Net} & 0.5963 & & 0.9131 & & 0.0498 & & 0.6021 & & 0.3936 & \silver & 0.3656 & & & 0.2400 & & 0.4000 & \bronze & 0.0800 \\
        {SGPA} & \bfseries 0.7570 & \gold & 0.9481 & \silver & 0.0742 & \bronze & 0.8534 & \silver & \bfseries 0.4383 & \gold & 0.7362 & & & 0.2000 & & 0.4000 & \bronze & 0.0800 \\
        {ASM-Net} & 0.1596 & & 0.1114 & & 0.0229 & & 0.5808 & & 0.1365 & & 0.2328 & & & 0.1600 & & 0.3600 & & 0.0000 \\
        {iCaps} & 0.2136 & & 0.5477 & & 0.0214 & & 0.2257 & & 0.1187 & & 0.2568 & & & 0.1600 & & 0.7600 & \silver & 0.1200 & \bronze \\
        {SDFEst} & 0.3935 & & 0.8679 & & 0.0011 & & 0.4984 & & 0.1946 & & \bfseries 0.8986 & \gold & & \bfseries 0.7200 & \gold & \bfseries 0.8400 & \gold & \bfseries 0.3600 & \gold \\
        {DPDN} & 0.7004 & \silver & \bfseries 0.9582 & \gold & 0.1562 & \silver & 0.8153 & \bronze & 0.3137 & \bronze & 0.8704 & \silver & & 0.4000 & \silver & 0.4000 & \bronze & 0.1600 & \silver \\
        {RBP-Pose} & 0.4559 & & 0.9338 & \bronze & \bfseries 0.2023 & \gold & 0.7340 & & 0.2756 & & 0.7707 & \bronze & & 0.0000 & & 0.2000 & & 0.0800 \\
         \bottomrule
    \end{tabular}
\end{table*}

\begin{figure}[htb]
    \centering
    \begin{tikzpicture}

\definecolor{crimson2143940}{RGB}{214,39,40}
\definecolor{darkgray176}{RGB}{176,176,176}
\definecolor{darkorange25512714}{RGB}{255,127,14}
\definecolor{forestgreen4416044}{RGB}{44,160,44}
\definecolor{darkturquoise23190207}{RGB}{23,190,207}
\definecolor{gray127}{RGB}{127,127,127}
\definecolor{lightgray204}{RGB}{204,204,204}
\definecolor{mediumpurple148103189}{RGB}{148,103,189}
\definecolor{orchid227119194}{RGB}{227,119,194}
\definecolor{sienna1408675}{RGB}{140,86,75}
\definecolor{steelblue31119180}{RGB}{31,119,180}

\scriptsize

\pgfplotsset{colormap name=viridis}

\begin{axis}[
    width=0.7\linewidth,
    height=0.7\linewidth,
    axis equal image,
    scale only axis=true,
    tick align=outside,
    tick pos=left,
    xlabel={REAL275},
    xmajorgrids,
    xmin=-0.05, xmax=0.85,
    x grid style={darkgray176},
    ylabel={REDWOOD75},
    ymajorgrids,
    ymin=-0.05, ymax=0.85,
    y grid style={darkgray176},
]
    \addplot[
        scatter, mark=*, only marks, 
        scatter src=explicit,
        clip=false,
        fill opacity=0.5,text opacity=1,
        visualization depends on={value \thisrow{xshift} \as \XShift},
        visualization depends on={value \thisrow{yshift} \as \YShift},
        visualization depends on={value \thisrow{anchor} \as \Anchor},
        visualization depends on={value \thisrow{label} \as \Label},
        point meta=\thisrow{color},
        visualization depends on = {2*\thisrow{radius} \as \radius}, 
        nodes near coords*={\Label},
        nodes near coords style={anchor=\Anchor,xshift=\XShift,yshift=\YShift},
        scatter/@pre marker code/.append style={
          /tikz/mark size=\radius
        }
    ] table {
        x y label radius anchor xshift yshift color
        0.0293 0.0 CASS 2 {south west} 0mm 0mm 0.1592
        0.4818 0.16 SPD 2 {west} 1mm 0mm 0.547257144397326
        0.4781 0.24 CR-Net 2 {south} 0mm 1mm 0.545682040308947
        0.6302 0.2267 SGPA 2 {west} 1mm 0mm 0.5478021441082043
        0.2101 0.1733 ASM-Net 2 {south west} 0mm 0mm 0.42511975665490076
        0.2243 0.3467 iCaps 2 {south west} 0mm 0mm 0.963893880319144
        0.4710 0.6400 SDFEst 2 {south west} 0mm 0mm 0.8565464684305338
        0.6315 0.3200 DPDN 2 {south west} 0mm 0mm 0.5650689573717497
        0.5559 0.0933 RBP-Pose 2 {north west} 0mm 0mm 0.49724917213131886
        0.9 0.8 $3\,\mathrm{s}$ 2 west 1mm 0mm 1.0
        0.9 0.72 $1\,\mathrm{s}$ 2 west 1mm 0mm 0.807
        0.9 0.64 $300\,\mathrm{ms}$ 2 west 1mm 0mm 0.596
        0.9 0.56 $100\,\mathrm{ms}$ 2 west 1mm 0mm 0.404
        0.9 0.48 $30\,\mathrm{ms}$ 2 west 1mm 0mm 0.193
        0.9 0.4 $10\,\mathrm{ms}$ 2 west 1mm 0mm 0.03123
    };
\end{axis}
\end{tikzpicture}
    \caption{Performance of the evaluated methods on the REAL275 and REDWOOD75 dataset for $\delta=10\degree,d=\SI{2}{cm},F_{1\mathrm{cm}}=0.6$. The color of the circle indicates the run time of the method for one inference.}
    \label{fig:overview}
\end{figure}
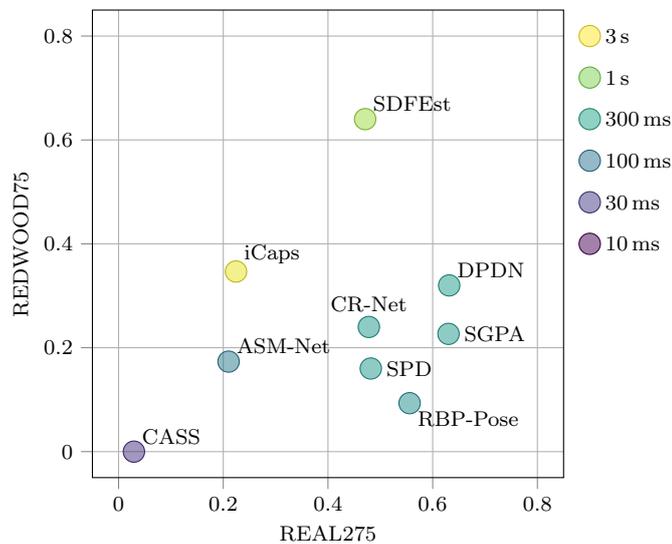

We now present the results using the metrics introduced in Section \ref{sec:metrics}. Following \cref{sec:metrics_summary}, we report precision with varying thresholds on translation error, orientation error, and reconstruction F-score in \cref{tab:precision}. To assess the joint estimate of pose and shape, we use $5\degree, \SI{1}{cm}, 0.8$ and $10\degree, \SI{2}{cm}, 0.6$ for $\delta$, $d$, and $F_{1\mathrm{cm}}$, respectively. To assess pose estimation in isolation (i.e., $F_{1\mathrm{cm}}=\infty$), we use $5\degree, \SI{1}{cm}$ and $10\degree, \SI{2}{cm}$. We picked these tuples of thresholds such that all thresholds in a tuple are roughly equally strict. In the past, some methods used pairs such as $10\degree, 5\mathrm{cm}$, or $10\degree, 10\mathrm{cm}$, where only the $10\degree$ threshold practically mattered.

The results from Table \ref{tab:precision} confirm the qualitative observations from before. DPDN performs best on REAL275, and SDFEst performs best on REDWOOD75. In general, the unbiased methods, SDFEst and iCaps, perform better on REDWOOD75 confirming limited generalization capability of methods trained on the CAMERA and REAL datasets. However, improved generalization performance comes with poorer performance on the REAL275 dataset, which opens the question, whether strong cross-dataset methods are possible. DPDN shows the third best performance on REDWOOD75, however, we note that this mostly stems from objects with similar orientations as REAL275. RBP-Pose while giving strong results for pose estimation on REAL275, shows poor performance on REDWOOD75 indicating a strong dataset-specific bias. Furthermore, its reconstruction quality falls short of simpler methods such as SGPA.

Note that for both datasets there is still a lot of room for improvement. Typically, significantly less than 50\% of the estimates are of sufficient quality to be considered correct in pose and shape for the stricter thresholds. This shows that categorical pose and shape estimation is still an open problem, especially for unconstrained orientations.

\begin{figure*}[htb]
    \centering
    \scriptsize
    \begin{subfigure}{\textwidth}
        \centering%
        \begin{tikzpicture}

\definecolor{crimson2143940}{RGB}{214,39,40}
\definecolor{darkgray176}{RGB}{176,176,176}
\definecolor{darkorange25512714}{RGB}{255,127,14}
\definecolor{forestgreen4416044}{RGB}{44,160,44}
\definecolor{darkturquoise23190207}{RGB}{23,190,207}
\definecolor{gray127}{RGB}{127,127,127}
\definecolor{lightgray204}{RGB}{204,204,204}
\definecolor{mediumpurple148103189}{RGB}{148,103,189}
\definecolor{orchid227119194}{RGB}{227,119,194}
\definecolor{sienna1408675}{RGB}{140,86,75}
\definecolor{steelblue31119180}{RGB}{31,119,180}

\begin{axis}[
width=0.24\textwidth,
height=0.2\textwidth,
scale only axis=true,
tick align=outside,
tick pos=left,
x grid style={darkgray176},
xlabel={$d\vphantom{ / \mathrm{cm}}$},
xmajorgrids,
xmin=-0.0025, xmax=0.0525,
xtick style={color=black},
xticklabel=$\mathclap{\pgfmathprintnumber{\tick}}\vphantom{\degree{}}$,
y grid style={darkgray176},
ymajorgrids,
ymin=-0.05, ymax=1.05,
ytick style={color=black},
change x base,
unit markings=slash space,
x SI prefix=centi,x unit=m,
every x tick scale label/.append style={overlay}
]
\addplot [very thick, steelblue31119180]
table {%
0 0
0.005 0.037969971460479
0.01 0.189167390495099
0.015 0.379017247797493
0.02 0.549385779873433
0.025 0.692207469909418
0.03 0.795756297307358
0.035 0.873929767961286
0.04 0.923749844893907
0.045 0.951048517185755
0.05 0.968792654175456
};

\addplot [very thick, darkorange25512714]
table {%
0 0
0.005 0.113165405137114
0.01 0.431753319270381
0.015 0.658642511477851
0.02 0.776709269140092
0.025 0.853393721305373
0.03 0.897133639409356
0.035 0.924742523886338
0.04 0.944347933986847
0.045 0.955453530214667
0.05 0.962154113413575
};

\addplot [very thick, forestgreen4416044]
table {%
0 0
0.005 0.095297183273359
0.01 0.358419158704554
0.015 0.612358853455764
0.02 0.750341233403648
0.025 0.82963146792406
0.03 0.892914753691525
0.035 0.937275096165777
0.04 0.95942424618439
0.045 0.968854696612483
0.05 0.977416552922199
};

\addplot [very thick, crimson2143940]
table {%
0 0
0.005 0.146916490879762
0.01 0.503412334036481
0.015 0.720809033378831
0.02 0.829135128427845
0.025 0.883794515448567
0.03 0.916304752450676
0.035 0.940935599950366
0.04 0.959920585680606
0.045 0.971956818463829
0.05 0.978099019729495
};

\addplot [very thick, mediumpurple148103189]
table {%
0 0
0.005 0.0259957811142822
0.01 0.129854820697357
0.015 0.306117384290855
0.02 0.518984985730239
0.025 0.688174711502668
0.03 0.798610249410597
0.035 0.866732845266162
0.04 0.90594366546718
0.045 0.925052736071473
0.05 0.941928278942797
};

\addplot [very thick, sienna1408675]
table {%
0 0
0.005 0.0434917483558754
0.01 0.185320759399429
0.015 0.371261943169128
0.02 0.535426231542375
0.025 0.664474500558382
0.03 0.76504529097903
0.035 0.836207966248914
0.04 0.876907804938578
0.045 0.903213798237995
0.05 0.918352152872565
};

\addplot [very thick, orchid227119194]
table {%
0 0
0.005 0.0702320387144807
0.01 0.319580593125698
0.015 0.552736071472888
0.02 0.675456011912148
0.025 0.753009058195806
0.03 0.817657277577863
0.035 0.871944409976424
0.04 0.914939818836084
0.045 0.94509244323117
0.05 0.961471646606279
};

\addplot [very thick, gray127]
table {%
0 0
0.005 0.152686437523266
0.01 0.508623898746743
0.015 0.765665715349299
0.02 0.887144807048021
0.025 0.930202258344708
0.03 0.949559498697109
0.035 0.960789179798982
0.04 0.96569053232411
0.045 0.972887455019233
0.05 0.981077056706787
};

\addplot [very thick, darkturquoise23190207]
table {%
0 0
0.005 0.116205484551433
0.01 0.477664722670306
0.015 0.719133887579104
0.02 0.836518178434049
0.025 0.888013401166398
0.03 0.914691649087976
0.035 0.936344459610373
0.04 0.953220002481697
0.045 0.966621168879514
0.05 0.976113661744633
};

\end{axis}%
\end{tikzpicture}%
%
        \begin{tikzpicture}

\definecolor{crimson2143940}{RGB}{214,39,40}
\definecolor{darkgray176}{RGB}{176,176,176}
\definecolor{darkorange25512714}{RGB}{255,127,14}
\definecolor{forestgreen4416044}{RGB}{44,160,44}
\definecolor{darkturquoise23190207}{RGB}{23,190,207}
\definecolor{gray127}{RGB}{127,127,127}
\definecolor{lightgray204}{RGB}{204,204,204}
\definecolor{mediumpurple148103189}{RGB}{148,103,189}
\definecolor{orchid227119194}{RGB}{227,119,194}
\definecolor{sienna1408675}{RGB}{140,86,75}
\definecolor{steelblue31119180}{RGB}{31,119,180}

\begin{axis}[
width=0.24\textwidth,
height=0.2\textwidth,
scale only axis=true,
tick align=outside,
tick pos=left,
x grid style={darkgray176},
xlabel={$\delta\vphantom{ / \mathrm{cm}}$},
xticklabel=$\mathclap{\pgfmathprintnumber{\tick}\degree{}}$,
xmajorgrids,
xmin=-1, xmax=21,
xtick style={color=black},
y grid style={darkgray176},
ymajorgrids,
ymin=-0.05, ymax=1.05,
ytick style={color=black}
]
\addplot [very thick, steelblue31119180]
table {%
0 0
1 0.00626628613971957
2 0.0272986722918476
3 0.0584439756793647
4 0.108015882863879
5 0.171857550564586
6 0.230983993051247
7 0.285643380071969
8 0.335463457004591
9 0.38292592133019
10 0.427472391115523
11 0.466310956694379
12 0.504653182777019
13 0.534743764735079
14 0.562662861397196
15 0.587107581585805
16 0.610993919841171
17 0.633577366918973
18 0.657773917359474
19 0.680915746370517
20 0.701265665715349
};

\addplot [very thick, darkorange25512714]
table {%
0 0
1 0.0279811390991438
2 0.0970343715101129
3 0.178061794267279
4 0.26423873929768
5 0.349360962898623
6 0.4218885717831
7 0.489142573520288
8 0.549820076932622
9 0.594738801340117
10 0.634011663978161
11 0.665777391735947
12 0.694316912768334
13 0.719444099764239
14 0.740600570790421
15 0.760516193076064
16 0.779066881747115
17 0.796190594366547
18 0.811825288497332
19 0.824916242710014
20 0.836456135997022
};

\addplot [very thick, forestgreen4416044]
table {%
0 0
1 0.0390246928899367
2 0.144993175331927
3 0.280866112420896
4 0.39967737932746
5 0.492244695371634
6 0.571845142077181
7 0.629048269016007
8 0.663543864002978
9 0.691400918228068
10 0.714542747239112
11 0.733217520784216
12 0.748790172477975
13 0.763059932994168
14 0.775778632584688
15 0.786077677131158
16 0.795756297307358
17 0.804069983868966
18 0.810336270008686
19 0.816850725896513
20 0.822124333043802
};

\addplot [very thick, crimson2143940]
table {%
0 0
1 0.0392728626380444
2 0.15523017744137
3 0.295384042685197
4 0.421206104975803
5 0.522707531951855
6 0.60578235513091
7 0.669810150142698
8 0.714915001861273
9 0.746308474996898
10 0.770318898126318
11 0.791599454026554
12 0.808599081771932
13 0.823054969599206
14 0.835587541878645
15 0.848058071721057
16 0.85910162551185
17 0.869586797369401
18 0.877900483931009
19 0.885966000744509
20 0.893659262935848
};

\addplot [very thick, mediumpurple148103189]
table {%
0 0
1 0.033751085742648
2 0.119990073210076
3 0.223414815733962
4 0.324171733465691
5 0.416490879761757
6 0.493485544112173
7 0.540761881126691
8 0.571783099640154
9 0.594614716466063
10 0.615709145055218
11 0.633143069859784
12 0.65057699466435
13 0.665839434172974
14 0.683831740910783
15 0.70387144807048
16 0.724717706911527
17 0.740476485916367
18 0.753629482566075
19 0.763494230053356
20 0.772862638044422
};

\addplot [very thick, sienna1408675]
table {%
0 0
1 0.0142697605161931
2 0.0496339496215411
3 0.096103734954709
4 0.149212061049758
5 0.19822558630103
6 0.250589403151756
7 0.301091946891674
8 0.35103610869835
9 0.396885469661248
10 0.44025313314307
11 0.48082888695868
12 0.517744136989701
13 0.545104851718575
14 0.572217396699342
15 0.594490631592009
16 0.610187368159821
17 0.622471770691153
18 0.630102990445465
19 0.638230549695992
20 0.642945774910038
};

\addplot [very thick, orchid227119194]
table {%
0 0
1 0.0374736319642636
2 0.140464077428961
3 0.274165529221988
4 0.38534557637424
5 0.464201513835463
6 0.516565330686189
7 0.555900235761261
8 0.574450924432312
9 0.584749968978782
10 0.59225710385904
11 0.595173098399305
12 0.599826281176325
13 0.604541506390371
14 0.607829755552798
15 0.611304132026306
16 0.614282169003598
17 0.616949993795756
18 0.619121479091699
19 0.621975431194937
20 0.624208958927907
};

\addplot [very thick, gray127]
table {%
0 0
1 0.0436778756669562
2 0.169251768209455
3 0.326777515820821
4 0.482938329817595
5 0.611924556396575
6 0.703499193448319
7 0.766720436778757
8 0.802829135128428
9 0.826529346072714
10 0.845824543988088
11 0.862451917111304
12 0.878955205360467
13 0.892914753691525
14 0.904082392356372
15 0.913947139843653
16 0.922322868842288
17 0.929830003722546
18 0.936778756669562
19 0.941742151631716
20 0.94670554659387
};

\addplot [very thick, darkturquoise23190207]
table {%
0 0
1 0.0402655416304752
2 0.166025561484055
3 0.321379823799479
4 0.467551805434917
5 0.586301029904455
6 0.686499565702941
7 0.759151259461472
8 0.810522397319767
9 0.849919344831865
10 0.881436902841544
11 0.903896265045291
12 0.921950614220127
13 0.936530586921454
14 0.947636183149274
15 0.955205360466559
16 0.960851222236009
17 0.964946023079787
18 0.968730611738429
19 0.971894776026802
20 0.974128303759772
};
\end{axis}%
\end{tikzpicture}%
%
        \begin{tikzpicture}

\definecolor{crimson2143940}{RGB}{214,39,40}
\definecolor{darkgray176}{RGB}{176,176,176}
\definecolor{darkorange25512714}{RGB}{255,127,14}
\definecolor{forestgreen4416044}{RGB}{44,160,44}
\definecolor{darkturquoise23190207}{RGB}{23,190,207}
\definecolor{gray127}{RGB}{127,127,127}
\definecolor{lightgray204}{RGB}{204,204,204}
\definecolor{mediumpurple148103189}{RGB}{148,103,189}
\definecolor{orchid227119194}{RGB}{227,119,194}
\definecolor{sienna1408675}{RGB}{140,86,75}
\definecolor{steelblue31119180}{RGB}{31,119,180}

\begin{axis}[
width=0.24\textwidth,
height=0.2\textwidth,
scale only axis=true,
tick align=outside,
tick pos=left,
x grid style={darkgray176},
xticklabel=$\mathclap{\pgfmathprintnumber{\tick}}\vphantom{\degree{}}$,
xlabel={$F_{1\mathrm{cm}}\vphantom{ / \mathrm{cm}}$},
xmajorgrids,
xmin=-0.05, xmax=1.05,
xtick style={color=black},
y grid style={darkgray176},
ymajorgrids,
ymin=-0.05, ymax=1.05,
ytick style={color=black}
]
\addplot [very thick, steelblue31119180]
table {%
0 1
0.05 0.94105968482442
0.1 0.909231914629607
0.15 0.872750961657774
0.2 0.813500434297059
0.25 0.722980518674774
0.3 0.636555403896265
0.35 0.532324109691029
0.4 0.418103983124457
0.4 0.418103983124457
0.45 0.323489266658394
0.5 0.241965504405013
0.55 0.140774289614096
0.6 0.0627869462712495
0.65 0.023638168507259
0.7 0.00465318277701948
0.75 0.000310212185134632
0.8 0
0.85 0
0.9 0
0.95 0
1 0
};

\addplot [very thick, darkorange25512714]
table {%
0 1
0.05 0.991996525623526
0.1 0.986722918476238
0.15 0.977292468048145
0.2 0.963705174339248
0.25 0.942920957935228
0.3 0.92430822682715
0.35 0.895148281424494
0.4 0.852090830127807
0.4 0.852090830127807
0.45 0.804007941431939
0.5 0.751333912396079
0.55 0.70225834470778
0.6 0.653741158952724
0.65 0.603983124457129
0.7 0.53834222608264
0.75 0.466931381064648
0.8 0.399429209579352
0.85 0.32721181288001
0.9 0.240166273731232
0.95 0.114033999255491
1 0.00111676386648468
};

\addplot [very thick, forestgreen4416044]
table {%
0 1
0.05 0.995594986971088
0.1 0.993857798734334
0.15 0.989080531083261
0.2 0.975927534433553
0.25 0.960851222236009
0.3 0.942610745750093
0.35 0.913140588162303
0.4 0.870083136865616
0.4 0.870083136865616
0.45 0.80742027546842
0.5 0.740104231294205
0.55 0.667080282913513
0.6 0.584501799230674
0.65 0.507693262191339
0.7 0.435724035240104
0.75 0.366856930140216
0.8 0.313934731356248
0.85 0.259213301898499
0.9 0.191400918228068
0.95 0.100074450924432
1 0.000248169748107706
};

\addplot [very thick, crimson2143940]
table {%
0 1
0.05 0.994043926045415
0.1 0.99212061049758
0.15 0.990197294949746
0.2 0.985792281920834
0.25 0.971584563841668
0.3 0.96345700459114
0.35 0.949621541134136
0.4 0.920213425983373
0.4 0.920213425983373
0.45 0.883422260826405
0.5 0.837821069611614
0.55 0.785395210323862
0.6 0.734210199776647
0.65 0.685196674525375
0.7 0.633639409355999
0.75 0.561111800471523
0.8 0.482007693262191
0.85 0.403648095297183
0.9 0.332175207842164
0.95 0.215101129172354
1 0.000682466807296191
};

\addplot [very thick, mediumpurple148103189]
table {%
0 1
0.05 0.981387268891922
0.1 0.973818091574637
0.15 0.965194192827894
0.2 0.954398808785209
0.25 0.9267278818712
0.3 0.89564462092071
0.35 0.847437647350788
0.4 0.75877900483931
0.4 0.75877900483931
0.45 0.654485668197047
0.5 0.549013525251272
0.55 0.440811515076312
0.6 0.354200272986723
0.65 0.274785953592257
0.7 0.206663357736692
0.75 0.157960044670555
0.8 0.112855192951979
0.85 0.0777391735947388
0.9 0.0526740290358605
0.95 0.0358605286015635
1 0.000620424370269264
};

\addplot [very thick, sienna1408675]
table {%
0 1
0.05 0.928775282293088
0.1 0.924866608760392
0.15 0.916056582702569
0.2 0.901972949497456
0.25 0.882057327211813
0.3 0.852463084749969
0.35 0.809343591016255
0.4 0.744633329197171
0.4 0.744633329197171
0.45 0.657960044670555
0.5 0.578049385779873
0.55 0.507321007569177
0.6 0.445030400794143
0.65 0.381002605782355
0.7 0.318649956570294
0.75 0.257041816602556
0.8 0.192703809405633
0.85 0.135872937088969
0.9 0.0858046904082392
0.95 0.0364809529718327
1 0.00223352773296935
};

\addplot [very thick, orchid227119194]
table {%
0 1
0.05 0.994974562600819
0.1 0.991624271001365
0.15 0.98324854200273
0.2 0.973197667204368
0.25 0.954336766348182
0.3 0.931256979774166
0.35 0.907060429333664
0.4 0.87665963519047
0.4 0.87665963519047
0.45 0.841047276337014
0.5 0.795446085122224
0.55 0.741841419530959
0.6 0.677317285022956
0.65 0.609691028663606
0.7 0.534247425238863
0.75 0.461533689043306
0.8 0.401600694875295
0.85 0.345638416677007
0.9 0.281548579228192
0.95 0.197853331678868
1 0.02947015758779
};

\addplot [very thick, gray127]
table {%
0 1
0.05 0.995470902097034
0.1 0.993795756297307
0.15 0.991252016379203
0.2 0.987095173098399
0.25 0.977664722670306
0.3 0.968048144931133
0.35 0.953406129792778
0.4 0.921206104975803
0.4 0.921206104975803
0.45 0.873929767961286
0.5 0.822806799851098
0.55 0.769202134259834
0.6 0.709641394713984
0.65 0.634818215659511
0.7 0.563903710137734
0.75 0.498448939074327
0.8 0.430450428092815
0.85 0.355751333912396
0.9 0.274227571659015
0.95 0.170244447201886
1 6.20424370269264e-05
};

\addplot [very thick, darkturquoise23190207]
table {%
0 1
0.05 0.999813872688919
0.1 0.998262811763246
0.15 0.99454026554163
0.2 0.978843528973818
0.25 0.964821938205733
0.3 0.948194565082516
0.35 0.924680481449311
0.4 0.869276585184266
0.4 0.869276585184266
0.45 0.802394838069239
0.5 0.739918103983124
0.55 0.68308723166646
0.6 0.619679861024941
0.65 0.546655912644249
0.7 0.465380320138975
0.75 0.380878520908301
0.8 0.28421640402035
0.85 0.212371261943169
0.9 0.146482193820573
0.95 0.0527981139099144
1 0
};
\end{axis}%
\end{tikzpicture}%
%
        \caption{REAL275}
    \end{subfigure}
    \begin{subfigure}{\textwidth}
        \centering%
        \begin{tikzpicture}

\definecolor{crimson2143940}{RGB}{214,39,40}
\definecolor{darkgray176}{RGB}{176,176,176}
\definecolor{darkorange25512714}{RGB}{255,127,14}
\definecolor{forestgreen4416044}{RGB}{44,160,44}
\definecolor{darkturquoise23190207}{RGB}{23,190,207}
\definecolor{gray127}{RGB}{127,127,127}
\definecolor{lightgray204}{RGB}{204,204,204}
\definecolor{mediumpurple148103189}{RGB}{148,103,189}
\definecolor{orchid227119194}{RGB}{227,119,194}
\definecolor{sienna1408675}{RGB}{140,86,75}
\definecolor{steelblue31119180}{RGB}{31,119,180}

\begin{axis}[
width=0.24\textwidth,
height=0.2\textwidth,
scale only axis=true,
tick align=outside,
tick pos=left,
x grid style={darkgray176},
xlabel={$d\vphantom{ / \mathrm{cm}}$},
xmajorgrids,
xmin=-0.0025, xmax=0.0525,
xtick style={color=black},
xticklabel=$\mathclap{\pgfmathprintnumber{\tick}}\vphantom{\degree{}}$,
y grid style={darkgray176},
ymajorgrids,
ymin=-0.05, ymax=1.05,
ytick style={color=black},
change x base,
unit markings=slash space,
x SI prefix=centi,x unit=m,
every x tick scale label/.append style={overlay}
]
\addplot [very thick, steelblue31119180]
table {%
0 0
0.005 0
0.01 0.0266666666666667
0.015 0.186666666666667
0.02 0.293333333333333
0.025 0.426666666666667
0.03 0.56
0.035 0.746666666666667
0.04 0.853333333333333
0.045 0.88
0.05 0.906666666666667
};

\addplot [very thick, darkorange25512714]
table {%
0 0
0.005 0.0133333333333333
0.01 0.08
0.015 0.226666666666667
0.02 0.4
0.025 0.493333333333333
0.03 0.613333333333333
0.035 0.693333333333333
0.04 0.76
0.045 0.84
0.05 0.866666666666667
};

\addplot [very thick, forestgreen4416044]
table {%
0 0
0.005 0.0933333333333333
0.01 0.253333333333333
0.015 0.32
0.02 0.493333333333333
0.025 0.613333333333333
0.03 0.666666666666667
0.035 0.733333333333333
0.04 0.813333333333333
0.045 0.826666666666667
0.05 0.84
};

\addplot [very thick, crimson2143940]
table {%
0 0
0.005 0.0533333333333333
0.01 0.266666666666667
0.015 0.413333333333333
0.02 0.48
0.025 0.626666666666667
0.03 0.653333333333333
0.035 0.693333333333333
0.04 0.746666666666667
0.045 0.8
0.05 0.826666666666667
};

\addplot [very thick, mediumpurple148103189]
table {%
0 0
0.005 0.0266666666666667
0.01 0.106666666666667
0.015 0.333333333333333
0.02 0.533333333333333
0.025 0.653333333333333
0.03 0.693333333333333
0.035 0.746666666666667
0.04 0.813333333333333
0.045 0.853333333333333
0.05 0.88
};

\addplot [very thick, sienna1408675]
table {%
0 0
0.005 0.04
0.01 0.2
0.015 0.426666666666667
0.02 0.56
0.025 0.693333333333333
0.03 0.773333333333333
0.035 0.813333333333333
0.04 0.826666666666667
0.045 0.826666666666667
0.05 0.84
};

\addplot [very thick, orchid227119194]
table {%
0 0
0.005 0.253333333333333
0.01 0.653333333333333
0.015 0.786666666666667
0.02 0.826666666666667
0.025 0.906666666666667
0.03 0.906666666666667
0.035 0.92
0.04 0.933333333333333
0.045 0.933333333333333
0.05 0.946666666666667
};

\addplot [very thick, gray127]
table {%
0 0
0.005 0.04
0.01 0.226666666666667
0.015 0.533333333333333
0.02 0.706666666666667
0.025 0.84
0.03 0.866666666666667
0.035 0.866666666666667
0.04 0.893333333333333
0.045 0.92
0.05 0.946666666666667
};

\addplot [very thick, darkturquoise23190207]
table {%
0 0
0.005 0
0.01 0.12
0.015 0.346666666666667
0.02 0.386666666666667
0.025 0.613333333333333
0.03 0.693333333333333
0.035 0.746666666666667
0.04 0.826666666666667
0.045 0.84
0.05 0.853333333333333
};
\end{axis}%
\end{tikzpicture}%
%
        \begin{tikzpicture}

\definecolor{crimson2143940}{RGB}{214,39,40}
\definecolor{darkgray176}{RGB}{176,176,176}
\definecolor{darkorange25512714}{RGB}{255,127,14}
\definecolor{forestgreen4416044}{RGB}{44,160,44}
\definecolor{darkturquoise23190207}{RGB}{23,190,207}
\definecolor{gray127}{RGB}{127,127,127}
\definecolor{lightgray204}{RGB}{204,204,204}
\definecolor{mediumpurple148103189}{RGB}{148,103,189}
\definecolor{orchid227119194}{RGB}{227,119,194}
\definecolor{sienna1408675}{RGB}{140,86,75}
\definecolor{steelblue31119180}{RGB}{31,119,180}

\begin{axis}[
width=0.24\textwidth,
height=0.2\textwidth,
scale only axis=true,
tick align=outside,
tick pos=left,
x grid style={darkgray176},
xlabel={$\delta\vphantom{ / \mathrm{cm}}$},
xticklabel=$\mathclap{\pgfmathprintnumber{\tick}\degree{}}$,
xmajorgrids,
xmin=-1, xmax=21,
xtick style={color=black},
y grid style={darkgray176},
ymajorgrids,
ymin=-0.05, ymax=1.05,
ytick style={color=black}
]
\addplot [very thick, steelblue31119180]
table {%
0 0
1 0
2 0.0133333333333333
3 0.0133333333333333
4 0.0133333333333333
5 0.0133333333333333
6 0.0266666666666667
7 0.04
8 0.04
9 0.04
10 0.0533333333333333
11 0.0533333333333333
12 0.0666666666666667
13 0.0666666666666667
14 0.0666666666666667
15 0.0666666666666667
16 0.0666666666666667
17 0.08
18 0.08
19 0.106666666666667
20 0.12
};

\addplot [very thick, darkorange25512714]
table {%
0 0
1 0.0133333333333333
2 0.0133333333333333
3 0.04
4 0.0533333333333333
5 0.08
6 0.08
7 0.106666666666667
8 0.146666666666667
9 0.173333333333333
10 0.2
11 0.213333333333333
12 0.213333333333333
13 0.226666666666667
14 0.24
15 0.253333333333333
16 0.28
17 0.28
18 0.306666666666667
19 0.306666666666667
20 0.32
};

\addplot [very thick, forestgreen4416044]
table {%
0 0
1 0.0133333333333333
2 0.0533333333333333
3 0.0933333333333333
4 0.12
5 0.16
6 0.186666666666667
7 0.213333333333333
8 0.253333333333333
9 0.253333333333333
10 0.28
11 0.28
12 0.293333333333333
13 0.306666666666667
14 0.333333333333333
15 0.36
16 0.373333333333333
17 0.373333333333333
18 0.373333333333333
19 0.373333333333333
20 0.373333333333333
};

\addplot [very thick, crimson2143940]
table {%
0 0
1 0.0133333333333333
2 0.04
3 0.146666666666667
4 0.213333333333333
5 0.213333333333333
6 0.24
7 0.24
8 0.266666666666667
9 0.266666666666667
10 0.266666666666667
11 0.266666666666667
12 0.266666666666667
13 0.293333333333333
14 0.293333333333333
15 0.293333333333333
16 0.293333333333333
17 0.293333333333333
18 0.293333333333333
19 0.306666666666667
20 0.32
};

\addplot [very thick, mediumpurple148103189]
table {%
0 0
1 0.0266666666666667
2 0.0533333333333333
3 0.12
4 0.186666666666667
5 0.24
6 0.293333333333333
7 0.333333333333333
8 0.346666666666667
9 0.386666666666667
10 0.413333333333333
11 0.426666666666667
12 0.44
13 0.44
14 0.453333333333333
15 0.466666666666667
16 0.48
17 0.493333333333333
18 0.506666666666667
19 0.546666666666667
20 0.546666666666667
};

\addplot [very thick, sienna1408675]
table {%
0 0
1 0
2 0.12
3 0.24
4 0.306666666666667
5 0.36
6 0.4
7 0.466666666666667
8 0.52
9 0.546666666666667
10 0.573333333333333
11 0.573333333333333
12 0.573333333333333
13 0.586666666666667
14 0.586666666666667
15 0.6
16 0.6
17 0.6
18 0.613333333333333
19 0.626666666666667
20 0.626666666666667
};

\addplot [very thick, orchid227119194]
table {%
0 0
1 0.133333333333333
2 0.293333333333333
3 0.44
4 0.56
5 0.6
6 0.64
7 0.68
8 0.693333333333333
9 0.693333333333333
10 0.693333333333333
11 0.72
12 0.76
13 0.773333333333333
14 0.773333333333333
15 0.786666666666667
16 0.786666666666667
17 0.8
18 0.813333333333333
19 0.813333333333333
20 0.813333333333333
};

\addplot [very thick, gray127]
table {%
0 0
1 0.04
2 0.106666666666667
3 0.12
4 0.173333333333333
5 0.213333333333333
6 0.24
7 0.293333333333333
8 0.333333333333333
9 0.346666666666667
10 0.373333333333333
11 0.426666666666667
12 0.44
13 0.466666666666667
14 0.466666666666667
15 0.48
16 0.493333333333333
17 0.493333333333333
18 0.493333333333333
19 0.52
20 0.533333333333333
};

\addplot [very thick, darkturquoise23190207]
table {%
0 0
1 0
2 0.04
3 0.0533333333333333
4 0.08
5 0.106666666666667
6 0.146666666666667
7 0.16
8 0.186666666666667
9 0.226666666666667
10 0.24
11 0.253333333333333
12 0.266666666666667
13 0.28
14 0.293333333333333
15 0.293333333333333
16 0.293333333333333
17 0.306666666666667
18 0.333333333333333
19 0.333333333333333
20 0.346666666666667
};
\end{axis}%
\end{tikzpicture}%
%
        \begin{tikzpicture}

\definecolor{crimson2143940}{RGB}{214,39,40}
\definecolor{darkgray176}{RGB}{176,176,176}
\definecolor{darkorange25512714}{RGB}{255,127,14}
\definecolor{forestgreen4416044}{RGB}{44,160,44}
\definecolor{darkturquoise23190207}{RGB}{23,190,207}
\definecolor{gray127}{RGB}{127,127,127}
\definecolor{lightgray204}{RGB}{204,204,204}
\definecolor{mediumpurple148103189}{RGB}{148,103,189}
\definecolor{orchid227119194}{RGB}{227,119,194}
\definecolor{sienna1408675}{RGB}{140,86,75}
\definecolor{steelblue31119180}{RGB}{31,119,180}

\begin{axis}[
width=0.24\textwidth,
height=0.2\textwidth,
scale only axis=true,
tick align=outside,
tick pos=left,
x grid style={darkgray176},
xticklabel=$\mathclap{\pgfmathprintnumber{\tick}}\vphantom{\degree{}}$,
xlabel={$F_{1\mathrm{cm}}\vphantom{ / \mathrm{cm}}$},
xmajorgrids,
xmin=-0.05, xmax=1.05,
xtick style={color=black},
y grid style={darkgray176},
ymajorgrids,
ymin=-0.05, ymax=1.05,
ytick style={color=black}
]
\addplot [very thick, steelblue31119180]
table {%
0 1
0.05 0.893333333333333
0.1 0.706666666666667
0.15 0.586666666666667
0.2 0.493333333333333
0.25 0.466666666666667
0.3 0.413333333333333
0.35 0.32
0.4 0.226666666666667
0.4 0.226666666666667
0.45 0.2
0.5 0.146666666666667
0.55 0.08
0.6 0.0266666666666667
0.65 0.0133333333333333
0.7 0
0.75 0
0.8 0
0.85 0
0.9 0
0.95 0
1 0
};

\addplot [very thick, darkorange25512714]
table {%
0 1
0.05 0.986666666666667
0.1 0.973333333333333
0.15 0.906666666666667
0.2 0.906666666666667
0.25 0.853333333333333
0.3 0.76
0.35 0.613333333333333
0.4 0.573333333333333
0.4 0.573333333333333
0.45 0.533333333333333
0.5 0.52
0.55 0.466666666666667
0.6 0.44
0.65 0.333333333333333
0.7 0.24
0.75 0.16
0.8 0.106666666666667
0.85 0.08
0.9 0.04
0.95 0.0133333333333333
1 0
};

\addplot [very thick, forestgreen4416044]
table {%
0 1
0.05 1
0.1 1
0.15 0.96
0.2 0.92
0.25 0.866666666666667
0.3 0.76
0.35 0.706666666666667
0.4 0.706666666666667
0.4 0.706666666666667
0.45 0.666666666666667
0.5 0.6
0.55 0.573333333333333
0.6 0.506666666666667
0.65 0.44
0.7 0.373333333333333
0.75 0.333333333333333
0.8 0.24
0.85 0.213333333333333
0.9 0.146666666666667
0.95 0.0533333333333333
1 0
};

\addplot [very thick, crimson2143940]
table {%
0 1
0.05 1
0.1 0.973333333333333
0.15 0.92
0.2 0.88
0.25 0.84
0.3 0.76
0.35 0.746666666666667
0.4 0.666666666666667
0.4 0.666666666666667
0.45 0.6
0.5 0.573333333333333
0.55 0.52
0.6 0.48
0.65 0.453333333333333
0.7 0.386666666666667
0.75 0.32
0.8 0.266666666666667
0.85 0.2
0.9 0.16
0.95 0.0533333333333333
1 0
};

\addplot [very thick, mediumpurple148103189]
table {%
0 1
0.05 1
0.1 0.986666666666667
0.15 0.96
0.2 0.96
0.25 0.946666666666667
0.3 0.813333333333333
0.35 0.786666666666667
0.4 0.746666666666667
0.4 0.746666666666667
0.45 0.693333333333333
0.5 0.626666666666667
0.55 0.533333333333333
0.6 0.426666666666667
0.65 0.306666666666667
0.7 0.24
0.75 0.173333333333333
0.8 0.12
0.85 0.0533333333333333
0.9 0.0133333333333333
0.95 0
1 0
};

\addplot [very thick, sienna1408675]
table {%
0 1
0.05 0.866666666666667
0.1 0.813333333333333
0.15 0.813333333333333
0.2 0.813333333333333
0.25 0.8
0.3 0.786666666666667
0.35 0.773333333333333
0.4 0.773333333333333
0.4 0.773333333333333
0.45 0.76
0.5 0.706666666666667
0.55 0.666666666666667
0.6 0.586666666666667
0.65 0.546666666666667
0.7 0.453333333333333
0.75 0.426666666666667
0.8 0.346666666666667
0.85 0.226666666666667
0.9 0.12
0.95 0.0666666666666667
1 0
};

\addplot [very thick, orchid227119194]
table {%
0 1
0.05 1
0.1 0.986666666666667
0.15 0.973333333333333
0.2 0.973333333333333
0.25 0.96
0.3 0.96
0.35 0.96
0.4 0.933333333333333
0.4 0.933333333333333
0.45 0.92
0.5 0.906666666666667
0.55 0.906666666666667
0.6 0.866666666666667
0.65 0.853333333333333
0.7 0.813333333333333
0.75 0.746666666666667
0.8 0.72
0.85 0.64
0.9 0.573333333333333
0.95 0.386666666666667
1 0
};

\addplot [very thick, gray127]
table {%
0 1
0.05 1
0.1 1
0.15 0.986666666666667
0.2 0.946666666666667
0.25 0.933333333333333
0.3 0.88
0.35 0.813333333333333
0.4 0.746666666666667
0.4 0.746666666666667
0.45 0.666666666666667
0.5 0.64
0.55 0.573333333333333
0.6 0.493333333333333
0.65 0.466666666666667
0.7 0.346666666666667
0.75 0.266666666666667
0.8 0.186666666666667
0.85 0.133333333333333
0.9 0.0533333333333333
0.95 0.0266666666666667
1 0
};

\addplot [very thick, darkturquoise23190207]
table {%
0 1
0.05 0.96
0.1 0.906666666666667
0.15 0.84
0.2 0.813333333333333
0.25 0.693333333333333
0.3 0.6
0.35 0.573333333333333
0.4 0.506666666666667
0.4 0.506666666666667
0.45 0.493333333333333
0.5 0.44
0.55 0.4
0.6 0.32
0.65 0.24
0.7 0.16
0.75 0.133333333333333
0.8 0.106666666666667
0.85 0.04
0.9 0.04
0.95 0.0133333333333333
1 0
};
\end{axis}%
\end{tikzpicture}%
%
        \caption{REDWOOD75}
    \end{subfigure}%
    \vspace{0.4cm}
    \begin{tikzpicture}

\definecolor{crimson2143940}{RGB}{214,39,40}
\definecolor{darkgray176}{RGB}{176,176,176}
\definecolor{darkorange25512714}{RGB}{255,127,14}
\definecolor{forestgreen4416044}{RGB}{44,160,44}
\definecolor{darkturquoise23190207}{RGB}{23,190,207}
\definecolor{gray127}{RGB}{127,127,127}
\definecolor{lightgray204}{RGB}{204,204,204}
\definecolor{mediumpurple148103189}{RGB}{148,103,189}
\definecolor{orchid227119194}{RGB}{227,119,194}
\definecolor{sienna1408675}{RGB}{140,86,75}
\definecolor{steelblue31119180}{RGB}{31,119,180}

\begin{axis}[%
legend columns=9,
scale only axis,width=1mm, %
hide axis,
legend image code/.code={%
    \draw[#1, fill] (0cm,0.025cm) circle (0.1cm);
},  
legend style={
  draw=none,
  column sep=0.3ex,
  font=\footnotesize,
  /tikz/every even column/.append style={column sep=0.4cm}
},
]

\addplot [thick, steelblue31119180]
table {%
0 0
1 1
};
\addlegendentry{CASS}

\addplot [thick, darkorange25512714]
table {%
0 0
1 1
};
\addlegendentry{SPD}

\addplot [thick, forestgreen4416044]
table {%
0 0
1 1
};
\addlegendentry{CR-Net}

\addplot [thick, crimson2143940]
table {%
0 0
1 1
};
\addlegendentry{SGPA}

\addplot [thick, mediumpurple148103189]
table {%
0 0
1 1
};
\addlegendentry{ASM-Net}

\addplot [thick, sienna1408675]
table {%
0 0
1 1
};
\addlegendentry{iCaps}

\addplot [thick, orchid227119194]
table {%
0 0
1 1
};
\addlegendentry{SDFEst}

\addplot [thick, gray127]
table {%
0 0
1 1
};
\addlegendentry{DPDN}

\addplot [thick, darkturquoise23190207]
table {%
0 0
1 1
};
\addlegendentry{RBP-Pose}

\end{axis}%
\end{tikzpicture}%
    \caption{Detailed precision results for varying thresholds of position, orientation and F-score thresholds.}
    \label{fig:varying_thresholds}
\end{figure*}
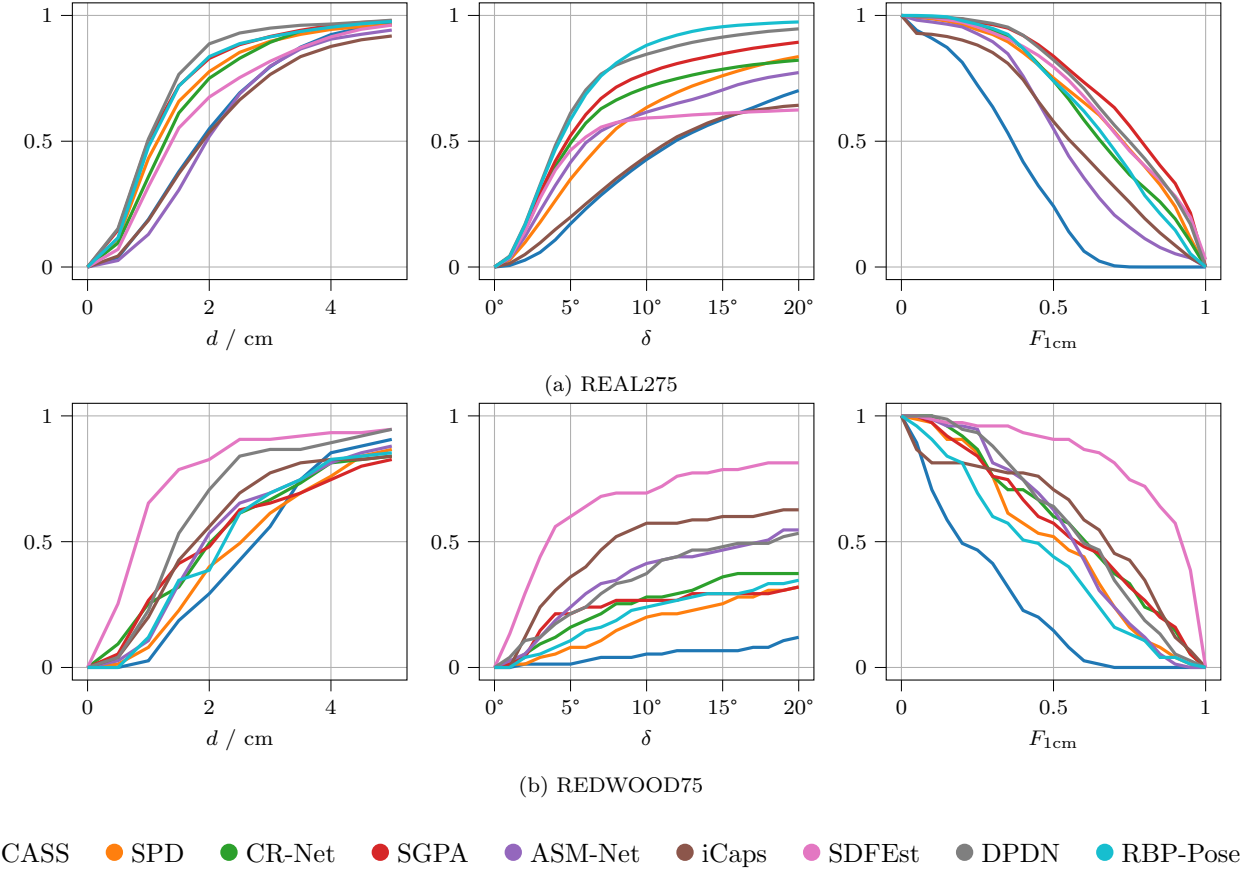

To gain further insight into the estimation quality of the methods, we show detailed results for varying thresholds in \cref{fig:varying_thresholds}. It can be seen that the difference between the two datasets is most pronounced for orientation-based thresholding. This confirms the issue of constrained orientations discussed in Section \ref{sec:datasets} (see \cref{fig:orientation_dist}). Unbiased methods (i.e., SDFEst and iCaps) that are trained on unconstrained synthetically generated data perform significantly better on REDWOOD75, however they level out earlier on REAL275, partly due to the aforementioned ambiguities, which other methods avoid through dataset specific biases. On REDWOOD75, ASM-Net which is trained on synthetic data (with the exact distribution unknown) performs comparatively well. Similarly DPDN, which is trained with augmented point sets, shows some generalization capability, however, less so than unbiased methods.

In \cref{tab:categories} we further report the precision per category for the more lenient thresholds $\delta=10\degree, d=\SI{2}{cm}, F_{1\mathrm{cm}}=0.6$. Note that CASS' reconstructions are relatively sparse and very noisy, and therefore rarely reach $F_{1\mathrm{cm}}>0.6$ (see also \cref{fig:varying_thresholds}). On REAL275, all methods fail at the camera category which contains significantly more shape variation than the other categories. RBP-Pose and DPDN show some encouraging results here, but still perform significantly worse than for other categories. SDFEst shows a drop in performance for geometrically ambiguous categories, such as can and laptop, for which other methods perform comparatively better.

\paragraph{Run Time} We also compute the mean run time of each method per prediction. We include the transfer time from CPU to GPU and mesh reconstruction for iCaps and SDFEst, but exclude the computation of the metrics. In our experiments\footnote{Intel Core i7-6850K, NVIDIA TITAN X (Pascal); mean run time on REDWOOD75 dataset}, CASS required $\SI{24.8}{ms}$, SPD $\SI{227}{ms}$, CR-Net $\SI{225}{ms}$, SGPA $\SI{227}{ms}$, ASM-Net $\SI{113}{ms}$,  iCaps $\SI{2.44}{s}$, SDFEst $\SI{1.32}{s}$, DPDN $\SI{251}{ms}$, and RBP-Pose $\SI{171}{ms}$. Figure \ref{fig:overview} visualizes the run time and performance of all methods on both datasets.

\section{Limitations}\label{sec:limitations}

\subsection{Comparability} The results from Section \ref{sec:experiments} suggest that training on synthetic data currently generalizes better to unconstrained orientations. This is expected, since synthetic data generation provides perfectly annotated and unconstrained training data. However, it introduces a synthetic-to-real domain gap, which needs to be addressed. This opens the question of how well methods such as DPDN and RBP-Pose would perform when trained on unconstrained synthetic images. In general, since methods currently vary significant parts of training datasets, architecture, pose parameterization, and losses, it is difficult to assess the contributions of individual changes.

\subsection{Multimodal Distributions} Unconstrained pose estimation introduces significant difficulties in the task, which were hidden due to the constraints present in the CAMERA and REAL datasets. Consider, for example, the bottom left mug in \cref{fig:redwood_examples}. From the given view, it is difficult to tell which way the opening of the mug faces. Another example are cans, which are geometry-wise nearly symmetric. Currently there are only few works \cite{manhardt2019explaining,deng2022deep} that consider this problem of ambiguous poses. Evaluation of such methods that predict multimodal posteriors is difficult. One possible way of extending the presented framework to such methods would be to allow methods to generate $N$ hypotheses. Two precision values $P_\mathrm{best}$ and $P_\mathrm{worst}$ could be computed based on the best and worst hypothesis, respectively. A strong method would generate the same hypothesis $N$ times if there is no ambiguity, leading to a higher  $P_\mathrm{best}$ and  $P_\mathrm{worst}$. Similarly, if there is ambiguity, the correct hypothesis would still be contained in the set of hypotheses leading to a higher $P_\mathrm{best}$.

\subsection{Dataset Size} The REDWOOD75 dataset is limited in size, but the results suggest a clear lack of generalization capability of current approaches. This shows the need for larger datasets for unconstrained pose and shape estimation. It is an open question how such a dataset could be collected in the most efficient way.

\section{Conclusion and Outlook}
In this work, we have discussed the current state of categorical pose and shape estimation and identified several limitations of the current evaluation protocol. In particular, existing evaluation datasets contain only a heavily constrained set of orientations, which simplifies the problem by removing pose and shape ambiguities. Furthermore, existing evaluation metrics are suboptimal and unnecessarily difficult to interpret. To alleviate these problems, we propose a new set of metrics and new annotations for the REDWOOD dataset, which contains less constrained orientations. We apply our evaluation protocol to nine state-of-the-art methods and confirm limited generalization capability as suggested by the constrained orientations in their training data.

Our experiments suggest that there is a need for larger, high-quality datasets for unconstrained pose and shape estimation as well as for robust methods that can handle unconstrained orientations and the resulting pose ambiguities in a principled way.

\section*{Acknowledgements}
This work was partially supported by the Wallenberg AI, Autonomous Systems and Software Program (WASP) funded by the Knut and Alice Wallenberg Foundation.

\bibliographystyle{elsarticle-num-names}
\bibliography{bib.bib}

\begin{thebibliography}{81}
\expandafter\ifx\csname natexlab\endcsname\relax\def\natexlab#1{#1}\fi
\providecommand{\url}[1]{\texttt{#1}}
\providecommand{\href}[2]{#2}
\providecommand{\path}[1]{#1}
\providecommand{\DOIprefix}{doi:}
\providecommand{\ArXivprefix}{arXiv:}
\providecommand{\URLprefix}{URL: }
\providecommand{\Pubmedprefix}{pmid:}
\providecommand{\doi}[1]{\href{http://dx.doi.org/#1}{\path{#1}}}
\providecommand{\Pubmed}[1]{\href{pmid:#1}{\path{#1}}}
\providecommand{\bibinfo}[2]{#2}
\ifx\xfnm\relax \def\xfnm[#1]{\unskip,\space#1}\fi
\bibitem[{Varley et~al.(2017)Varley, DeChant, Richardson, Ruales, and
  Allen}]{varley2017shape}
\bibinfo{author}{J.~Varley}, \bibinfo{author}{C.~DeChant},
  \bibinfo{author}{A.~Richardson}, \bibinfo{author}{J.~Ruales},
  \bibinfo{author}{P.~Allen},
\newblock \bibinfo{title}{Shape completion enabled robotic grasping},
\newblock in: \bibinfo{booktitle}{Proceedings of the IEEE/RSJ International
  Conference on Intelligent Robots and Systems}, \bibinfo{year}{2017}, pp.
  \bibinfo{pages}{2442--2447}.
\bibitem[{Sucar et~al.(2020)Sucar, Wada, and Davison}]{sucar2020nodeslam}
\bibinfo{author}{E.~Sucar}, \bibinfo{author}{K.~Wada},
  \bibinfo{author}{A.~Davison},
\newblock \bibinfo{title}{{NodeSLAM}: Neural object descriptors for multi-view
  shape reconstruction},
\newblock in: \bibinfo{booktitle}{Proceedings of the International Conference
  on {3D} Vision}, \bibinfo{year}{2020}, pp. \bibinfo{pages}{949--958}.
\bibitem[{Wang et~al.(2019)Wang, Sridhar, Huang, Valentin, Song, and
  Guibas}]{wang2019normalized}
\bibinfo{author}{H.~Wang}, \bibinfo{author}{S.~Sridhar},
  \bibinfo{author}{J.~Huang}, \bibinfo{author}{J.~Valentin},
  \bibinfo{author}{S.~Song}, \bibinfo{author}{L.~J. Guibas},
\newblock \bibinfo{title}{Normalized object coordinate space for category-level
  {6D} object pose and size estimation},
\newblock in: \bibinfo{booktitle}{Proceedings of the IEEE/CVF Conference on
  Computer Vision and Pattern Recognition}, \bibinfo{year}{2019}, pp.
  \bibinfo{pages}{2642--2651}.
\bibitem[{Choi et~al.(2016)Choi, Zhou, Miller, and Koltun}]{choi2016large}
\bibinfo{author}{S.~Choi}, \bibinfo{author}{Q.-Y. Zhou},
  \bibinfo{author}{S.~Miller}, \bibinfo{author}{V.~Koltun},
\newblock \bibinfo{title}{A large dataset of object scans},
\newblock \bibinfo{journal}{arXiv preprint arXiv:1602.02481}
  (\bibinfo{year}{2016}).
\bibitem[{Chen et~al.(2020)Chen, Li, Wang, and Xu}]{chen2020learning}
\bibinfo{author}{D.~Chen}, \bibinfo{author}{J.~Li}, \bibinfo{author}{Z.~Wang},
  \bibinfo{author}{K.~Xu},
\newblock \bibinfo{title}{Learning canonical shape space for category-level
  {6D} object pose and size estimation},
\newblock in: \bibinfo{booktitle}{Proceedings of the IEEE/CVF Conference on
  Computer Vision and Pattern Recognition}, \bibinfo{year}{2020}, pp.
  \bibinfo{pages}{11973--11982}.
\bibitem[{Tian et~al.(2020)Tian, Ang, and Lee}]{tian2020shape}
\bibinfo{author}{M.~Tian}, \bibinfo{author}{M.~H. Ang}, \bibinfo{author}{G.~H.
  Lee},
\newblock \bibinfo{title}{Shape prior deformation for categorical {6D} object
  pose and size estimation},
\newblock in: \bibinfo{booktitle}{Proceedings of the European Conference on
  Computer Vision}, \bibinfo{organization}{Springer}, \bibinfo{year}{2020}, pp.
  \bibinfo{pages}{530--546}.
\bibitem[{Tatarchenko et~al.(2019)Tatarchenko, Richter, Ranftl, Li, Koltun, and
  Brox}]{tatarchenko2019single}
\bibinfo{author}{M.~Tatarchenko}, \bibinfo{author}{S.~R. Richter},
  \bibinfo{author}{R.~Ranftl}, \bibinfo{author}{Z.~Li},
  \bibinfo{author}{V.~Koltun}, \bibinfo{author}{T.~Brox},
\newblock \bibinfo{title}{What do single-view {3D} reconstruction networks
  learn?},
\newblock in: \bibinfo{booktitle}{Proceedings of the IEEE/CVF Conference on
  Computer Vision and Pattern Recognition}, \bibinfo{year}{2019}, pp.
  \bibinfo{pages}{3405--3414}.
\bibitem[{Hoda{\v{n}} et~al.(2020)Hoda{\v{n}}, Sundermeyer, Drost, Labb{\'e},
  Brachmann, Michel, Rother, and Matas}]{hodavn2020bop}
\bibinfo{author}{T.~Hoda{\v{n}}}, \bibinfo{author}{M.~Sundermeyer},
  \bibinfo{author}{B.~Drost}, \bibinfo{author}{Y.~Labb{\'e}},
  \bibinfo{author}{E.~Brachmann}, \bibinfo{author}{F.~Michel},
  \bibinfo{author}{C.~Rother}, \bibinfo{author}{J.~Matas},
\newblock \bibinfo{title}{{BOP} challenge 2020 on {6D} object localization},
\newblock in: \bibinfo{booktitle}{Proceedings of the European Conference on
  Computer Vision}, \bibinfo{organization}{Springer}, \bibinfo{year}{2020}, pp.
  \bibinfo{pages}{577--594}.
\bibitem[{Chang et~al.(2015)Chang, Funkhouser, Guibas, Hanrahan, Huang, Li,
  Savarese, Savva, Song, Su, Xiao, Yi, and Yu}]{chang2015}
\bibinfo{author}{A.~X. Chang}, \bibinfo{author}{T.~Funkhouser},
  \bibinfo{author}{L.~Guibas}, \bibinfo{author}{P.~Hanrahan},
  \bibinfo{author}{Q.~Huang}, \bibinfo{author}{Z.~Li},
  \bibinfo{author}{S.~Savarese}, \bibinfo{author}{M.~Savva},
  \bibinfo{author}{S.~Song}, \bibinfo{author}{H.~Su},
  \bibinfo{author}{J.~Xiao}, \bibinfo{author}{L.~Yi}, \bibinfo{author}{F.~Yu},
  \bibinfo{title}{{ShapeNet}: An Information-Rich {3D} Model Repository},
  \bibinfo{type}{Technical Report} \bibinfo{number}{1512.03012}, arXiv
  preprint, \bibinfo{year}{2015}.
\bibitem[{He et~al.(2017)He, Gkioxari, Doll{\'a}r, and Girshick}]{he2017mask}
\bibinfo{author}{K.~He}, \bibinfo{author}{G.~Gkioxari},
  \bibinfo{author}{P.~Doll{\'a}r}, \bibinfo{author}{R.~Girshick},
\newblock \bibinfo{title}{{Mask R-CNN}},
\newblock in: \bibinfo{booktitle}{Proceedings of the IEEE International
  Conference on Computer Vision}, \bibinfo{year}{2017}, pp.
  \bibinfo{pages}{2961--2969}.
\bibitem[{Umeyama(1991)}]{umeyama1991least}
\bibinfo{author}{S.~Umeyama},
\newblock \bibinfo{title}{Least-squares estimation of transformation parameters
  between two point patterns},
\newblock \bibinfo{journal}{IEEE Transactions on Pattern Analysis and Machine
  Intelligence} \bibinfo{volume}{13} (\bibinfo{year}{1991})
  \bibinfo{pages}{376--380}.
\bibitem[{Fischler and Bolles(1981)}]{fischler1981random}
\bibinfo{author}{M.~A. Fischler}, \bibinfo{author}{R.~C. Bolles},
\newblock \bibinfo{title}{Random sample consensus: a paradigm for model fitting
  with applications to image analysis and automated cartography},
\newblock \bibinfo{journal}{Communications of the ACM} \bibinfo{volume}{24}
  (\bibinfo{year}{1981}) \bibinfo{pages}{381--395}.
\bibitem[{Kingma and Welling(2014)}]{kingma2013auto}
\bibinfo{author}{D.~P. Kingma}, \bibinfo{author}{M.~Welling},
\newblock \bibinfo{title}{Auto-encoding variational {B}ayes},
\newblock in: \bibinfo{booktitle}{Proceedings of the International Conference
  on Learning Representations}, \bibinfo{year}{2014}.
\bibitem[{Wang et~al.(2021)Wang, Chen, and Dou}]{wang2021category}
\bibinfo{author}{J.~Wang}, \bibinfo{author}{K.~Chen}, \bibinfo{author}{Q.~Dou},
\newblock \bibinfo{title}{Category-level {6D} object pose estimation via
  cascaded relation and recurrent reconstruction networks},
\newblock in: \bibinfo{booktitle}{Proceedings of the IEEE/RSJ International
  Conference on Intelligent Robots and Systems}, \bibinfo{year}{2021}, pp.
  \bibinfo{pages}{4807--4814}.
\bibitem[{Chen and Dou(2021)}]{chen2021sgpa}
\bibinfo{author}{K.~Chen}, \bibinfo{author}{Q.~Dou},
\newblock \bibinfo{title}{{SGPA}: Structure-guided prior adaptation for
  category-level {6D} object pose estimation},
\newblock in: \bibinfo{booktitle}{Proceedings of the International Conference
  on Computer Vision}, \bibinfo{year}{2021}, pp. \bibinfo{pages}{2773--2782}.
\bibitem[{Lin et~al.(2022)Lin, Wei, Ding, and Jia}]{lin2022category}
\bibinfo{author}{J.~Lin}, \bibinfo{author}{Z.~Wei}, \bibinfo{author}{C.~Ding},
  \bibinfo{author}{K.~Jia},
\newblock \bibinfo{title}{Category-level {6D} object pose and size estimation
  using self-supervised deep prior deformation networks},
\newblock in: \bibinfo{booktitle}{Proceedings of the European Conference on
  Computer Vision}, \bibinfo{organization}{Springer}, \bibinfo{year}{2022}, pp.
  \bibinfo{pages}{19--34}.
\bibitem[{Li et~al.(2020)Li, Wang, Ji, Xiang, and Fox}]{li2020deepim}
\bibinfo{author}{Y.~Li}, \bibinfo{author}{G.~Wang}, \bibinfo{author}{X.~Ji},
  \bibinfo{author}{Y.~Xiang}, \bibinfo{author}{D.~Fox},
\newblock \bibinfo{title}{{DeepIM}: Deep iterative matching for {6D} pose
  estimation},
\newblock \bibinfo{journal}{International Journal of Computer Vision}
  \bibinfo{volume}{128} (\bibinfo{year}{2020}) \bibinfo{pages}{657--678}.
\bibitem[{Wang et~al.(2019)Wang, Xu, Zhu, Mart{\'\i}n-Mart{\'\i}n, Lu, Fei-Fei,
  and Savarese}]{wang2019densefusion}
\bibinfo{author}{C.~Wang}, \bibinfo{author}{D.~Xu}, \bibinfo{author}{Y.~Zhu},
  \bibinfo{author}{R.~Mart{\'\i}n-Mart{\'\i}n}, \bibinfo{author}{C.~Lu},
  \bibinfo{author}{L.~Fei-Fei}, \bibinfo{author}{S.~Savarese},
\newblock \bibinfo{title}{{DenseFusion}: {6D} object pose estimation by
  iterative dense fusion},
\newblock in: \bibinfo{booktitle}{Proceedings of the IEEE/CVF conference on
  computer vision and pattern recognition}, \bibinfo{year}{2019}, pp.
  \bibinfo{pages}{3343--3352}.
\bibitem[{Labb{\'e} et~al.(2020)Labb{\'e}, Carpentier, Aubry, and
  Sivic}]{labbe2020cosypose}
\bibinfo{author}{Y.~Labb{\'e}}, \bibinfo{author}{J.~Carpentier},
  \bibinfo{author}{M.~Aubry}, \bibinfo{author}{J.~Sivic},
\newblock \bibinfo{title}{{CosyPose}: Consistent multi-view multi-object {6D}
  pose estimation},
\newblock in: \bibinfo{booktitle}{European Conference on Computer Vision},
  \bibinfo{organization}{Springer}, \bibinfo{year}{2020}, pp.
  \bibinfo{pages}{574--591}.
\bibitem[{Shugurov et~al.(2021)Shugurov, Pavlov, Zakharov, and
  Ilic}]{shugurov2021multi}
\bibinfo{author}{I.~Shugurov}, \bibinfo{author}{I.~Pavlov},
  \bibinfo{author}{S.~Zakharov}, \bibinfo{author}{S.~Ilic},
\newblock \bibinfo{title}{Multi-view object pose refinement with differentiable
  renderer},
\newblock \bibinfo{journal}{IEEE Robotics and Automation Letters}
  \bibinfo{volume}{6} (\bibinfo{year}{2021}) \bibinfo{pages}{2579--2586}.
\bibitem[{Chen et~al.(2020)Chen, Dong, Song, Geiger, and
  Hilliges}]{chen2020category}
\bibinfo{author}{X.~Chen}, \bibinfo{author}{Z.~Dong},
  \bibinfo{author}{J.~Song}, \bibinfo{author}{A.~Geiger},
  \bibinfo{author}{O.~Hilliges},
\newblock \bibinfo{title}{Category level object pose estimation via neural
  analysis-by-synthesis},
\newblock in: \bibinfo{booktitle}{Proceedings of the European Conference on
  Computer Vision}, \bibinfo{organization}{Springer}, \bibinfo{year}{2020}, pp.
  \bibinfo{pages}{139--156}.
\bibitem[{Deng et~al.(2022)Deng, Geng, Bretl, Xiang, and Fox}]{deng2022icaps}
\bibinfo{author}{X.~Deng}, \bibinfo{author}{J.~Geng},
  \bibinfo{author}{T.~Bretl}, \bibinfo{author}{Y.~Xiang},
  \bibinfo{author}{D.~Fox},
\newblock \bibinfo{title}{icaps: Iterative category-level object pose and shape
  estimation},
\newblock \bibinfo{journal}{IEEE Robotics and Automation Letters}
  \bibinfo{volume}{7} (\bibinfo{year}{2022}) \bibinfo{pages}{1784--1791}.
\bibitem[{Bruns and Jensfelt(2022)}]{bruns2022sdfest}
\bibinfo{author}{L.~Bruns}, \bibinfo{author}{P.~Jensfelt},
\newblock \bibinfo{title}{{SDFEst}: Categorical pose and shape estimation of
  objects from {RGB-D} using signed distance fields},
\newblock \bibinfo{journal}{IEEE Robotics and Automation Letters}
  \bibinfo{volume}{7} (\bibinfo{year}{2022}) \bibinfo{pages}{9597--9604}.
\bibitem[{Irshad et~al.(2022)Irshad, Zakharov, Ambrus, Kollar, Kira, and
  Gaidon}]{irshad2022shapo}
\bibinfo{author}{M.~Z. Irshad}, \bibinfo{author}{S.~Zakharov},
  \bibinfo{author}{R.~Ambrus}, \bibinfo{author}{T.~Kollar},
  \bibinfo{author}{Z.~Kira}, \bibinfo{author}{A.~Gaidon},
\newblock \bibinfo{title}{{ShAPO}: Implicit representations for multi-object
  shape, appearance, and pose optimization},
\newblock in: \bibinfo{booktitle}{Proceedings of the European Conference on
  Computer Vision}, \bibinfo{organization}{Springer}, \bibinfo{year}{2022}, pp.
  \bibinfo{pages}{275--292}.
\bibitem[{Manhardt et~al.(2020)Manhardt, Wang, Busam, Nickel, Meier, Minciullo,
  Ji, and Navab}]{manhardt2020cps++}
\bibinfo{author}{F.~Manhardt}, \bibinfo{author}{G.~Wang},
  \bibinfo{author}{B.~Busam}, \bibinfo{author}{M.~Nickel},
  \bibinfo{author}{S.~Meier}, \bibinfo{author}{L.~Minciullo},
  \bibinfo{author}{X.~Ji}, \bibinfo{author}{N.~Navab},
\newblock \bibinfo{title}{{CPS++}: Improving class-level {6D} pose and shape
  estimation from monocular images with self-supervised learning},
\newblock \bibinfo{journal}{arXiv preprint arXiv:2003.05848}
  (\bibinfo{year}{2020}).
\bibitem[{Lee et~al.(2021)Lee, Lee, Kim, and Kweon}]{lee2021category}
\bibinfo{author}{T.~Lee}, \bibinfo{author}{B.-U. Lee},
  \bibinfo{author}{M.~Kim}, \bibinfo{author}{I.~S. Kweon},
\newblock \bibinfo{title}{Category-level metric scale object shape and pose
  estimation},
\newblock \bibinfo{journal}{IEEE Robotics and Automation Letters}
  \bibinfo{volume}{6} (\bibinfo{year}{2021}) \bibinfo{pages}{8575--8582}.
\bibitem[{Gkioxari et~al.(2019)Gkioxari, Malik, and Johnson}]{gkioxari2019mesh}
\bibinfo{author}{G.~Gkioxari}, \bibinfo{author}{J.~Malik},
  \bibinfo{author}{J.~Johnson},
\newblock \bibinfo{title}{Mesh {R-CNN}},
\newblock in: \bibinfo{booktitle}{Proceedings of the International Conference
  on Computer Vision}, \bibinfo{year}{2019}, pp. \bibinfo{pages}{9785--9795}.
\bibitem[{Engelmann et~al.(2021)Engelmann, Rematas, Leibe, and
  Ferrari}]{engelmann2021points}
\bibinfo{author}{F.~Engelmann}, \bibinfo{author}{K.~Rematas},
  \bibinfo{author}{B.~Leibe}, \bibinfo{author}{V.~Ferrari},
\newblock \bibinfo{title}{From points to multi-object {3D} reconstruction},
\newblock in: \bibinfo{booktitle}{Proceedings of the IEEE/CVF Conference on
  Computer Vision and Pattern Recognition}, \bibinfo{year}{2021}, pp.
  \bibinfo{pages}{4588--4597}.
\bibitem[{Chen et~al.(2021)Chen, Jia, Chang, Duan, Shen, and
  Leonardis}]{chen2021fs}
\bibinfo{author}{W.~Chen}, \bibinfo{author}{X.~Jia}, \bibinfo{author}{H.~J.
  Chang}, \bibinfo{author}{J.~Duan}, \bibinfo{author}{L.~Shen},
  \bibinfo{author}{A.~Leonardis},
\newblock \bibinfo{title}{{FS-Net}: Fast shape-based network for category-level
  {6D} object pose estimation with decoupled rotation mechanism},
\newblock in: \bibinfo{booktitle}{Proceedings of the IEEE/CVF Conference on
  Computer Vision and Pattern Recognition}, \bibinfo{year}{2021}, pp.
  \bibinfo{pages}{1581--1590}.
\bibitem[{Li et~al.(2021)Li, Weng, Yi, Guibas, Abbott, Song, and
  Wang}]{li2021leveraging}
\bibinfo{author}{X.~Li}, \bibinfo{author}{Y.~Weng}, \bibinfo{author}{L.~Yi},
  \bibinfo{author}{L.~J. Guibas}, \bibinfo{author}{A.~Abbott},
  \bibinfo{author}{S.~Song}, \bibinfo{author}{H.~Wang},
\newblock \bibinfo{title}{Leveraging {SE(3)} equivariance for self-supervised
  category-level object pose estimation from point clouds},
\newblock \bibinfo{journal}{Advances in Neural Information Processing Systems}
  \bibinfo{volume}{34} (\bibinfo{year}{2021}) \bibinfo{pages}{15370--15381}.
\bibitem[{Lin et~al.(2021)Lin, Wei, Li, Xu, Jia, and Li}]{lin2021dualposenet}
\bibinfo{author}{J.~Lin}, \bibinfo{author}{Z.~Wei}, \bibinfo{author}{Z.~Li},
  \bibinfo{author}{S.~Xu}, \bibinfo{author}{K.~Jia}, \bibinfo{author}{Y.~Li},
\newblock \bibinfo{title}{{DualPoseNet}: Category-level {6D} object pose and
  size estimation using dual pose network with refined learning of pose
  consistency},
\newblock in: \bibinfo{booktitle}{Proceedings of the International Conference
  on Computer Vision}, \bibinfo{year}{2021}, pp. \bibinfo{pages}{3560--3569}.
\bibitem[{Di et~al.(2022)Di, Zhang, Lou, Manhardt, Ji, Navab, and
  Tombari}]{di2022gpv}
\bibinfo{author}{Y.~Di}, \bibinfo{author}{R.~Zhang}, \bibinfo{author}{Z.~Lou},
  \bibinfo{author}{F.~Manhardt}, \bibinfo{author}{X.~Ji},
  \bibinfo{author}{N.~Navab}, \bibinfo{author}{F.~Tombari},
\newblock \bibinfo{title}{{GPV-Pose}: Category-level object pose estimation via
  geometry-guided point-wise voting},
\newblock in: \bibinfo{booktitle}{Proceedings of the IEEE/CVF Conference on
  Computer Vision and Pattern Recognition}, \bibinfo{year}{2022}, pp.
  \bibinfo{pages}{6781--6791}.
\bibitem[{Lee et~al.(2022)Lee, Lee, Shin, Choe, Shin, Kweon, and
  Yoon}]{lee2022uda}
\bibinfo{author}{T.~Lee}, \bibinfo{author}{B.-U. Lee},
  \bibinfo{author}{I.~Shin}, \bibinfo{author}{J.~Choe},
  \bibinfo{author}{U.~Shin}, \bibinfo{author}{I.~S. Kweon},
  \bibinfo{author}{K.-J. Yoon},
\newblock \bibinfo{title}{{UDA-COPE}: Unsupervised domain adaptation for
  category-level object pose estimation},
\newblock in: \bibinfo{booktitle}{Proceedings of the IEEE/CVF Conference on
  Computer Vision and Pattern Recognition}, \bibinfo{year}{2022}, pp.
  \bibinfo{pages}{14891--14900}.
\bibitem[{R\"unz et~al.(2020)R\"unz, Li, Tang, Ma, Kong, Schmidt, Reid,
  Agapito, Straub, Lovegrove et~al.}]{runz2020frodo}
\bibinfo{author}{M.~R\"unz}, \bibinfo{author}{K.~Li},
  \bibinfo{author}{M.~Tang}, \bibinfo{author}{L.~Ma},
  \bibinfo{author}{C.~Kong}, \bibinfo{author}{T.~Schmidt},
  \bibinfo{author}{I.~Reid}, \bibinfo{author}{L.~Agapito},
  \bibinfo{author}{J.~Straub}, \bibinfo{author}{S.~Lovegrove}, et~al.,
\newblock \bibinfo{title}{{FroDO}: From detections to {3D} objects},
\newblock in: \bibinfo{booktitle}{Proceedings of the IEEE/CVF Conference on
  Computer Vision and Pattern Recognition}, \bibinfo{year}{2020}, pp.
  \bibinfo{pages}{14720--14729}.
\bibitem[{Park et~al.(2019)Park, Florence, Straub, Newcombe, and
  Lovegrove}]{park2019deepsdf}
\bibinfo{author}{J.~J. Park}, \bibinfo{author}{P.~Florence},
  \bibinfo{author}{J.~Straub}, \bibinfo{author}{R.~Newcombe},
  \bibinfo{author}{S.~Lovegrove},
\newblock \bibinfo{title}{{DeepSDF}: Learning continuous signed distance
  functions for shape representation},
\newblock in: \bibinfo{booktitle}{Proceedings of the IEEE/CVF Conference on
  Computer Vision and Pattern Recognition}, \bibinfo{year}{2019}, pp.
  \bibinfo{pages}{165--174}.
\bibitem[{Akizuki and Hashimoto(2021)}]{akizuki2021}
\bibinfo{author}{S.~Akizuki}, \bibinfo{author}{M.~Hashimoto},
\newblock \bibinfo{title}{{ASM-Net}: Category-level pose and shape estimation
  using parametric deformation},
\newblock in: \bibinfo{booktitle}{Proceedings of the British Machine Vision
  Conference}, \bibinfo{year}{2021}, pp. \bibinfo{pages}{1--13}.
\bibitem[{Fan et~al.(2017)Fan, Su, and Guibas}]{fan2017point}
\bibinfo{author}{H.~Fan}, \bibinfo{author}{H.~Su}, \bibinfo{author}{L.~J.
  Guibas},
\newblock \bibinfo{title}{A point set generation network for {3D} object
  reconstruction from a single image},
\newblock in: \bibinfo{booktitle}{Proceedings of the IEEE conference on
  computer vision and pattern recognition}, \bibinfo{year}{2017}, pp.
  \bibinfo{pages}{605--613}.
\bibitem[{Knapitsch et~al.(2017)Knapitsch, Park, Zhou, and
  Koltun}]{knapitsch2017}
\bibinfo{author}{A.~Knapitsch}, \bibinfo{author}{J.~Park},
  \bibinfo{author}{Q.-Y. Zhou}, \bibinfo{author}{V.~Koltun},
\newblock \bibinfo{title}{Tanks and temples: Benchmarking large-scale scene
  reconstruction},
\newblock \bibinfo{journal}{ACM Transactions on Graphics} \bibinfo{volume}{36}
  (\bibinfo{year}{2017}).
\bibitem[{Pitteri et~al.(2019)Pitteri, Ramamonjisoa, Ilic, and
  Lepetit}]{pitteri2019object}
\bibinfo{author}{G.~Pitteri}, \bibinfo{author}{M.~Ramamonjisoa},
  \bibinfo{author}{S.~Ilic}, \bibinfo{author}{V.~Lepetit},
\newblock \bibinfo{title}{On object symmetries and {6D} pose estimation from
  images},
\newblock in: \bibinfo{booktitle}{Proceedings of the International Conference
  on {3D} Vision (3DV)}, \bibinfo{organization}{IEEE}, \bibinfo{year}{2019},
  pp. \bibinfo{pages}{614--622}.
\bibitem[{Lin et~al.(2022)Lin, Liu, Cheang, Fu, Guo, and Xue}]{lin2022sar}
\bibinfo{author}{H.~Lin}, \bibinfo{author}{Z.~Liu},
  \bibinfo{author}{C.~Cheang}, \bibinfo{author}{Y.~Fu},
  \bibinfo{author}{G.~Guo}, \bibinfo{author}{X.~Xue},
\newblock \bibinfo{title}{{SAR-Net}: Shape alignment and recovery network for
  category-level {6D} object pose and size estimation},
\newblock in: \bibinfo{booktitle}{Proceedings of the IEEE/CVF Conference on
  Computer Vision and Pattern Recognition}, \bibinfo{year}{2022}, pp.
  \bibinfo{pages}{6707--6717}.
\bibitem[{Wen et~al.(2022)Wen, Li, Pan, Yang, Wang, Komura, and
  Wang}]{wen2022disp6d}
\bibinfo{author}{Y.~Wen}, \bibinfo{author}{X.~Li}, \bibinfo{author}{H.~Pan},
  \bibinfo{author}{L.~Yang}, \bibinfo{author}{Z.~Wang},
  \bibinfo{author}{T.~Komura}, \bibinfo{author}{W.~Wang},
\newblock \bibinfo{title}{{DISP6D}: Disentangled implicit shape and pose
  learning for scalable {6D} pose estimation},
\newblock in: \bibinfo{booktitle}{Proceedings of the European Conference on
  Computer Vision}, \bibinfo{organization}{Springer}, \bibinfo{year}{2022}, pp.
  \bibinfo{pages}{404--421}.
\bibitem[{Sundermeyer et~al.(2018)Sundermeyer, Marton, Durner, Brucker, and
  Triebel}]{sundermeyer2018implicit}
\bibinfo{author}{M.~Sundermeyer}, \bibinfo{author}{Z.-C. Marton},
  \bibinfo{author}{M.~Durner}, \bibinfo{author}{M.~Brucker},
  \bibinfo{author}{R.~Triebel},
\newblock \bibinfo{title}{Implicit {3D} orientation learning for {6D} object
  detection from {RGB} images},
\newblock in: \bibinfo{booktitle}{Proceedings of the European Conference on
  Computer Vision}, \bibinfo{organization}{Springer}, \bibinfo{year}{2018}, pp.
  \bibinfo{pages}{699--715}.
\bibitem[{Zou et~al.(2022)Zou, Huang, Gu, and Wang}]{zou20226d}
\bibinfo{author}{L.~Zou}, \bibinfo{author}{Z.~Huang}, \bibinfo{author}{N.~Gu},
  \bibinfo{author}{G.~Wang},
\newblock \bibinfo{title}{{6D-ViT}: Category-level {6D} object pose estimation
  via transformer-based instance representation learning},
\newblock \bibinfo{journal}{IEEE Transactions on Image Processing}
  \bibinfo{volume}{31} (\bibinfo{year}{2022}) \bibinfo{pages}{6907--6921}.
\bibitem[{Fan et~al.(2021)Fan, Song, Xu, Wang, Wu, Liu, and He}]{fan2021acr}
\bibinfo{author}{Z.~Fan}, \bibinfo{author}{Z.~Song}, \bibinfo{author}{J.~Xu},
  \bibinfo{author}{Z.~Wang}, \bibinfo{author}{K.~Wu}, \bibinfo{author}{H.~Liu},
  \bibinfo{author}{J.~He},
\newblock \bibinfo{title}{{ACR-Pose}: Adversarial canonical representation
  reconstruction network for category level {6D} object pose estimation},
\newblock \bibinfo{journal}{arXiv preprint arXiv:2111.10524}
  (\bibinfo{year}{2021}).
\bibitem[{Cootes et~al.(1995)Cootes, Taylor, Cooper, and
  Graham}]{cootes1995active}
\bibinfo{author}{T.~F. Cootes}, \bibinfo{author}{C.~J. Taylor},
  \bibinfo{author}{D.~H. Cooper}, \bibinfo{author}{J.~Graham},
\newblock \bibinfo{title}{Active shape models-their training and application},
\newblock \bibinfo{journal}{Computer vision and image understanding}
  \bibinfo{volume}{61} (\bibinfo{year}{1995}) \bibinfo{pages}{38--59}.
\bibitem[{He et~al.(2022)He, Fan, Huang, Chen, and Sun}]{he2022towards}
\bibinfo{author}{Y.~He}, \bibinfo{author}{H.~Fan}, \bibinfo{author}{H.~Huang},
  \bibinfo{author}{Q.~Chen}, \bibinfo{author}{J.~Sun},
\newblock \bibinfo{title}{Towards self-supervised category-level object pose
  and size estimation},
\newblock \bibinfo{journal}{arXiv preprint arXiv:2203.02884}
  (\bibinfo{year}{2022}).
\bibitem[{Lin et~al.(2022)Lin, Wei, Ding, and Jia}]{fan2022old}
\bibinfo{author}{J.~Lin}, \bibinfo{author}{Z.~Wei}, \bibinfo{author}{C.~Ding},
  \bibinfo{author}{K.~Jia},
\newblock \bibinfo{title}{Object level depth reconstruction for category level
  {6D} object pose estimation from monocular {RGB} image},
\newblock in: \bibinfo{booktitle}{Proceedings of the European Conference on
  Computer Vision}, \bibinfo{organization}{Springer}, \bibinfo{year}{2022}, pp.
  \bibinfo{pages}{220--236}.
\bibitem[{Peng et~al.(2022)Peng, Yan, Wen, and Sun}]{peng2022self}
\bibinfo{author}{W.~Peng}, \bibinfo{author}{J.~Yan}, \bibinfo{author}{H.~Wen},
  \bibinfo{author}{Y.~Sun},
\newblock \bibinfo{title}{Self-supervised category-level {6D} object pose
  estimation with deep implicit shape representation},
\newblock in: \bibinfo{booktitle}{Proceedings of the AAAI Conference on
  Artificial Intelligence}, volume~\bibinfo{volume}{36}, \bibinfo{year}{2022},
  pp. \bibinfo{pages}{2082--2090}.
\bibitem[{Irshad et~al.(2022)Irshad, Kollar, Laskey, Stone, and
  Kira}]{irshad2022centersnap}
\bibinfo{author}{M.~Z. Irshad}, \bibinfo{author}{T.~Kollar},
  \bibinfo{author}{M.~Laskey}, \bibinfo{author}{K.~Stone},
  \bibinfo{author}{Z.~Kira},
\newblock \bibinfo{title}{{CenterSnap}: Single-shot multi-object {3D} shape
  reconstruction and categorical {6D} pose and size estimation},
\newblock in: \bibinfo{booktitle}{Proceedings of the IEEE International
  Conference on Robotics and Automation}, \bibinfo{year}{2022}, pp.
  \bibinfo{pages}{10632--10640}.
\bibitem[{Fu and Wang(2022)}]{fu2022wild6d}
\bibinfo{author}{Y.~Fu}, \bibinfo{author}{X.~Wang},
\newblock \bibinfo{title}{Category-level {6D} object pose estimation in the
  wild: A semi-supervised learning approach and a new dataset},
\newblock \bibinfo{journal}{arXiv preprint arXiv:2212.10428}
  (\bibinfo{year}{2022}).
\bibitem[{Xiang et~al.(2018)Xiang, Schmidt, Narayanan, and Fox}]{xiangposecnn}
\bibinfo{author}{Y.~Xiang}, \bibinfo{author}{T.~Schmidt},
  \bibinfo{author}{V.~Narayanan}, \bibinfo{author}{D.~Fox},
\newblock \bibinfo{title}{{PoseCNN}: A convolutional neural network for {6D}
  object pose estimation in cluttered scenes},
\newblock in: \bibinfo{booktitle}{Robotics: Science and Systems},
  \bibinfo{year}{2018}.
\bibitem[{Wang et~al.(2021)Wang, Manhardt, Tombari, and Ji}]{wang2021gdr}
\bibinfo{author}{G.~Wang}, \bibinfo{author}{F.~Manhardt},
  \bibinfo{author}{F.~Tombari}, \bibinfo{author}{X.~Ji},
\newblock \bibinfo{title}{{GDR-Net}: Geometry-guided direct regression network
  for monocular {6D} object pose estimation},
\newblock in: \bibinfo{booktitle}{Proceedings of the IEEE/CVF Conference on
  Computer Vision and Pattern Recognition}, \bibinfo{year}{2021}, pp.
  \bibinfo{pages}{16611--16621}.
\bibitem[{Zhang et~al.(2022{\natexlab{a}})Zhang, Di, Manhardt, Tombari, and
  Ji}]{zhang2022ssp}
\bibinfo{author}{R.~Zhang}, \bibinfo{author}{Y.~Di},
  \bibinfo{author}{F.~Manhardt}, \bibinfo{author}{F.~Tombari},
  \bibinfo{author}{X.~Ji},
\newblock \bibinfo{title}{{SSP-Pose}: Symmetry-aware shape prior deformation},
\newblock in: \bibinfo{booktitle}{Proceedings of the IEEE/RSJ International
  Conference on Intelligent Robots and Systems},
  \bibinfo{year}{2022}{\natexlab{a}}, pp. \bibinfo{pages}{7452--7458}.
\bibitem[{Zhang et~al.(2022{\natexlab{b}})Zhang, Di, Lou, Manhardt, Tombari,
  and Ji}]{zhang2022rbp}
\bibinfo{author}{R.~Zhang}, \bibinfo{author}{Y.~Di}, \bibinfo{author}{Z.~Lou},
  \bibinfo{author}{F.~Manhardt}, \bibinfo{author}{F.~Tombari},
  \bibinfo{author}{X.~Ji},
\newblock \bibinfo{title}{{RBP-Pose}: Residual bounding box projection for
  category-level pose estimation},
\newblock in: \bibinfo{booktitle}{Proceedings of the European Conference on
  Computer Vision}, \bibinfo{organization}{Springer},
  \bibinfo{year}{2022}{\natexlab{b}}, pp. \bibinfo{pages}{655--672}.
\bibitem[{Li et~al.(2022)Li, Li, Ye, Zhang, Kong, Cui, and
  Zhang}]{li2022generative}
\bibinfo{author}{G.~Li}, \bibinfo{author}{Y.~Li}, \bibinfo{author}{Z.~Ye},
  \bibinfo{author}{Q.~Zhang}, \bibinfo{author}{T.~Kong},
  \bibinfo{author}{Z.~Cui}, \bibinfo{author}{G.~Zhang},
\newblock \bibinfo{title}{Generative category-level shape and pose estimation
  with semantic primitives},
\newblock \bibinfo{journal}{arXiv preprint arXiv:2210.01112}
  (\bibinfo{year}{2022}).
\bibitem[{Hao et~al.(2020)Hao, Averbuch-Elor, Snavely, and
  Belongie}]{hao2020dualsdf}
\bibinfo{author}{Z.~Hao}, \bibinfo{author}{H.~Averbuch-Elor},
  \bibinfo{author}{N.~Snavely}, \bibinfo{author}{S.~Belongie},
\newblock \bibinfo{title}{{DualSDF}: Semantic shape manipulation using a
  two-level representation},
\newblock in: \bibinfo{booktitle}{Proceedings of the IEEE/CVF Conference on
  Computer Vision and Pattern Recognition}, \bibinfo{year}{2020}, pp.
  \bibinfo{pages}{7631--7641}.
\bibitem[{Zhang et~al.(2022)Zhang, Fu, Borse, Cai, Porikli, and
  Wang}]{zhang2022self}
\bibinfo{author}{K.~Zhang}, \bibinfo{author}{Y.~Fu},
  \bibinfo{author}{S.~Borse}, \bibinfo{author}{H.~Cai},
  \bibinfo{author}{F.~Porikli}, \bibinfo{author}{X.~Wang},
\newblock \bibinfo{title}{Self-supervised geometric correspondence for
  category-level {6D} object pose estimation in the wild},
\newblock \bibinfo{journal}{arXiv preprint arXiv:2210.07199}
  (\bibinfo{year}{2022}).
\bibitem[{You et~al.(2022)You, Shi, Wang, and Lu}]{you2022cppf}
\bibinfo{author}{Y.~You}, \bibinfo{author}{R.~Shi}, \bibinfo{author}{W.~Wang},
  \bibinfo{author}{C.~Lu},
\newblock \bibinfo{title}{{CPPF}: Towards robust category-level {9D} pose
  estimation in the wild},
\newblock in: \bibinfo{booktitle}{Proceedings of the IEEE/CVF Conference on
  Computer Vision and Pattern Recognition}, \bibinfo{year}{2022}, pp.
  \bibinfo{pages}{6866--6875}.
\bibitem[{Wang et~al.(2020)Wang, Mart{\'\i}n-Mart{\'\i}n, Xu, Lv, Lu, Fei-Fei,
  Savarese, and Zhu}]{wang20206pack}
\bibinfo{author}{C.~Wang}, \bibinfo{author}{R.~Mart{\'\i}n-Mart{\'\i}n},
  \bibinfo{author}{D.~Xu}, \bibinfo{author}{J.~Lv}, \bibinfo{author}{C.~Lu},
  \bibinfo{author}{L.~Fei-Fei}, \bibinfo{author}{S.~Savarese},
  \bibinfo{author}{Y.~Zhu},
\newblock \bibinfo{title}{{6-PACK}: Category-level {6D} pose tracker with
  anchor-based keypoints},
\newblock in: \bibinfo{booktitle}{Proceedings of the IEEE International
  Conference on Robotics and Automation}, \bibinfo{year}{2020}, pp.
  \bibinfo{pages}{10059--10066}.
\bibitem[{Weng et~al.(2021)Weng, Wang, Zhou, Qin, Duan, Fan, Chen, Su, and
  Guibas}]{weng2021captra}
\bibinfo{author}{Y.~Weng}, \bibinfo{author}{H.~Wang},
  \bibinfo{author}{Q.~Zhou}, \bibinfo{author}{Y.~Qin},
  \bibinfo{author}{Y.~Duan}, \bibinfo{author}{Q.~Fan},
  \bibinfo{author}{B.~Chen}, \bibinfo{author}{H.~Su}, \bibinfo{author}{L.~J.
  Guibas},
\newblock \bibinfo{title}{{CAPTRA}: Category-level pose tracking for rigid and
  articulated objects from point clouds},
\newblock in: \bibinfo{booktitle}{Proceedings of the International Conference
  on Computer Vision}, \bibinfo{year}{2021}, pp. \bibinfo{pages}{13209--13218}.
\bibitem[{Wen and Bekris(2021)}]{wen2021bundletrack}
\bibinfo{author}{B.~Wen}, \bibinfo{author}{K.~Bekris},
\newblock \bibinfo{title}{{BundleTrack}: {6D} pose tracking for novel objects
  without instance or category-level {3D} models},
\newblock in: \bibinfo{booktitle}{Proceedings of the IEEE/RSJ International
  Conference on Intelligent Robots and Systems}, \bibinfo{year}{2021}, pp.
  \bibinfo{pages}{8067--8074}.
\bibitem[{Li et~al.(2021)Li, Rezatofighi, and Reid}]{li2021moltr}
\bibinfo{author}{K.~Li}, \bibinfo{author}{H.~Rezatofighi},
  \bibinfo{author}{I.~Reid},
\newblock \bibinfo{title}{{MOLTR}: Multiple object localization, tracking and
  reconstruction from monocular {RGB} videos},
\newblock \bibinfo{journal}{IEEE Robotics and Automation Letters}
  \bibinfo{volume}{6} (\bibinfo{year}{2021}) \bibinfo{pages}{3341--3348}.
\bibitem[{Yang et~al.(2018)Yang, Feng, Shen, and Tian}]{yang2018foldingnet}
\bibinfo{author}{Y.~Yang}, \bibinfo{author}{C.~Feng},
  \bibinfo{author}{Y.~Shen}, \bibinfo{author}{D.~Tian},
\newblock \bibinfo{title}{{FoldingNet}: Point cloud auto-encoder via deep grid
  deformation},
\newblock in: \bibinfo{booktitle}{Proceedings of the IEEE Conference on
  Computer Vision and Pattern Recognition}, \bibinfo{year}{2018}, pp.
  \bibinfo{pages}{206--215}.
\bibitem[{Lorensen and Cline(1987)}]{lorensen1987marching}
\bibinfo{author}{W.~E. Lorensen}, \bibinfo{author}{H.~E. Cline},
\newblock \bibinfo{title}{Marching cubes: A high resolution {3D} surface
  construction algorithm},
\newblock in: \bibinfo{booktitle}{Proceedings of the 14th Annual Conference on
  Computer Graphics and Interactive Techniques}, \bibinfo{year}{1987}, pp.
  \bibinfo{pages}{163--169}.
\bibitem[{Xie et~al.(2022)Xie, Takikawa, Saito, Litany, Yan, Khan, Tombari,
  Tompkin, Sitzmann, and Sridhar}]{xie2022neural}
\bibinfo{author}{Y.~Xie}, \bibinfo{author}{T.~Takikawa},
  \bibinfo{author}{S.~Saito}, \bibinfo{author}{O.~Litany},
  \bibinfo{author}{S.~Yan}, \bibinfo{author}{N.~Khan},
  \bibinfo{author}{F.~Tombari}, \bibinfo{author}{J.~Tompkin},
  \bibinfo{author}{V.~Sitzmann}, \bibinfo{author}{S.~Sridhar},
\newblock \bibinfo{title}{Neural fields in visual computing and beyond},
\newblock in: \bibinfo{booktitle}{Computer Graphics Forum},
  volume~\bibinfo{volume}{41}, \bibinfo{organization}{Wiley Online Library},
  \bibinfo{year}{2022}, pp. \bibinfo{pages}{641--676}.
\bibitem[{Mescheder et~al.(2019)Mescheder, Oechsle, Niemeyer, Nowozin, and
  Geiger}]{mescheder2019occupancy}
\bibinfo{author}{L.~Mescheder}, \bibinfo{author}{M.~Oechsle},
  \bibinfo{author}{M.~Niemeyer}, \bibinfo{author}{S.~Nowozin},
  \bibinfo{author}{A.~Geiger},
\newblock \bibinfo{title}{Occupancy networks: Learning {3D} reconstruction in
  function space},
\newblock in: \bibinfo{booktitle}{Proceedings of the IEEE/CVF Conference on
  Computer Vision and Pattern Recognition}, \bibinfo{year}{2019}, pp.
  \bibinfo{pages}{4460--4470}.
\bibitem[{Chen and Zhang(2019)}]{chen2019learning}
\bibinfo{author}{Z.~Chen}, \bibinfo{author}{H.~Zhang},
\newblock \bibinfo{title}{Learning implicit fields for generative shape
  modeling},
\newblock in: \bibinfo{booktitle}{Proceedings of the IEEE/CVF Conference on
  Computer Vision and Pattern Recognition}, \bibinfo{year}{2019}, pp.
  \bibinfo{pages}{5939--5948}.
\bibitem[{Deng et~al.(2022)Deng, Bui, Navab, Guibas, Ilic, and
  Birdal}]{deng2022deep}
\bibinfo{author}{H.~Deng}, \bibinfo{author}{M.~Bui},
  \bibinfo{author}{N.~Navab}, \bibinfo{author}{L.~Guibas},
  \bibinfo{author}{S.~Ilic}, \bibinfo{author}{T.~Birdal},
\newblock \bibinfo{title}{Deep {Bingham} networks: Dealing with uncertainty and
  ambiguity in pose estimation},
\newblock \bibinfo{journal}{International Journal of Computer Vision}
  (\bibinfo{year}{2022}) \bibinfo{pages}{1--28}.
\bibitem[{Zhou et~al.(2019)Zhou, Barnes, Lu, Yang, and Li}]{zhou2019continuity}
\bibinfo{author}{Y.~Zhou}, \bibinfo{author}{C.~Barnes},
  \bibinfo{author}{J.~Lu}, \bibinfo{author}{J.~Yang}, \bibinfo{author}{H.~Li},
\newblock \bibinfo{title}{On the continuity of rotation representations in
  neural networks},
\newblock in: \bibinfo{booktitle}{Proceedings of the IEEE/CVF Conference on
  Computer Vision and Pattern Recognition}, \bibinfo{year}{2019}, pp.
  \bibinfo{pages}{5745--5753}.
\bibitem[{Qi et~al.(2017)Qi, Yi, Su, and Guibas}]{qi2017pointnet++}
\bibinfo{author}{C.~R. Qi}, \bibinfo{author}{L.~Yi}, \bibinfo{author}{H.~Su},
  \bibinfo{author}{L.~J. Guibas},
\newblock \bibinfo{title}{{PointNet++}: Deep hierarchical feature learning on
  point sets in a metric space},
\newblock \bibinfo{journal}{Advances in Neural Information Processing Systems}
  \bibinfo{volume}{30} (\bibinfo{year}{2017}) \bibinfo{pages}{5099--5108}.
\bibitem[{Ahmadyan et~al.(2021)Ahmadyan, Zhang, Ablavatski, Wei, and
  Grundmann}]{ahmadyan2021objectron}
\bibinfo{author}{A.~Ahmadyan}, \bibinfo{author}{L.~Zhang},
  \bibinfo{author}{A.~Ablavatski}, \bibinfo{author}{J.~Wei},
  \bibinfo{author}{M.~Grundmann},
\newblock \bibinfo{title}{Objectron: A large scale dataset of object-centric
  videos in the wild with pose annotations},
\newblock in: \bibinfo{booktitle}{Proceedings of the IEEE/CVF Conference on
  Computer Vision and Pattern Recognition}, \bibinfo{year}{2021}, pp.
  \bibinfo{pages}{7822--7831}.
\bibitem[{Salton and McGill(1983)}]{salton1983introduction}
\bibinfo{author}{G.~Salton}, \bibinfo{author}{M.~J. McGill},
  \bibinfo{title}{Introduction to modern information retrieval},
  \bibinfo{publisher}{McGraw Hill}, \bibinfo{year}{1983}.
\bibitem[{Everingham et~al.(2010)Everingham, Van~Gool, Williams, Winn, and
  Zisserman}]{everingham2010pascal}
\bibinfo{author}{M.~Everingham}, \bibinfo{author}{L.~Van~Gool},
  \bibinfo{author}{C.~K. Williams}, \bibinfo{author}{J.~Winn},
  \bibinfo{author}{A.~Zisserman},
\newblock \bibinfo{title}{The {PASCAL} visual object classes ({VOC})
  challenge},
\newblock \bibinfo{journal}{International Journal of Computer Vision}
  \bibinfo{volume}{88} (\bibinfo{year}{2010}) \bibinfo{pages}{303--338}.
\bibitem[{Lin et~al.(2014)Lin, Maire, Belongie, Hays, Perona, Ramanan,
  Doll{\'a}r, and Zitnick}]{lin2014microsoft}
\bibinfo{author}{T.-Y. Lin}, \bibinfo{author}{M.~Maire},
  \bibinfo{author}{S.~Belongie}, \bibinfo{author}{J.~Hays},
  \bibinfo{author}{P.~Perona}, \bibinfo{author}{D.~Ramanan},
  \bibinfo{author}{P.~Doll{\'a}r}, \bibinfo{author}{C.~L. Zitnick},
\newblock \bibinfo{title}{Microsoft {COCO}: Common objects in context},
\newblock in: \bibinfo{booktitle}{Proceedings of the European Conference on
  Computer Vision}, \bibinfo{year}{2014}, pp. \bibinfo{pages}{740--755}.
\bibitem[{Liu et~al.(2020)Liu, Jonschkowski, Angelova, and
  Konolige}]{liu2020keypose}
\bibinfo{author}{X.~Liu}, \bibinfo{author}{R.~Jonschkowski},
  \bibinfo{author}{A.~Angelova}, \bibinfo{author}{K.~Konolige},
\newblock \bibinfo{title}{{KeyPose}: Multi-view {3D} labeling and keypoint
  estimation for transparent objects},
\newblock in: \bibinfo{booktitle}{Proceedings of the IEEE/CVF Conference on
  Computer Vision and Pattern Recognition}, \bibinfo{year}{2020}, pp.
  \bibinfo{pages}{11602--11610}.
\bibitem[{Wang et~al.(2022)Wang, Jung, Li, Shen, Srikanth, Garattoni, Meier,
  Navab, and Busam}]{wang2022phocal}
\bibinfo{author}{P.~Wang}, \bibinfo{author}{H.~Jung}, \bibinfo{author}{Y.~Li},
  \bibinfo{author}{S.~Shen}, \bibinfo{author}{R.~P. Srikanth},
  \bibinfo{author}{L.~Garattoni}, \bibinfo{author}{S.~Meier},
  \bibinfo{author}{N.~Navab}, \bibinfo{author}{B.~Busam},
\newblock \bibinfo{title}{{PhoCaL}: A multi-modal dataset for category-level
  object pose estimation with photometrically challenging objects},
\newblock in: \bibinfo{booktitle}{Proceedings of the IEEE/CVF Conference on
  Computer Vision and Pattern Recognition}, \bibinfo{year}{2022}, pp.
  \bibinfo{pages}{21222--21231}.
\bibitem[{Jung et~al.(2022)Jung, Wu, Ruhkamp, Schieber, Wang, Rizzoli, Zhao,
  Meier, Roth, Navab et~al.}]{jung2022housecat}
\bibinfo{author}{H.~Jung}, \bibinfo{author}{S.-C. Wu},
  \bibinfo{author}{P.~Ruhkamp}, \bibinfo{author}{H.~Schieber},
  \bibinfo{author}{P.~Wang}, \bibinfo{author}{G.~Rizzoli},
  \bibinfo{author}{H.~Zhao}, \bibinfo{author}{S.~D. Meier},
  \bibinfo{author}{D.~Roth}, \bibinfo{author}{N.~Navab}, et~al.,
\newblock \bibinfo{title}{Housecat6d--a large-scale multi-modal category level
  6d object pose dataset with household objects in realistic scenarios},
\newblock \bibinfo{journal}{arXiv preprint arXiv:2212.10428}
  (\bibinfo{year}{2022}).
\bibitem[{{Blender Online Community}(2022)}]{blender}
\bibinfo{author}{{Blender Online Community}}, \bibinfo{title}{Blender - a {3D}
  modelling and rendering package}, \bibinfo{organization}{Blender Foundation},
  \bibinfo{address}{Blender Institute, Amsterdam}, \bibinfo{year}{2022}.
  \URLprefix \url{http://www.blender.org}.
\bibitem[{Zhou et~al.(2018)Zhou, Park, and Koltun}]{zhou2018open3d}
\bibinfo{author}{Q.-Y. Zhou}, \bibinfo{author}{J.~Park},
  \bibinfo{author}{V.~Koltun},
\newblock \bibinfo{title}{{Open3D}: A modern library for {3D} data processing},
\newblock \bibinfo{journal}{arXiv preprint arXiv:1801.09847}
  (\bibinfo{year}{2018}).
\bibitem[{Paszke et~al.(2019)Paszke, Gross, Massa, Lerer, Bradbury, Chanan,
  Killeen, Lin, Gimelshein, Antiga et~al.}]{paszke2019pytorch}
\bibinfo{author}{A.~Paszke}, \bibinfo{author}{S.~Gross},
  \bibinfo{author}{F.~Massa}, \bibinfo{author}{A.~Lerer},
  \bibinfo{author}{J.~Bradbury}, \bibinfo{author}{G.~Chanan},
  \bibinfo{author}{T.~Killeen}, \bibinfo{author}{Z.~Lin},
  \bibinfo{author}{N.~Gimelshein}, \bibinfo{author}{L.~Antiga}, et~al.,
\newblock \bibinfo{title}{{PyTorch}: An imperative style, high-performance deep
  learning library},
\newblock \bibinfo{journal}{Advances in Neural Information Processing Systems}
  \bibinfo{volume}{32} (\bibinfo{year}{2019}) \bibinfo{pages}{8026--8037}.
\bibitem[{Manhardt et~al.(2019)Manhardt, Arroyo, Rupprecht, Busam, Birdal,
  Navab, and Tombari}]{manhardt2019explaining}
\bibinfo{author}{F.~Manhardt}, \bibinfo{author}{D.~M. Arroyo},
  \bibinfo{author}{C.~Rupprecht}, \bibinfo{author}{B.~Busam},
  \bibinfo{author}{T.~Birdal}, \bibinfo{author}{N.~Navab},
  \bibinfo{author}{F.~Tombari},
\newblock \bibinfo{title}{Explaining the ambiguity of object detection and {6D}
  pose from visual data},
\newblock in: \bibinfo{booktitle}{Proceedings of the International Conference
  on Computer Vision}, \bibinfo{year}{2019}, pp. \bibinfo{pages}{6841--6850}.

\end{thebibliography}
\end{document}